%% file: arxiv_submission.tex
\documentclass{article}

\usepackage{microtype}
\usepackage{graphicx}
\usepackage{subcaption}
\usepackage{booktabs} 

\usepackage{hyperref}

\usepackage[accepted]{icml2023}

\usepackage{amsmath}
\usepackage{amssymb}
\usepackage{mathtools}
\usepackage{amsthm}
\usepackage{graphicx}
\usepackage[capitalize,noabbrev]{cleveref}

\theoremstyle{plain}
\newtheorem{theorem}{Theorem}[section]

\theoremstyle{definition}

\newtheorem{assumption}[theorem]{Assumption}
\theoremstyle{remark}

\usepackage[textsize=tiny]{todonotes}

\usepackage{cleveref}       
\usepackage{url}            
\usepackage{booktabs}       
\usepackage{nicefrac}       
\usepackage{microtype}      
\usepackage{xcolor}         

\usepackage{microtype}
\usepackage{graphicx}
\usepackage{makecell}
\usepackage{multirow}

 \usepackage{array}
 \usepackage{float}

\usepackage{colortbl}
\definecolor{bgcolor}{rgb}{0.66,0.88,1.00}

\usepackage{tablefootnote}
\usepackage{threeparttable}

\usepackage{xspace}
\usepackage{math_commands}

\xspaceaddexceptions{[]\{\}}

\usepackage{thm-restate}

\usepackage{eqparbox}

\usepackage{enumitem}

\newcommand{\ns}[1]{{\color{purple}[NS: #1]}}
\newcommand{\haoyu}[1]{{\color{orange}[HZ: #1]}}

\input{math_commands.tex}

\usepackage{nicefrac}

\usepackage{ifthen}
\newboolean{arxiv}
\setboolean{arxiv}{true}
\newboolean{arxiv1}
\setboolean{arxiv1}{false}

\usepackage{numprint}

\icmltitlerunning{Skill Localization in Fine-tuned Language Models}

\begin{document}

\twocolumn[
\icmltitle{Task-Specific Skill Localization in Fine-tuned Language Models} 



\icmlsetsymbol{equal}{*}

\begin{icmlauthorlist}
\icmlauthor{Abhishek Panigrahi}{equal,sch}
\icmlauthor{Nikunj Saunshi}{equal,sch}
\icmlauthor{Haoyu Zhao}{sch}
\icmlauthor{Sanjeev Arora}{sch}
\end{icmlauthorlist}

\icmlaffiliation{sch}{Department of Computer Science, Princeton University}

\icmlcorrespondingauthor{Abhishek Panigrahi}{ap34@princeton.edu}
\icmlcorrespondingauthor{Nikunj Saunshi}{nsaunshi@google.com}

\icmlkeywords{Machine Learning, ICML}

\vskip 0.3in
]



\printAffiliationsAndNotice{\icmlEqualContribution} 


\input{abstract.tex}

\input{Introduction.tex}

\input{table_fig/Grafting_illustration}

\input{SparsePatches.tex}

\input{PatchProperties.tex}

\input{OODCalibration.tex}

\input{MTContinual.tex}

\input{RelatedWorks.tex}

\input{Conclusions.tex}



\bibliographystyle{icml2023}
\bibliography{references.bib}

\appendix
\input{Appendix}

\end{document}

%% file: math_commands.tex
\usepackage{amsmath,amsfonts,bm}

\def\1{\bm{1}}

\makeatletter
\newcommand{\ve}{\@ifnextchar\bgroup{\velong}{{\bm{e}}}}
\newcommand{\velong}[1]{{\bm{#1}}}
\makeatother

\DeclareMathAlphabet{\mathsfit}{\encodingdefault}{\sfdefault}{m}{sl}
\SetMathAlphabet{\mathsfit}{bold}{\encodingdefault}{\sfdefault}{bx}{n}

\def\gC{{\mathcal{C}}}

\def\gL{{\mathcal{L}}}

\def\gO{{\mathcal{O}}}



%% file: abstract.tex
\begin{abstract}
 
\looseness-1  Pre-trained language models can be fine-tuned to solve diverse NLP tasks, including in few-shot settings.
Thus fine-tuning allows the model to quickly pick up task-specific ``skills,'' but there has been  limited study of  {\em where} these newly-learnt skills reside inside the massive model.
This paper introduces the term {\em skill localization} for this problem and proposes a solution. Given the downstream task and a model fine-tuned on that task, a simple optimization is used to identify a very small subset of parameters ($\sim0.01$\% of model parameters) responsible for ($>95$\%) of the model's performance, in the sense that {\em grafting} the fine-tuned values for just this tiny subset onto the pre-trained model gives a performance almost as well as the fine-tuned model. 
While reminiscent of recent works on parameter-efficient fine-tuning, the novel aspects here are that: (i) No further re-training is needed on the subset (unlike, say, with lottery tickets). (ii) Notable improvements are seen over vanilla fine-tuning with respect to calibration of predictions in-distribution ($40$-$90$\% error reduction) as well as the quality of predictions out-of-distribution (OOD).
In models trained on multiple tasks, a stronger notion of skill localization is observed, where the sparse regions corresponding to different tasks are almost disjoint, and their overlap (when it happens) is a proxy for task similarity. 
Experiments suggest that localization via grafting can assist certain forms of continual learning. Our code is available at \href{https://github.com/abhishekpanigrahi1996/Skill-Localization-by-grafting}{Skill-Localization-by-grafting}\footnote{https://github.com/abhishekpanigrahi1996/Skill-Localization-by-grafting}.
\ifthenelse{\boolean{arxiv}}{}{
\vspace{-0.1in}
}

\end{abstract}

%% file: Introduction.tex
\vspace{-0.1in} 
\section{Introduction}

\looseness-1Pre-trained language models ~\citep{liu2019roberta,devlin2019bert} have shown huge success after fine-tuning (FT) on many downstream tasks. With as few as 32 training examples, fine-tuning these giant models beats head tuning \citep{gao2021adapting}. 
Thus fine-tuning is quick to acquire new ``skills'' like sentiment identification, but locating where these skills reside in the net is an open question. {\em A priori}, one expects them to be diffused  and not be ``localized'' in a meaningful way. Better understanding of the skill's location could improve fine-tuned models -- say, with respect to accuracy,  calibration, or out-of-domain generalization--- or help mitigate catastrophic forgetting in multi-task settings.
Existing methods for parameter-efficient fine-tuning (e.g., using {\em lottery tickets}~\citep{frankle2018lottery,chen2020lottery}) suggest  the existence of more compact descriptions of the fine-tuned model, but they involve re-training the net and do not give insight into where the skills existed after vanilla fine-tuning.  (\Cref{sec:related_work} discusses past works.).

\looseness-1This paper gives a simple way to pinpoint tiny regions in the pre-trained model where the skills acquired via fine-tuning can be localized.
In particular, although fine-tuning could have updated hundreds of millions of parameters, we can identify (\Cref{sec:sparse_patches}) a tiny subset of (a few thousand) parameters whose  values are sufficient to solve the task, in the following sense: ``grafting'' the values of this tiny subset of parameters onto the pre-trained model without changing any of the other parameters, almost recovers the performance of the original fine-tuned model.
We call this tiny subset of parameters a ``grafting region.''
Note that finding sparse grafting regions allows for compact storage of the fine-tuned model, which could be important in settings where the model is being fine-tuned for a large number of tasks or users.
Crucially we find that grafted models have other desirable properties like calibration and OOD generalization.
Our main contributions are as follows. 
\begin{itemize}[leftmargin=*]
\vspace{-0.1in}
\itemsep0em 
    \item \looseness-1\Cref{sec:sparse_patches} formalizes skill localization via  {\em grafting}, and gives a simple optimization procedure to find them. \Cref{sec:patch_properties} shows that in a RoBERTa (GPT-2 resp.) model fine-tuned on GLUE tasks, regions with just $0.01$\% ($0.05$\% resp.) of parameters can recover $95$\% accuracy of the fine-tuned models, without further re-training. 

    \item \looseness-1\Cref{sec:patch_application} shows that our grafted models have much better calibrated outputs than vanilla fine-tuning --- a significant result because calibrating models can be difficult on small datasets.
    Grafted models, without re-training, also often have much better out-of-distribution (OOD) performance.
    However, re-training a sparse region (including prior parameter-efficient fine-tuning methods like BitFit) does not afford the same OOD benefits.
    These findings suggest that retaining sparse grafting regions provide a purer, more transferable way to capture the skill, while avoiding over-fitting to idiosyncrasies of the specific dataset.
    The section also discusses the generalization mystery of fine-tuning and how the graft regions begin to explain it.

    \item \Cref{sec:MT_Continual} explores consequences of our skill localization in multi-task and continual learning settings.  We show that when FT learns multiple tasks together, skills from different tasks localize in somewhat disjoint regions, where the degree of overlap between regions for two tasks seems to correlate with their intuitive similarity. We also observed some degree of compositionality: grafting the pre-trained net using regions of a subset of tasks works well for only those (and related) tasks but not others.

\end{itemize}

%% file: table_fig/Grafting_illustration.tex
\begin{figure}[!t]
    \centering
    \begin{subfigure}{0.45\textwidth}
        \centering
        \includegraphics[width=\textwidth]{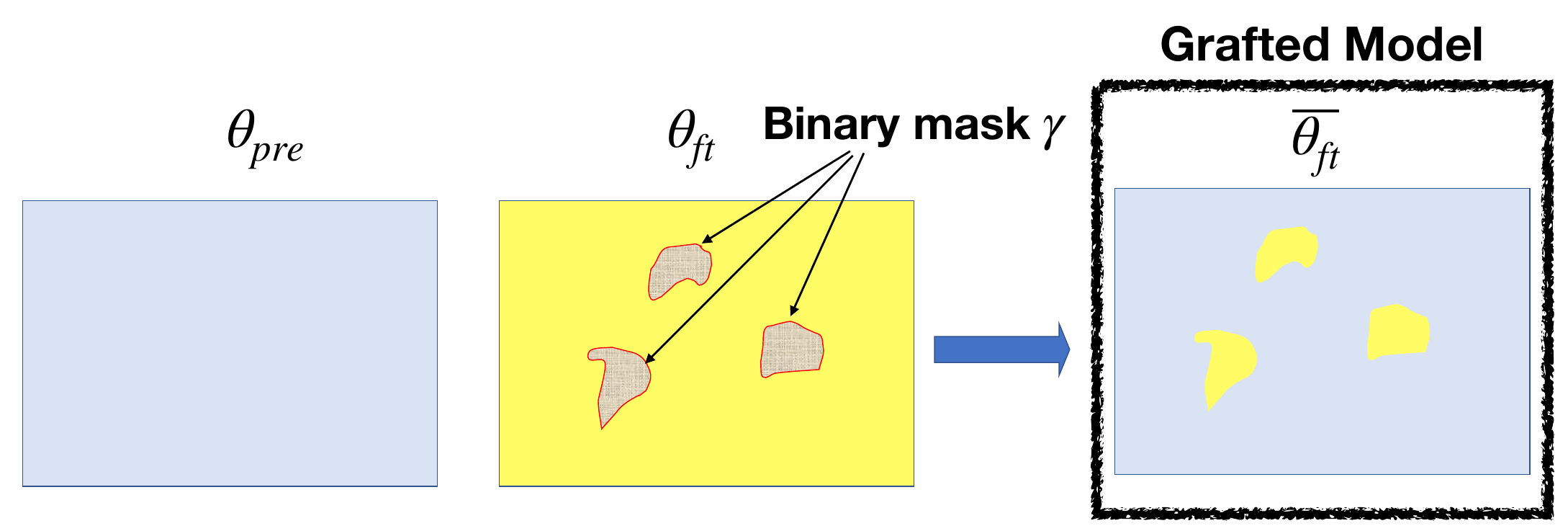}
    \end{subfigure}
    \caption{Grafting learns a binary mask $\patch$ using the fine-tuned ($\paramft$) and pre-trained ($\parampre$) models, and creates a grafted model $\patchparam(\patch)$. For parameters in the region corresponding to $\patch$, $\patchparam(\patch)$ gets its values from $\paramft$, while all other parameters default to $\parampre$. \vspace{-0.2in}}
    \label{fig:graft_illustration}
\end{figure}

%% file: SparsePatches.tex
\ifthenelse{\boolean{arxiv}}{}{
\vspace{-0.2in}
}
\section{Skill Localization through Model Grafting}
\label{sec:sparse_patches}

\looseness-1

\looseness-1Humans have concise ways of describing their skills in a modular fashion using natural language, typically as a combination of more basic skills. Such descriptions are challenging for skills in deep nets. In context of fine-tuning, recent papers have approached by equating ``skill''  with the ability to perform  a specific fine-tuning task. They correlate this ability with activations of certain subsets of neurons; for instance, \citet{wang2022finding} find ``skill neurons'' in prompt-tuned nets (not FT nets). While interesting, such notions suffer from the limitation that the activations depend on both the input to the net, as well as on a large set of parameters. Ideally, we would pinpoint the skill for the entire task in terms of specific net parameters, and   in a compact way.
 
A naive attempt at a parameter-centered formalization would be to identify parameters that change a lot during fine-tuning. 
However, this turns out to be neither concise  nor closely connected to the task in question; see \Cref{fig:topk_vs_patch}.

\ifthenelse{\boolean{arxiv}}{}{
\vspace{-0.1in}
}
\subsection{Model Grafting}
\label{sec:stitces_defn}

\input{table_fig/topk_vs_patch.tex}

\looseness-1 To formalize skill localization, we introduce {\em model grafting}.
Given a pre-trained model with parameters $\parampre$, fine-tuned model $\paramft$, we can think of a binary mask $\patch \in \{0,1\}^{|\paramft|}$ as identifying a subset of parameters, also called a ``region.'' A {\em grafting} of $\paramft$ in the region $\patch$ onto $\parampre$ is defined as 
\begin{align}
    \patchparam(\patch) &= \patch \odot \paramft + (1-\patch) \odot \parampre.
    \label{eq:patch_params}
\end{align}
\looseness-1In other words, for parameters in the region corresponding to $\patch$, $\patchparam(\patch)$ gets its values from $\paramft$, while all other parameters default to $\parampre$, yielding a ``grafted model.'' This is reminiscent of model stitching \citep{bansal2021revisiting}, where layers of one model are grafted onto the remaining layers of another model. But we allow any subset of parameters to be grafted, thus potentially affecting very tiny fraction of parameters. We desire two competing properties:
\begin{itemize}[leftmargin=*,label={}]
    \item \looseness-1{\em Good Localization:} the region $\patch$ is sparse, i.e. $\|\patch\|_{0}$ is tiny
    \item {\em Skill retention}: $\loss_{\task}(\patchparam(\patch)) \approx \loss_{\task}(\paramft)$ are both small, i.e., grafted model recovers the fine-tuning performance
\end{itemize}
\looseness-1where $\loss_{\task}$ is some metric for performance on $\task$ (e.g. classification error or logistic loss).
We note that we can rewrite \Cref{eq:patch_params} as $\patchparam(\patch) = \parampre + \patch \odot \left(\paramft - \parampre\right)$.
Thus while $\patch$ denotes the location of the skill, $\patch \odot \left(\paramft - \parampre\right)$ gives a succinct representation of the core skills acquired.
This characterization suggests a natural way to learn a grafting region $\patch$ as well; see \Cref{sec:learning_patches}.

\ifthenelse{\boolean{arxiv}}{}{
\vspace{-0.1in}
}

\subsection{Differences from Sparsity-based Fine-Tuning}

\looseness-1Grafting with sparse regions (aka {\em sparse grafting}), while reminiscent of efficient FT methods has key differences.

\textbf{Lottery tickets:}
The lottery ticket hypothesis (LTH)  \citep{frankle2018lottery} aims to prune to model by finding a sparse sub-network that when re-trained from scratch can recover the performance of vanilla training. Grafting is fundamentally different in two ways:  (a) parameters outside the grafting region are set to pre-trained values rather than 0, and (b) no re-training is needed.
Our underlying motivation is to find skills in the standard fine-tuned model, whereas re-training can change the training mechanism itself.

\textbf{Parameter-efficient fine-tuning:}
Methods like BitFit \citep{ben2022bitfit} find that updating only a small subset of parameters during fine-tuning (e.g. just biases) is very effective. However, the mechanism of parameter-efficient FT is very different from vanilla FT (e.g. biases are not important for vanilla FT).
Furthermore, the sparsity of biases ($0.1\%$) is $10 \times$ than what task-dependent grafting can achieve. 
Besides, BitFit fails to provide the calibration benefits of grafting; see \Cref{tab:bitfit_graftretrain}.

\input{table_fig/cls_vs_prompt.tex}

\ifthenelse{\boolean{arxiv}}{}{
\vspace{-0.1in}
}
\subsection{Optimization Procedure to Learn Grafted Models}
\label{sec:learning_patches}

We now describe a simple optimization procedure to learn a mask $\patch$ such that the grafted model $\patchparam(\patch)$ from \Cref{eq:patch_params} retains the skills to do well on the task $\task$:
\begin{align}
    \argmin_{ \patch \in \{0, 1\}^{|\paramft|} : \norm{ \patch }_0 \le s } \loss_{\task} (  \patch \odot \paramft + (1 - \patch) \odot \parampre )
\end{align}

Due to optimization considerations, we re-parametrize $\patch$ as a sigmoid of a real-valued vector $\optpatch$, i.e. $\patch = \sigma(\optpatch)$.
Furthermore to control the sparsity level $s$, we build the mask $\patch$ on top of an initial candidate mask $\basepatch \in \{0, 1\}^{|\paramft|}$.
So the general optimization problem reduces to solving
\begin{align}
    \argmin_{ \optpatch \in \real^{|\paramft|}}~~ &\loss_{T} (   \patch \odot \paramft + (1 - \patch) \odot \parampre ) 
    \label{eq:opt_patch}\\
    \patch \coloneqq \basepatch &\odot (1 - \sigmoid ( \optpatch ) ) + (1  - \basepatch)  \odot \sigmoid ( \optpatch )
    \label{eq:base_patch}
\end{align}
\looseness-1Our optimization procedure aims to make minimal changes (addition or deletion) to $\basepatch$ while getting low task loss.
We achieve minimal changes by initializing $\optpatch$ such that $\sigma(\optpatch)\approx \bm{0}$ and by taking only a few gradient steps to train $\optpatch$. (One could also  use $\ell_{1}$ regularization on $\sigma(\optpatch)$, but using a few gradient steps seems to suffice.)
A natural and effective choice for $\basepatch$ turns out to be the top few parameters based on their movement $|\paramft - \parampre|$.
While $\basepatch$ by itself is not great  (see \Cref{fig:topk_vs_patch}),  it tends to agree with the final localization in many coordinates.

\input{table_fig/table1_2.tex}

%% file: table_fig/topk_vs_patch.tex
\ifthenelse{\boolean{arxiv1}}
{
\begin{figure}[!t]
    \centering
    \begin{subfigure}{0.48\textwidth}
    \centering
    \includegraphics[width=\textwidth]{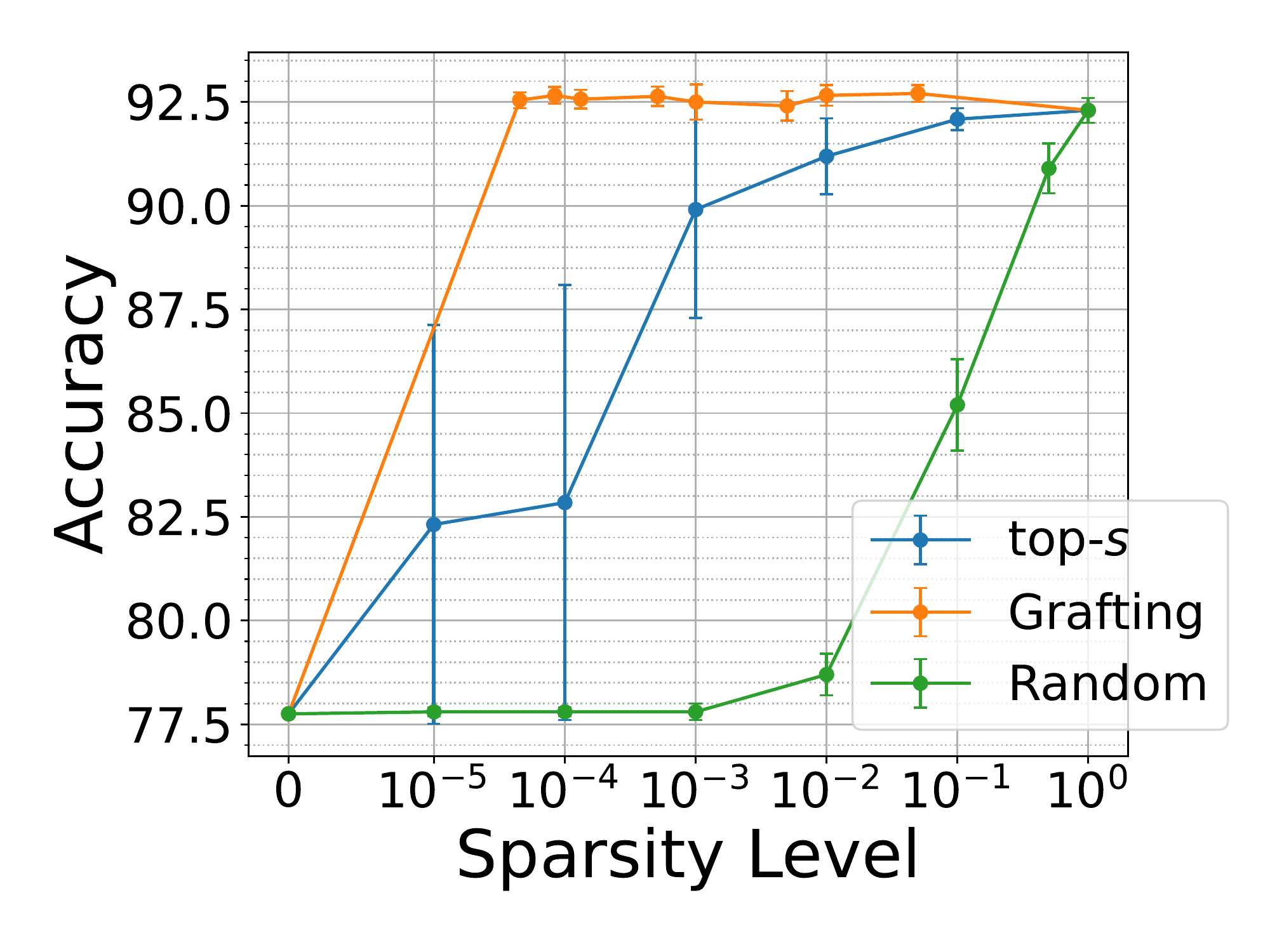}\caption { SST-2, 4096-shot}
    \end{subfigure}\hfill
    \begin{subfigure}{0.48\textwidth}
    \centering
    \includegraphics[width=\textwidth]{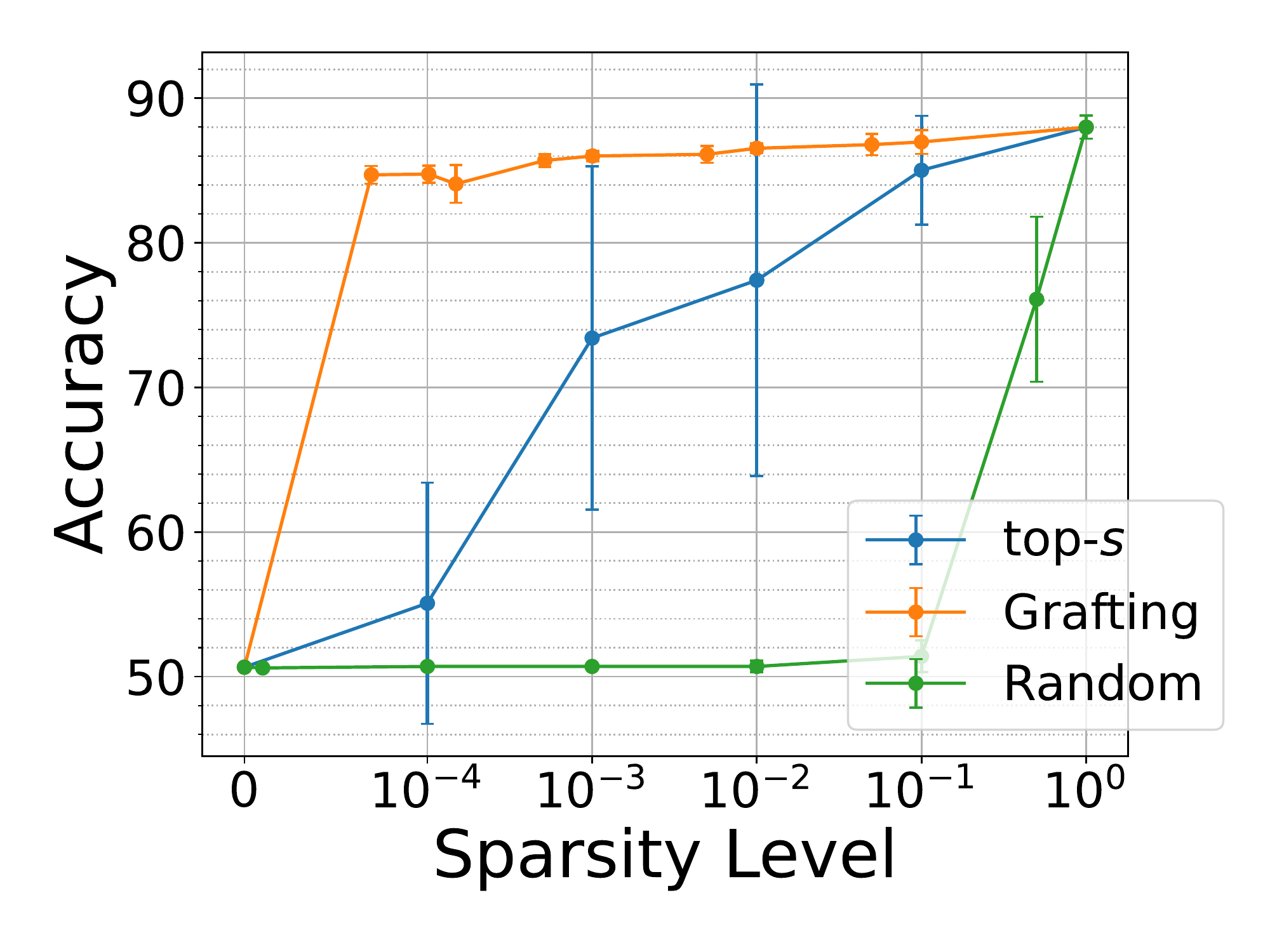}\caption { QNLI, 4096-shot }
    \end{subfigure}\hfill
    \caption{Accuracies of the grafting regions learned using our procedure in \Cref{sec:learning_patches} and regions corresponding to the top-$\sparse$ parameters based on magnitude of movement during FT. The learned region performs much better at low sparsity levels.
    }
    \label{fig:topk_vs_patch}
\end{figure}
}
{
\begin{figure}[!t]
    \centering
    \begin{subfigure}{0.24\textwidth}
    \centering
    \includegraphics[width=\textwidth]{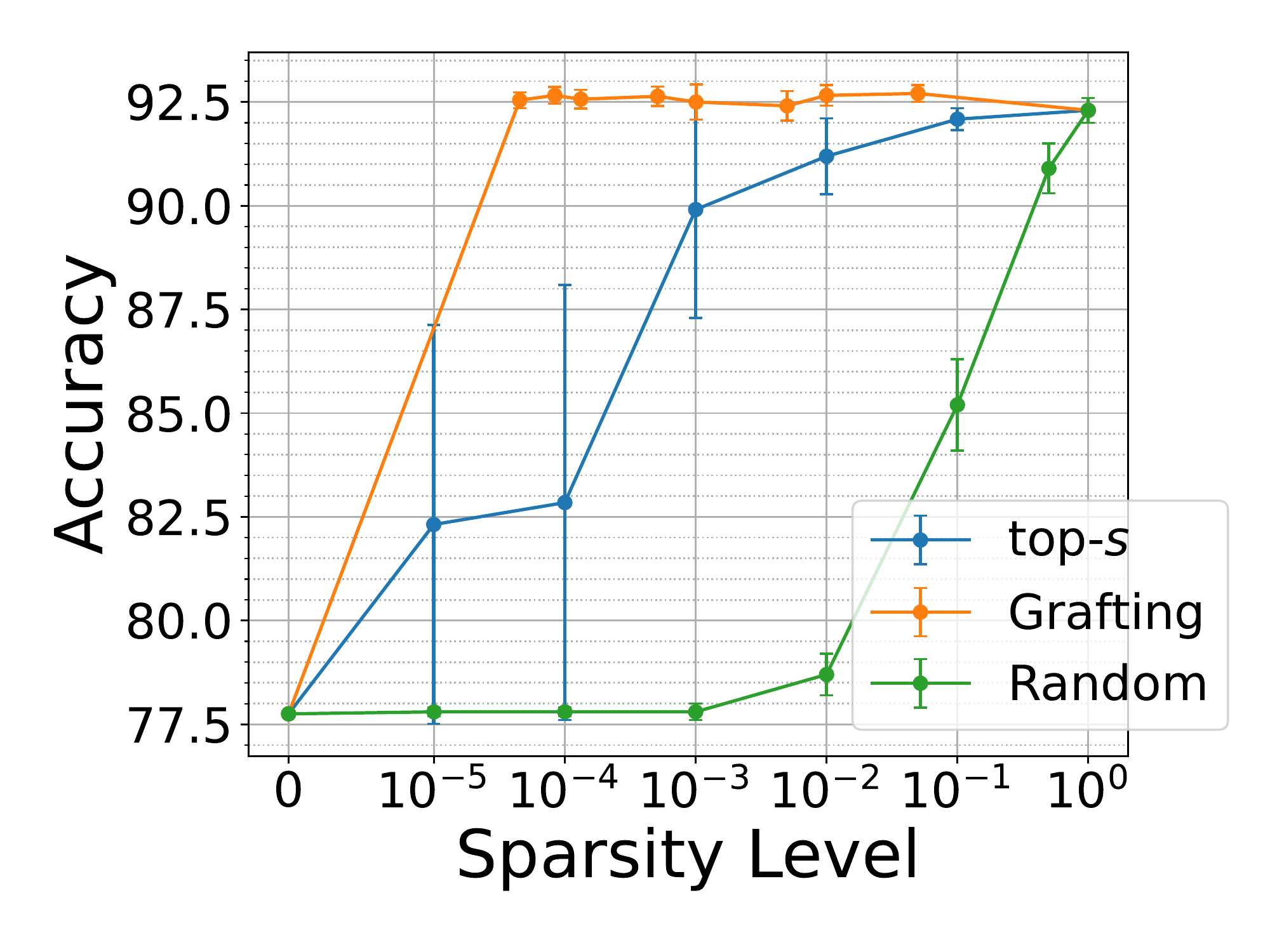}\caption { SST-2, 4096-shot}
    \end{subfigure}\hfill
    \begin{subfigure}{0.24\textwidth}
    \centering
    \includegraphics[width=\textwidth]{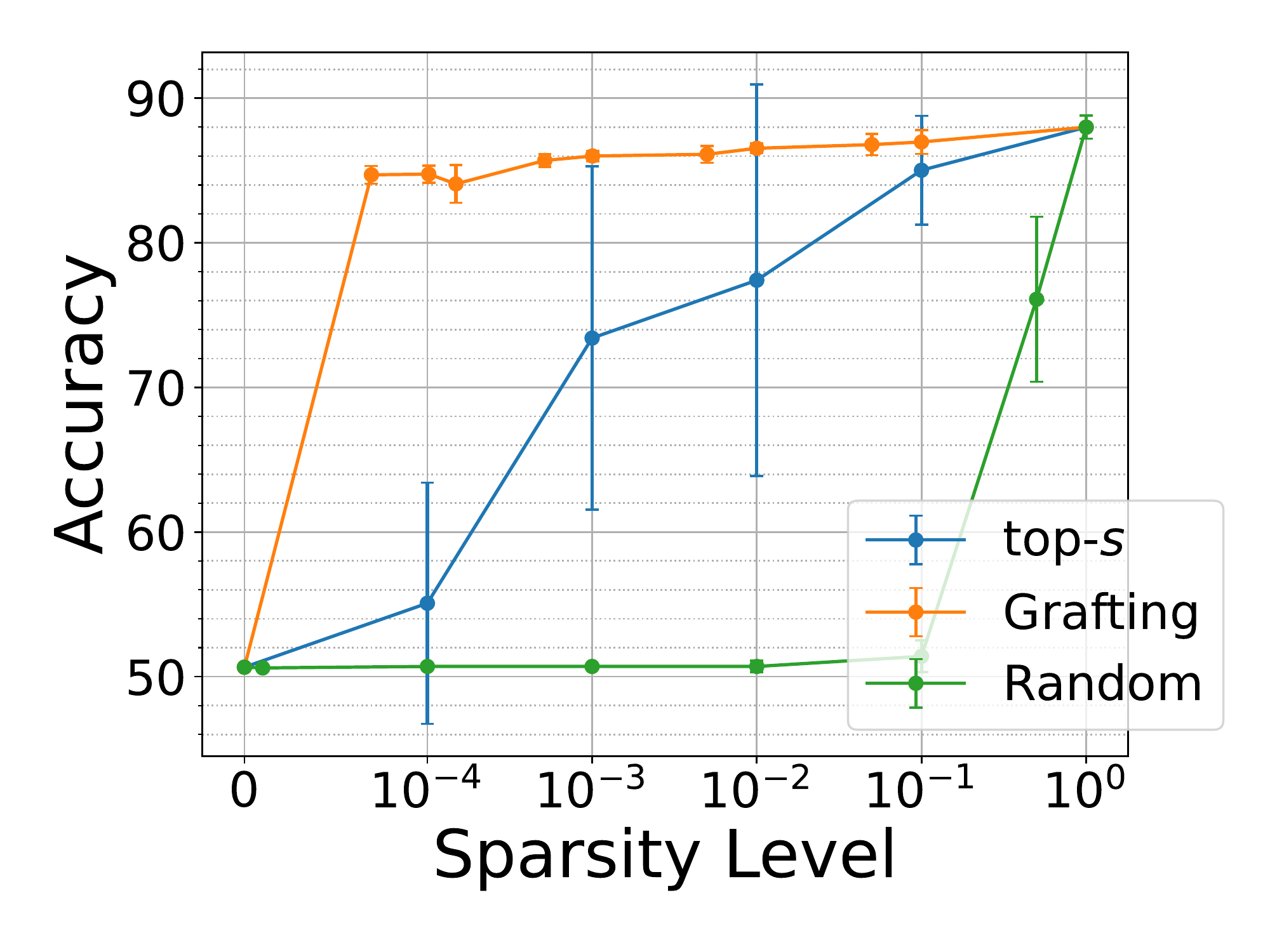}\caption { QNLI, 4096-shot }
    \end{subfigure}\hfill
    \caption{Accuracies of the grafting regions learned using our procedure in \Cref{sec:learning_patches}, regions corresponding to the top-$\sparse$ parameters based on the magnitude of movement during FT, and random regions. The learned region performs much better at low sparsity levels.
    \vspace{-0.2in}
    }
    \label{fig:topk_vs_patch}
\end{figure}
}

%% file: table_fig/cls_vs_prompt.tex
\begin{figure}[!t]
    \centering
    \begin{subfigure}{0.24\textwidth}
    \centering
    \includegraphics[width=\textwidth]{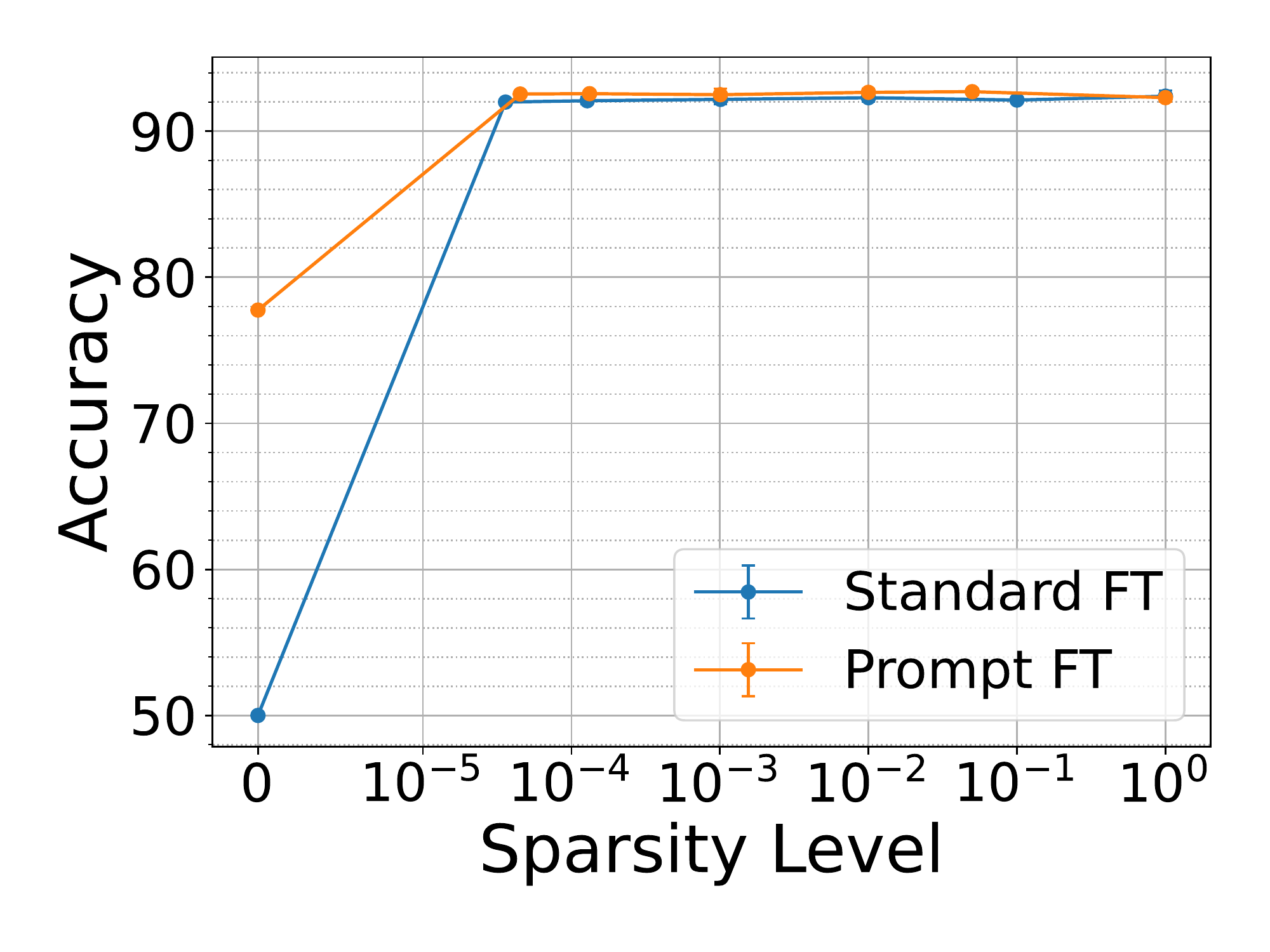}\\\caption{SST-2, 4096-shot}
    \label{fig:cls_vs_prompt_a}
    \end{subfigure}\hfill
    \begin{subfigure}{0.24\textwidth}
    \centering
    \includegraphics[width=\textwidth]{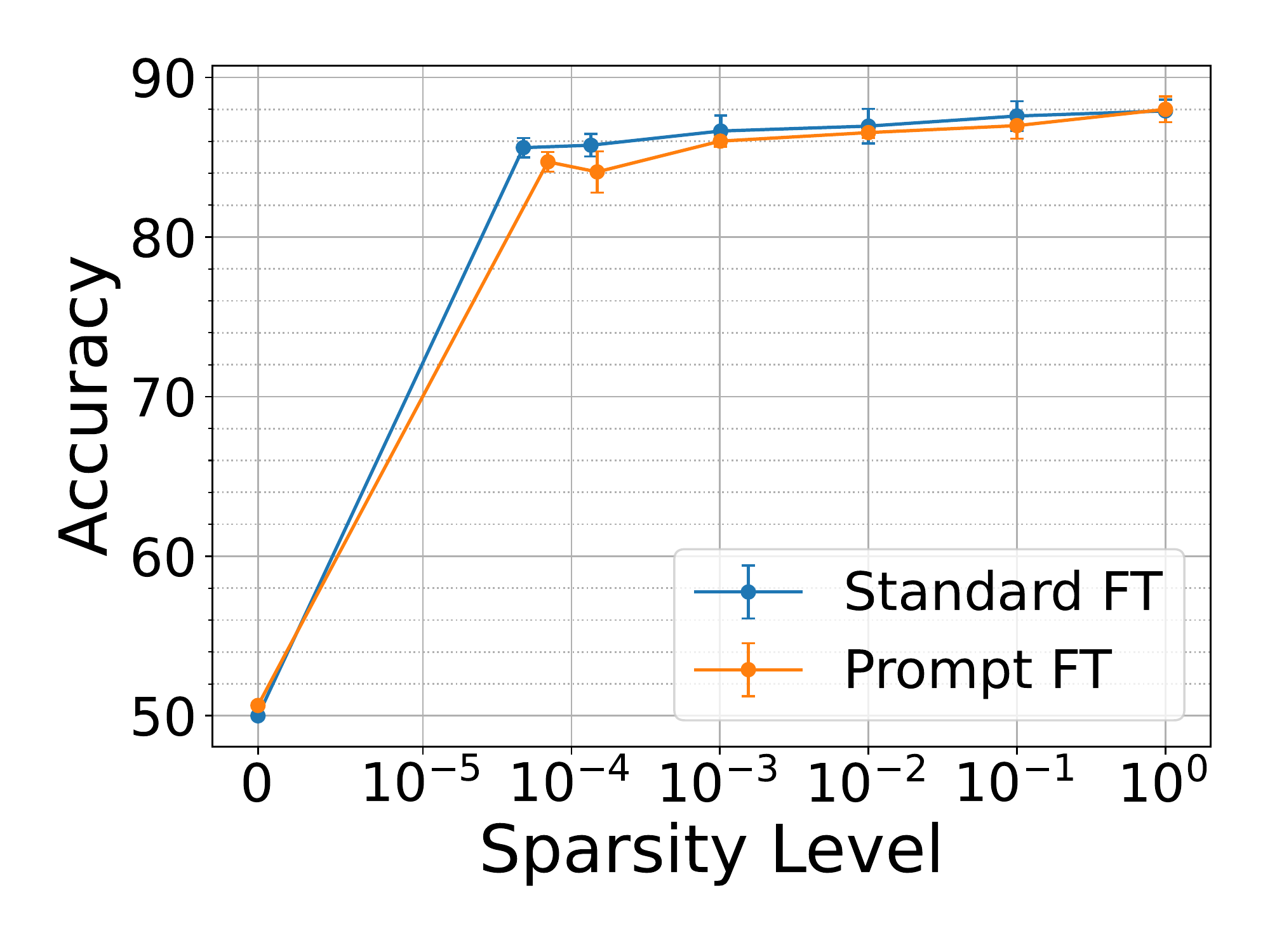}\\\caption{QNLI, 4096-shot}
    \label{fig:cls_vs_prompt_b}
    \end{subfigure}
    \caption{\looseness-1Testing existence of sparse grafting regions for prompt-based FT and standard FT fine-tuning (uses a linear head on top of [CLS] token). Skill localization is equally good for FT approaches.}
    \label{fig:cls_vs_prompt}
\end{figure}

%% file: table_fig/table1_2.tex
\begin{table*}[!t]
\centering
\caption{ \looseness-1Comparing the accuracy (${}^\dagger$F1) in \% and the calibration error (ECE) for fine-tuned model (FT), model grafting, graft re-training for grafting region with sparsity $0.01$\%, and BitFit. Results are in the 4096-shot setting. Re-training the grafting regions from scratch is not only good, but performs slightly better than the grafted model, implying that the grafting regions form a sub-network of their own. It also retains the good calibration of non re-trained grafting.
BitFit updates all biases (sparsity $0.1$\%) and achieves slightly better accuracy than grafting, but it has significantly worse calibration error compared to the grafted model.\vspace{0.1in} }
\label{tab:bitfit_graftretrain}
\footnotesize
\begin{tabular}{l|cc|cc|cc|cc} \toprule
  \midrule
     & \multicolumn{2}{c|}{FT} &  \multicolumn{2}{c|}{Grafting} &  \multicolumn{2}{c|}{Graft re-training} &  \multicolumn{2}{c}{BitFit}\\
  \midrule
 Dataset & Acc./F1 & ECE & Acc./F1 & ECE & Acc./F1 & ECE & Acc./F1 & ECE  \\
 \midrule
 SST-2 & 92.3 {\tiny (0.3)} & 7.4 {\tiny (0.3)} & 92.4 {\tiny (0.1)} & 3.1 {\tiny (0.4)} & 92.2 {\tiny (0.7)} & 3.9 {\tiny (0.7)} & 92.4 {\tiny (0.6)} & 6.7 {\tiny (0.8)}\\
AGNews & 92.7 {\tiny (0.4)} & 6.8 {\tiny (0.3)} & 91.1 {\tiny (0.9)} & 0.9 {\tiny (0.2)} & 91.2 {\tiny (0.1)} & 1.0 {\tiny (0.2)} & 93.0 {\tiny (0.2)} & 4.4 {\tiny (0.2)}\\
QNLI & 88.0 {\tiny (0.8)} & 10.2 {\tiny (0.0)} & 84.7 {\tiny (0.6)} & 1.0 {\tiny (0.3)} & 87.4 {\tiny (0.5)} & 2.0 {\tiny (1.1)} & 87.8 {\tiny (0.6)} & 8.1 {\tiny (2.3)}\\
QQP$^\dagger$ & 79.6 {\tiny (0.1)} & 10.1 {\tiny (4.2)} & 76.3 {\tiny (0.4)} & 3.5 {\tiny (0.7)} & 78.1 {\tiny (0.7)} & 4.3 {\tiny (1.3)} & 79.4 {\tiny (0.3)} & 9.7 {\tiny (1.2)}\\
 \bottomrule
\end{tabular}
\vspace{-0.1in}
\end{table*}

%% file: PatchProperties.tex
\ifthenelse{\boolean{arxiv}}{}{
\vspace{-0.1in}
}
\section{Evaluating Grafting for Skill Localization}
\label{sec:patch_properties}


\paragraph{Experimental Setup.}
\label{sec:exp_setup}

\looseness-1We fine-tuned the pre-trained RoBERTa-base \citep{liu2019roberta} model on 13 different tasks, with the majority from GLUE \citep{wang2018glue}, including sentiment analysis,  topic classification, natural language inference, and paraphrase detection datasets.
All experiments, unless specified otherwise, use prompt-based fine-tuning with the human-generated prompts from \citet{gao2021making}, and SGD optimizer, which achieves similar performance as AdamW \citep{loshchilov2017decoupled} after fixing the embedding layer (which was also observed in \citet{kumar2022fine} but in the vision setting). 
For $64$-shot and $4096$-shot experiments, we report performance across $5$ randomly sampled datasets. Unless mentioned otherwise, we always report our performance on 4096-shot setting.
Other training and hyperparameter details can be found in \Cref{sec:hyperparam}.

\looseness-1Model grafting experiments optimize \Cref{eq:opt_patch} using SGD with batch size $1024$ (full-batch GD for $64$-shot) for $100$ steps with learning rate $10^7$.
For patches of varying sizes, we use $\basepatch$ as the top-$\sparse$ fraction of parameters based their movement $|\paramft - \parampre|$ for $\sparse$ in $[0, 1]$.

\input{table_fig/roberta_patch_single.tex}

\input{table_fig/bias_adam.tex}

\ifthenelse{\boolean{arxiv}}{}{
\vspace{-0.1in}
}
\subsection{Sparse Grafting Retain Skills}
\label{sec:patch_skills}

\looseness-1The first experiment  compares  the performance of model grafting with sparse regions versus full model (vanilla) fine-tuning. 
For each downstream task and prompt fine-tuned model, we learn a grafting region $\patch$ of sparsity at most $0.01$\% by building on top of $\basepatch \coloneqq \text{top-}(10^{-5})$, where $\text{top-}\sparse$ selects the top $s$ fraction parameters based on parameter movement.
We report the accuracies of these grafted models in  \Cref{tab:roberta_patch_perf}.
The main observation is that sparse grafting can recover at least $95$\% of FT performance for all datasets.
Additionally, the grafted models have a high agreement (on test set labels) with the original FT models: $93\%$ (resp.\ $86\%$) for single-sentence (resp.\ two-sentence) experiments, suggesting good skill ``localization.'' Please see \cref{sec:distribution} for a closer analysis on the distribution of the graft parameters. We have similar observations for models trained with larger training data. Please see \cref{sec:fullshot}.

\looseness-1\textbf{Performance of $\basepatch$}:  We compare the performance of the learned regions $\gamma$ using optimization versus that of $\basepatch$ using top-$\sparse$ most-changed parameters, at different levels of sparsity in \cref{fig:topk_vs_patch}. We  find that the optimization method  is much more effective, especially at lower sparsity levels.
Exploring other ways to learn the regions, perhaps directly from the pre-trained model, is an interesting open question.

\looseness-1\textbf{Gains from re-training the grafting regions.} 
To check whether the learned region forms a meaningful sub-network within the model, we re-train starting from pre-trained initialization, but only make updates to parameters in the learned region $\patch$, akin to parameter-efficient FT.
 \cref{tab:bitfit_graftretrain} shows that re-training the sparse regions from scratch  performs well, almost always better than the grafted models. This suggests that the sparse sub-network represented by $\patch$ is also trainable while being significantly sparser ($0.01$\%) than the set of biases used in BitFit ($0.1$\%).

\textbf{Differences with BitFit and lottery tickets:}  Since BitFit succeeds by only training biases, we check whether biases can form good grafting regions. \Cref{fig:bias} answers this in the negative, which implies a stark difference in mechanism between standard fine-tuning and BitFit. 
Furthermore, we check whether lottery tickets style pruning can work, without re-training, i.e. can we learn a sparse region such that setting other parameters to 0 can yield a good model. 
We find in \Cref{fig:lth} that even regions as dense as $90$\% of parameters fail to capture any skill, without re-training.
The high density is consistent with prior works on lottery tickets for language model fine-tuning \citep{chen2020lottery}, where the sparsity is usually higher than $10$\%, much denser compared to our grafting regions of $0.01$\%.

\subsection{Other Fine-Tuning Paradigms}
\label{sec:other_FT}

Experiments in the previous sections use prompt-based fine-tuning with SGD optimizer.
In this section, we check whether sparse grafting regions (i.e. skill localization) exist in models fine-tuned differently.

\looseness-1\textbf{Standard FT.} Instead of prompt-based FT, we consider fine-tuning with a linear head on top of the [CLS] token representation, which was the {\em standard FT} approach before prompt-tuning \citep{liu2019roberta}.
\cref{fig:cls_vs_prompt} confirms that similar sparse localization is also possible for Standard FT.

\looseness-1\textbf{AdamW optimizer.} In \cref{fig:adam_sgd} we test skill localization with AdamW \citep{loshchilov2017decoupled} optimizer on prompt-based FT.
Unlike SGD, we find that fine-tuning with AdamW {\em does not} contain sparse grafted models with good performance. However, adding an explicit $\ell_1$ regularization (with strength $0.001$) on the parameter movement $\|\paramft - \parampre\|_{1}$ can recover sparse grafts. This suggests that $\ell_{1}$ regularization could be a way to encourage skill localization. An extensive exploration of this is left for future work.

\input{table_fig/adam_sgd_l1.tex}

%% file: table_fig/roberta_patch_single.tex
\begin{table*}
\centering
\caption{\looseness-1For each downstream task, we learn a grafting region $\patch$ using our optimization procedure in \cref{eq:opt_patch}. The grafting regions for all tasks have sparsity at most $0.01\%$ ($<8500$ parameters).
We report test accuracy (${}^\dagger$F1) and the calibration error using ECE of the fine-tuned model and the grafted model for each task. The main findings are (1) The grafted model can retrieve $>95\%$ of the FT accuracy, while being better calibrated than the original model itself.
For single-sentence tasks (4096-shot) the grafted model shows only a $0.7\%$ drop in accuracy but an improvement of $5\%$ in the calibration error. Similarly for two-sentence tasks, the grafted model shows a $3.4\%$ drop in accuracy with an improvement of $9.6\%$ in the calibration error.\vspace{0.1in}}
\label{tab:roberta_patch_perf}
\footnotesize
\begin{tabular}{c|l|cc|cc|cc|cc|c} 
\toprule
  & & \multicolumn{4}{c|}{64-shot} &  \multicolumn{5}{c}{4096-shot} \\
  \midrule
  &    & \multicolumn{2}{c|}{FT} & \multicolumn{2}{c|}{Graft} & \multicolumn{2}{c|}{FT} & \multicolumn{2}{c|}{Graft} &  \\
  \midrule
 & Dataset & Acc. & ECE & Acc. & ECE & Acc. & ECE & Acc. & ECE & Agreement\\
 \midrule
\parbox[t]{2mm}{\multirow{8}{*}{\rotatebox[origin=c]{90}{Single sent. tasks}}} &
 SST-2 & 90.5 {\tiny (0.4)} & 9.7 {\tiny (0.3)} & 89.7 {\tiny (0.2)} & 7.8 {\tiny (0.6)} & 92.3 {\tiny (0.3)} & 7.4 {\tiny (0.3)} & 92.4 {\tiny (0.1)} & 3.1 {\tiny (0.4)} & 95.3 {\tiny (0.6)}\\
& CR & 90.2 {\tiny (0.6)} & 8.2 {\tiny (2.3)} & 89.5 {\tiny (1.1)} & 5.7 {\tiny (1.9)} & 91.7 {\tiny (0.2)} & 8.0 {\tiny (0.3)} & 91.7 {\tiny (0.5)} & 5.0 {\tiny (0.3)} & 96.6 {\tiny (0.5)}\\
& MR & 85.0 {\tiny (1.2)} & 22.9 {\tiny (2.1)} & 85.2 {\tiny (1.7)} & 10.8 {\tiny (2.3)} & 89.7 {\tiny (0.3)} & 9.0 {\tiny (0.6)} & 89.1 {\tiny (1.1)} & 1.5 {\tiny (0.2)} & 93.6 {\tiny (0.8)}\\
& MPQA & 85.4 {\tiny (0.9)} & 14.2 {\tiny (0.9)} & 84.1 {\tiny (1.2)} & 11.4 {\tiny (1.8)} & 88.9 {\tiny (0.6)} & 10.5 {\tiny (0.6)} & 88.1 {\tiny (0.4)} & 3.3 {\tiny (0.2)} & 93.3 {\tiny (0.2)}\\
& TREC & 93.1 {\tiny (1.7)} & 6.1 {\tiny (1.1)} & 86.8 {\tiny (0.7)} & 4.8 {\tiny (1.2)} & - & - & - & - & -\\
& AGNews & 88.2 {\tiny (0.3)} & 10.1 {\tiny (0.5)} & 86.8 {\tiny (0.3)} & 7.1 {\tiny (0.5)} & 92.7 {\tiny (0.4)} & 6.8 {\tiny (0.3)} & 91.1 {\tiny (0.2)} & 0.9 {\tiny (0.2)} & 95.1 {\tiny (0.5)}\\
& Subj & 91.2 {\tiny (1.2)} & 5.9 {\tiny (1.2)} & 91.7 {\tiny (1.2)} & 2.6 {\tiny (1.2)} & 96.7 {\tiny (0.1)} & 3.0 {\tiny (0.1)} & 95.5 {\tiny (0.1)} & 1.2 {\tiny (0.1)} & 97.3 {\tiny (0.2)}\\
\cmidrule{2-11}
& Avg. & 89.1 & 11.0 & 87.7 & 7.2 & 92 & 7.5 & 91.3 & 2.5 & 95.2 \\
\midrule
\parbox[t]{2mm}{\multirow{7}{*}{\rotatebox[origin=c]{90}{Two sent. tasks}}} &
 QNLI & 77.8 {\tiny (0.9)} & 21.1 {\tiny (0.9)} & 76.7 {\tiny (1.5)} & 12.3 {\tiny (0.8)} & 88.0 {\tiny (0.8)} & 10.2 {\tiny (0.0)} & 84.7 {\tiny (0.6)} & 1.0 {\tiny (0.3)} & 88.9 {\tiny (0.3)}\\
& SNLI & 76.5 {\tiny (1.7)} & 20.8 {\tiny (1.6)} & 72.1 {\tiny (1.9)} & 14.4 {\tiny (3.1)} & 86.4 {\tiny (0.3)} & 10.6 {\tiny (1.7)} & 82.7 {\tiny (0.5)} & 1.1 {\tiny (0.4)} & 87.5 {\tiny (1.5)}\\
& MNLI & 67.5 {\tiny (2.1)} & 29.5 {\tiny (2.0)} & 64.6 {\tiny (2.5)} & 20.4 {\tiny (1.4)} & 81.8 {\tiny (0.1)} & 14.8 {\tiny (0.7)} & 78.0 {\tiny (0.4)} & 1.5 {\tiny (0.0)} & 86.4 {\tiny (0.4)}\\
& RTE & 66.9 {\tiny (3.5)} & 31.0 {\tiny (3.5)} & 66.2 {\tiny (5.0)} & 21.5 {\tiny (3.9)} & 80.0 {\tiny (2.2)} & 20.2 {\tiny (1.7)} & 77.7 {\tiny (0.5)} & 8.5 {\tiny (1.7)} & 88.6 {\tiny (1.7)}\\
& MRPC$^{\dagger}$ & 82.5 {\tiny (1.9)} & 22.9 {\tiny (2.1)} & 76.9 {\tiny (2.4)} & 19.1 {\tiny (3.5)} & 90.0 {\tiny (0.7)} & 13.0 {\tiny (0.8)} & 86.2 {\tiny (0.5)} & 6.2 {\tiny (0.1)} & 88.6 {\tiny (1.7)}\\
& QQP$^{\dagger}$ & 68.7 {\tiny (1.3)} & 26.5 {\tiny (3.7)} & 66.9 {\tiny (1.0)} & 17.3 {\tiny (2.1)} & 79.6 {\tiny (0.6)} & 10.1 {\tiny (4.2)} & 76.3 {\tiny (0.4)} & 3.5 {\tiny (0.7)} & 93.3 {\tiny (4.9)}\\
\cmidrule{2-11}
& Avg. & 73.3 & 25.3 & 70.6 & 17.5 &  84.3 & 13.2 & 80.9 & 3.6 & 88.9 \\
 \bottomrule
\end{tabular}
\end{table*}

%% file: table_fig/bias_adam.tex
\begin{figure}[!t]
    \centering
    \begin{subfigure}{0.24\textwidth}
    \centering
    \includegraphics[width=\textwidth]{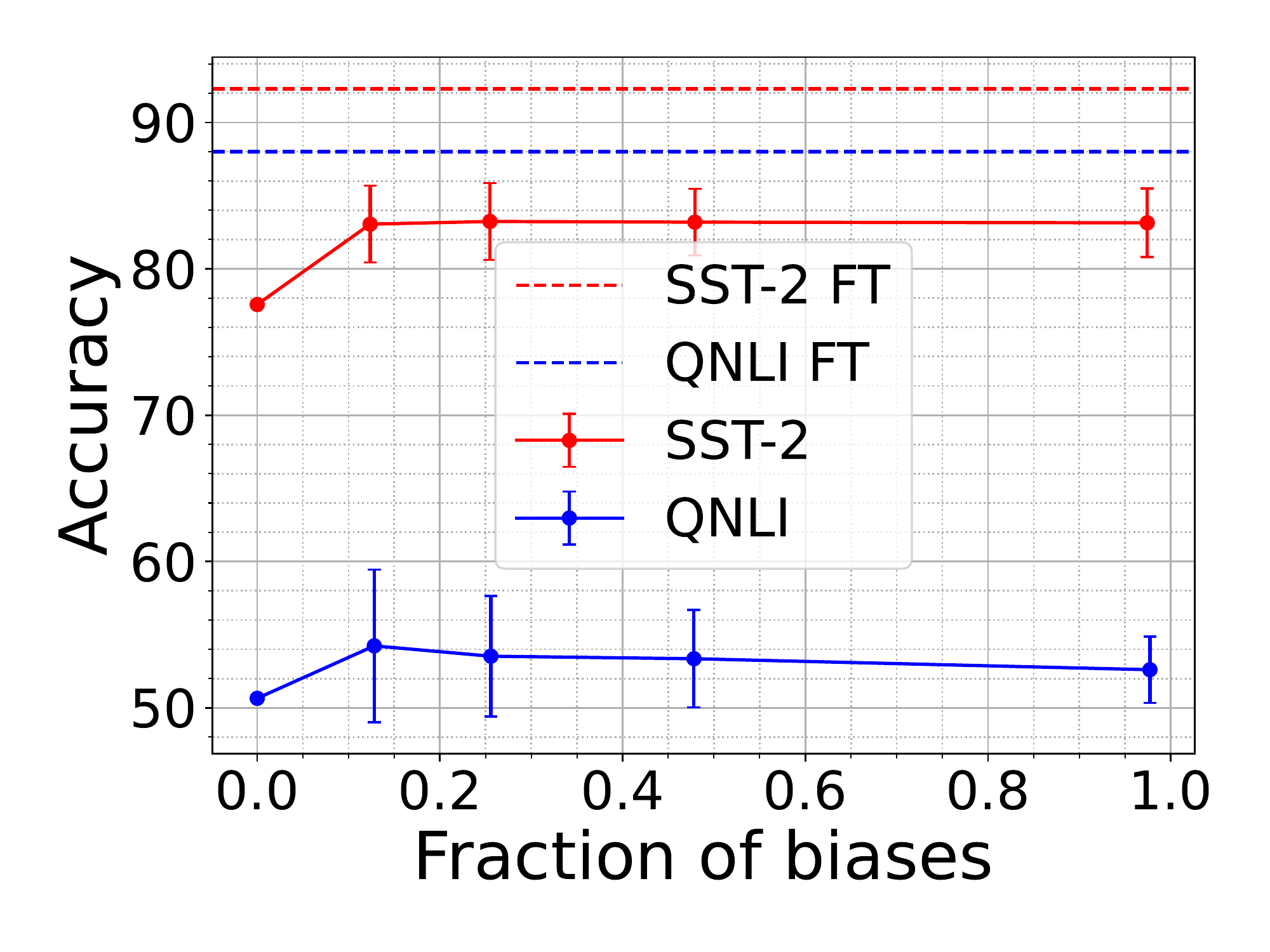}\caption{Bias-only grafting}
    \label{fig:bias}
    \end{subfigure}\hfill
    \begin{subfigure}{0.24\textwidth}
    \centering
    \includegraphics[width=\textwidth]{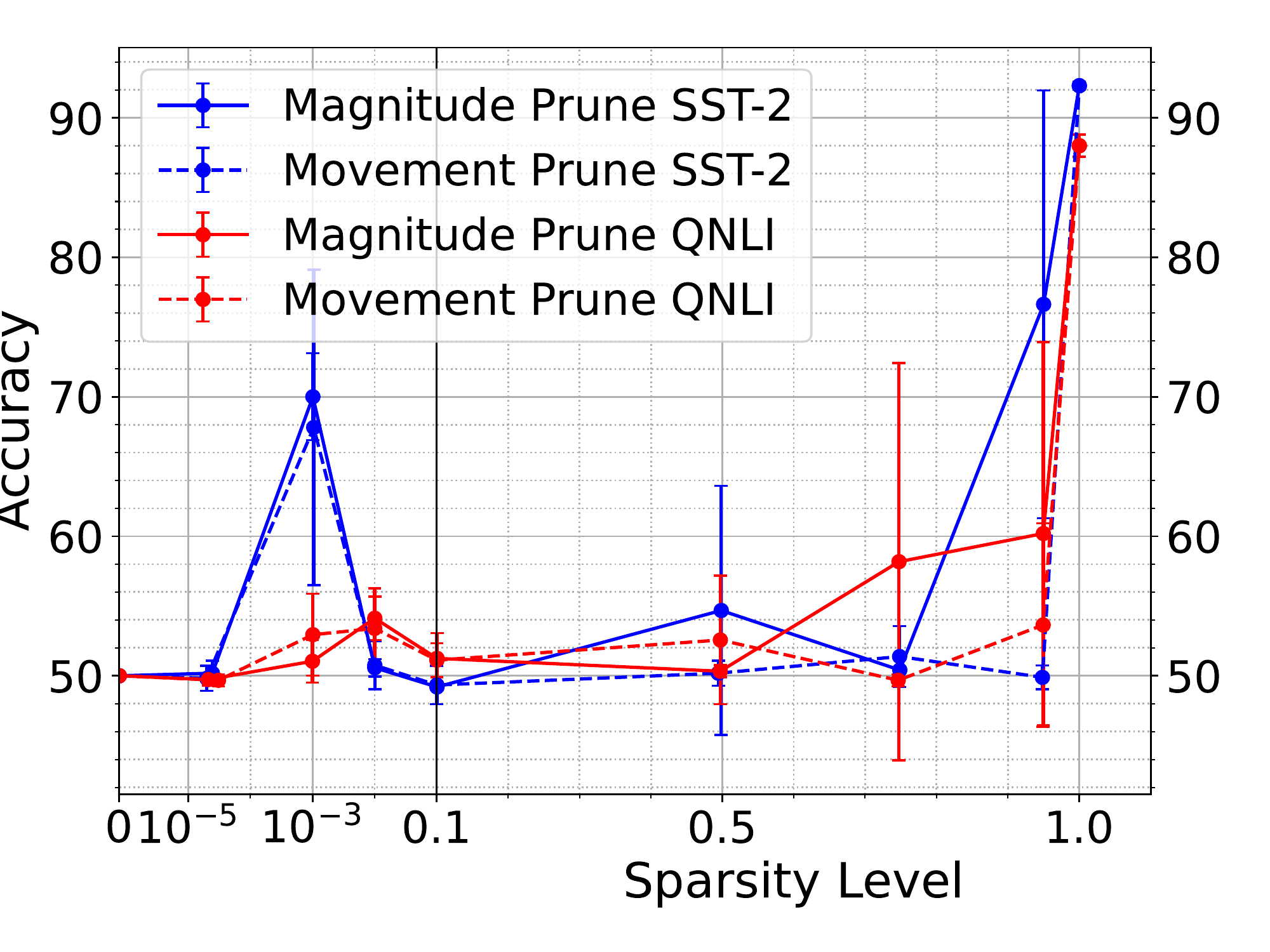} \caption{Lottery tickets pruning}
    \label{fig:lth}
    \end{subfigure}\hfill
    \caption{\looseness-1(a) Grafting regions that only contain biases of the model don't give good skill localization. (b) Localizing with lottery ticket pruning (setting remaining parameters to $0$) does not perform well at any sparsity level without re-training.\vspace{-0.2in}}
    \label{fig:bias_lth}
\end{figure}

%% file: table_fig/adam_sgd_l1.tex
\begin{figure}[!t]
    \centering
    \begin{subfigure}{0.24\textwidth}
    \centering
    \includegraphics[width=\textwidth]{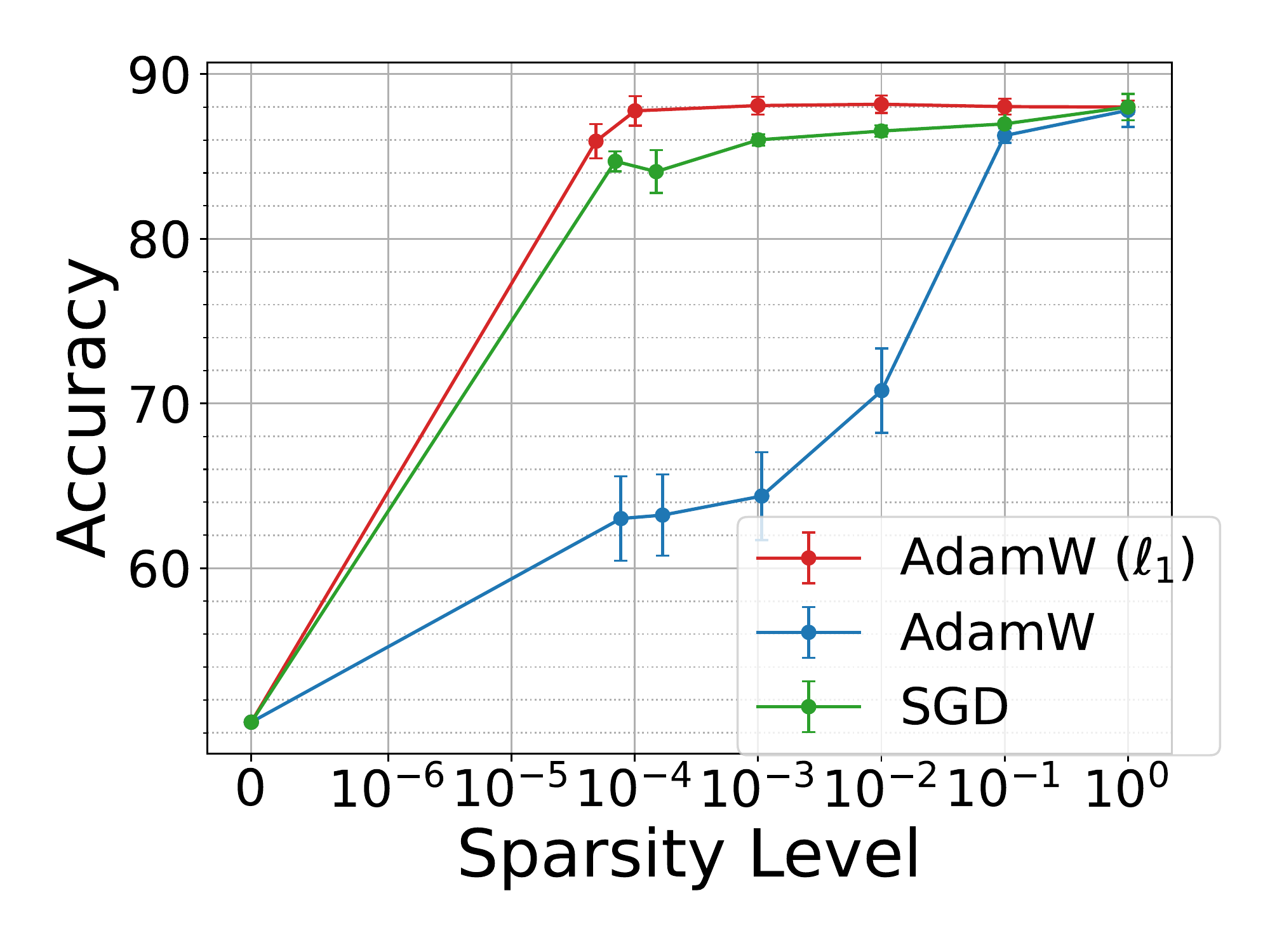}
    \caption{ QNLI, 4096-shot }
    \label{fig:adam_sgd_qnli}
    \end{subfigure}\hfill
    \begin{subfigure}{0.24\textwidth}
    \centering
    \includegraphics[width=\textwidth]{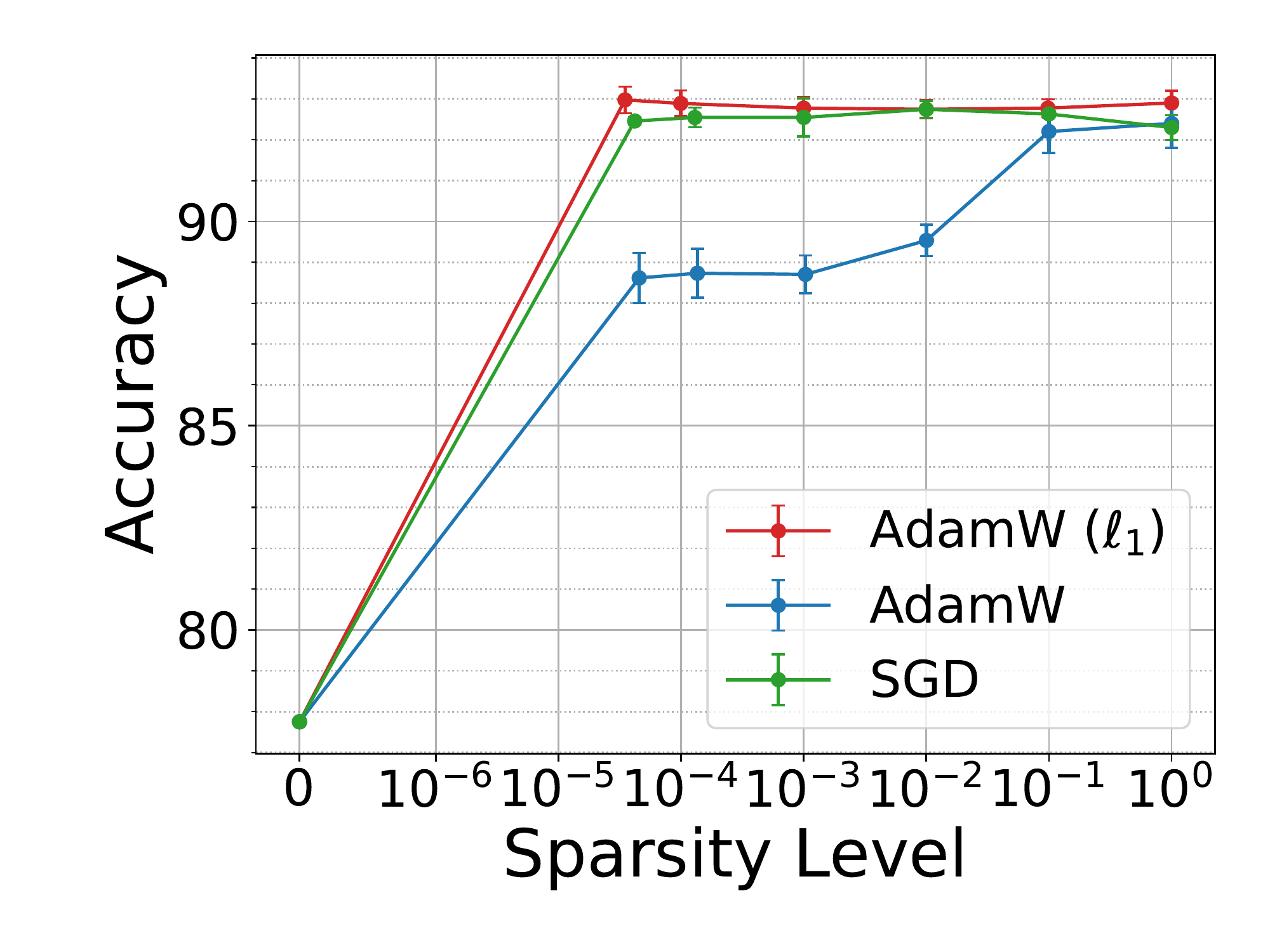}\caption{ SST-2, 4096-shot}
    \label{fig:adam_sgd_sst-2}
    \end{subfigure}\hfill
    \caption{Grafting accuracy for FT with SGD and AdamW. For both SST-2 and  QNLI, the AdamW trained model is much worse at skill localization through grafting. However, a small $\ell_1$ regularization on the parameter movement during FT recovers localization.\vspace{-0.2in}}
    \label{fig:adam_sgd}
\end{figure}

%% file: OODCalibration.tex
\ifthenelse{\boolean{arxiv}}{}{
\vspace{-0.1in}
}
\section{Calibration, OOD and Generalization}
\label{sec:patch_application}

Usually ``skill'' denotes flexibility and competence at a task, and  machine learning has  standard notions for testing this.
\ifthenelse{\boolean{arxiv}}{}{
\vspace{-0.1in}
}
\input{Calibration.tex}

\input{OOD.tex}
\input{Generalization.tex}

%% file: Calibration.tex
\subsection{Skill Localization improves Calibration}
\label{sec:calibration}
\looseness-1 Human skill in a task usually includes some awareness of when the task has not been performed well.   In ML, {\em calibration}  is an attempt to formalize this. Suppose a classification model outputs, given input $x$ and each possibly label $y$, a probability $\Pr[y|x]$ that the label $y$ is correct for $x$. In order to be {\em well-calibrated}, this probability should be meaningful, i.e., among all $x, y$ such that $\Pr[y|x]=a$, the expected fraction of these where $y$ is the correct (ground truth) label should also be around $a$. It is well-known that usual softmax outputs in ML models are not well-calibrated. (This can be mitigated given enough held-out data by re-calibrating the output.) Could skill localization help with calibration? This could be interesting in low-data settings where re-calibration is impossible. 

\Cref{tab:roberta_patch_perf} reports the calibration error using the ECE metric \citep{naeini2015obtaining} (described in \Cref{sec:apx_calibration}) in a pre-trained model with sparse grafting  using tasks from GLUE dataset. Sparsity levels that cause $<5$\% reduction in accuracy lead to $40-90$\% reduction in ECE. 
Vanilla fine-tuning is highly overconfident in  wrong predictions, and grafted models avoid this; see  histograms  in \cref{fig:prediction_qnli_4096}, 

\textbf{Comparison with re-training.} Our sparse grafting involves no re-training.  Does re-training just the grafted parameters affect calibration? \Cref{tab:bitfit_graftretrain} shows that  impressive calibration persists after re-training. This suggests that the sparse grafting region identified by our method is, in some sense, fundamentally suited to the task.

Note that several recent papers have tried to sparsify fine-tuned nets by identifying sub-networks of interest and re-training their parameters -- one example is BitFit.  \Cref{tab:bitfit_graftretrain} finds that  BitFit is better calibrated than vanilla fine-tuning, but worse than our grafted model after re-training. This suggests that the sparse regions identified by our procedure are better at localizing the skill.

\input{table_fig/confidence.tex}

%% file: table_fig/confidence.tex
\begin{figure}[!htbp]
    \centering
    \begin{subfigure}{0.24\textwidth}
        \centering
        \includegraphics[width=\textwidth]{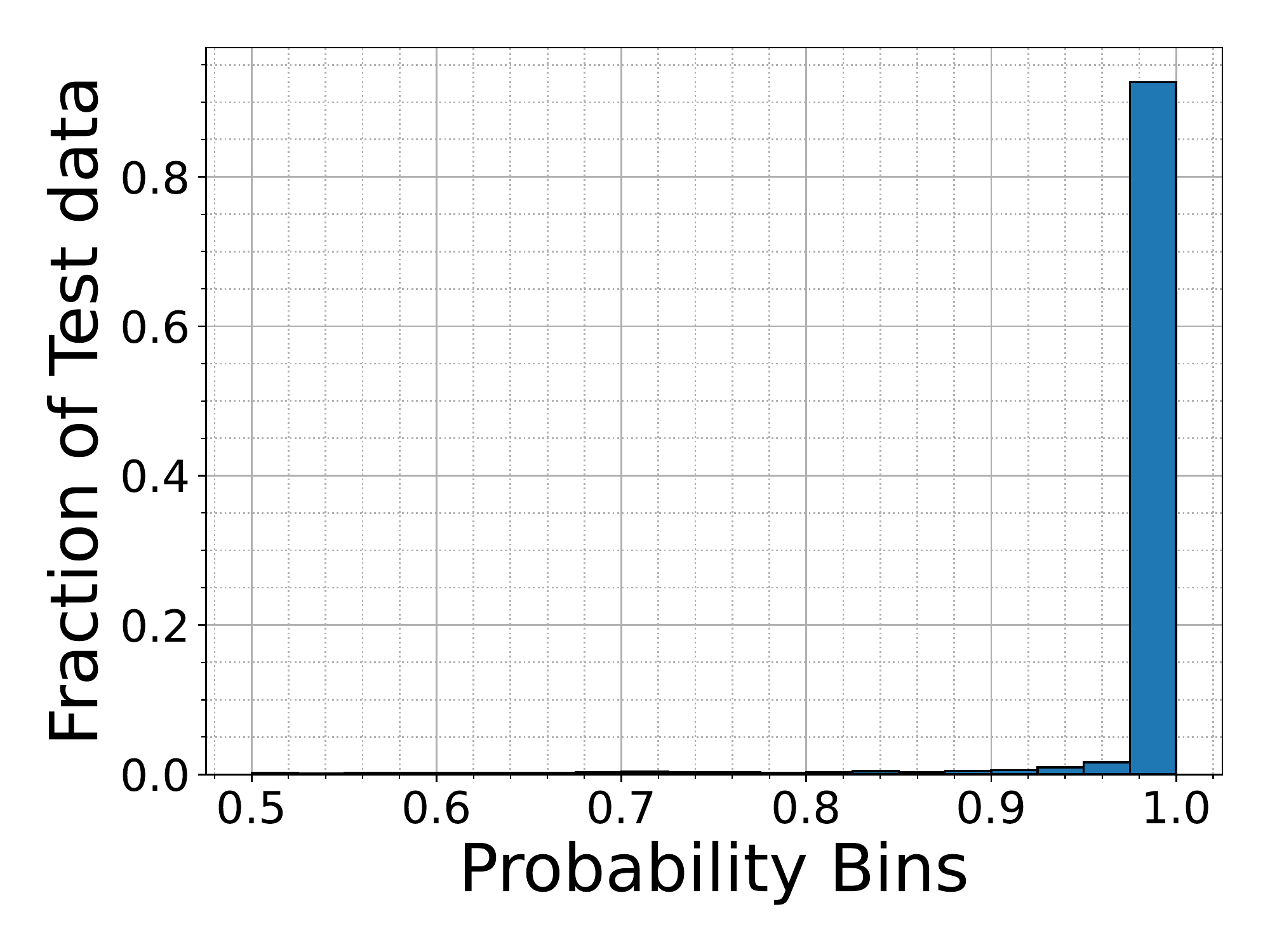}
    \end{subfigure}\hfill
    \begin{subfigure}{0.24\textwidth}
        \centering
        \includegraphics[width=\textwidth]{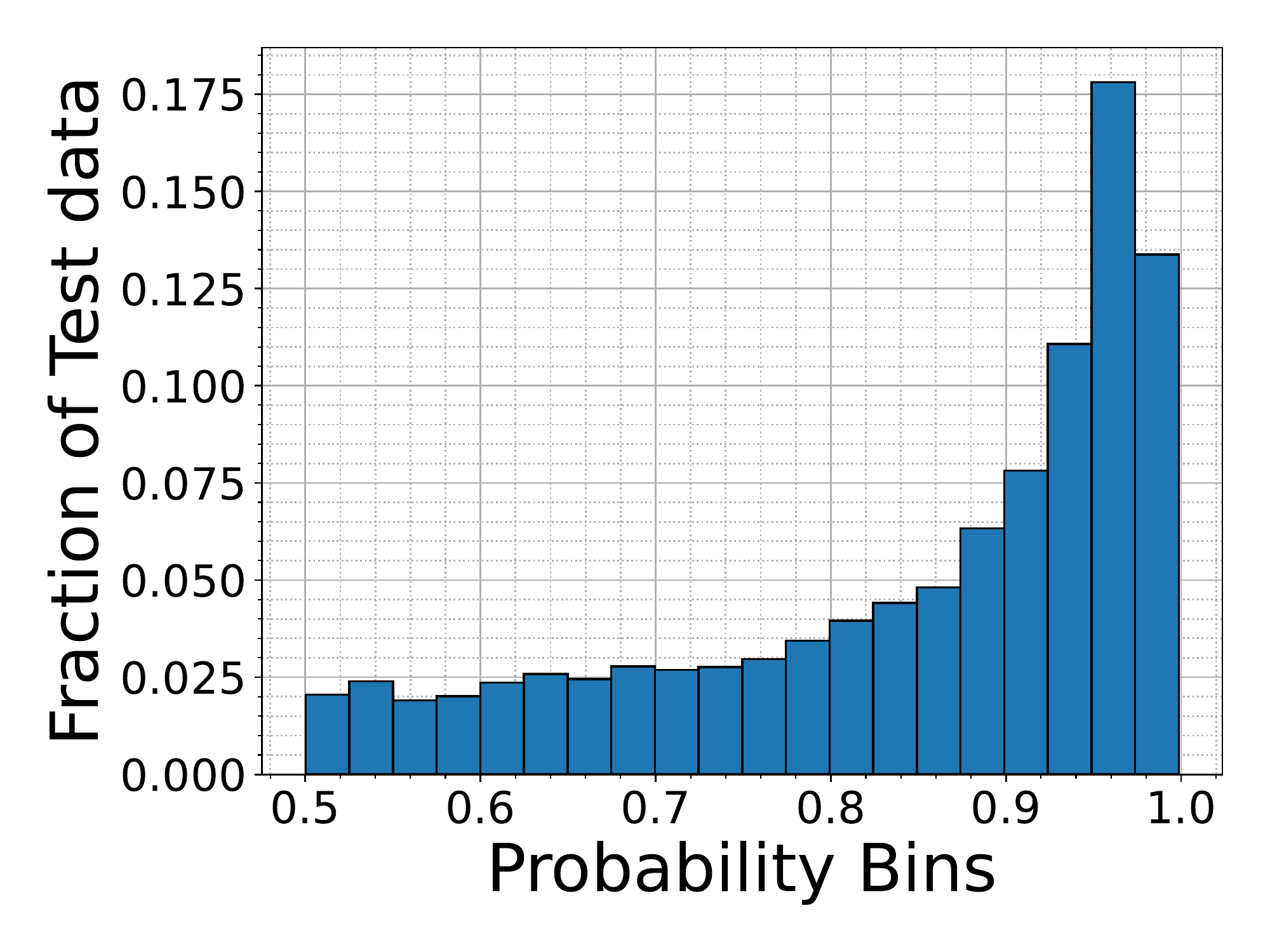}
    \end{subfigure}\hfill
    \caption{Histogram of top prediction probabilities for FT model and grafted model on QNLI (4096-shot). ({\bf left}) The model assigns high confidence on most examples. ({\bf right}) The grafted model has diverse confidence levels, explaining its superior calibration.\vspace{-0.2in}}
    \label{fig:prediction_qnli_4096}
\end{figure}

%% file: OOD.tex
\ifthenelse{\boolean{arxiv}}{}{
\vspace{-0.1in}
}
\subsection{Out-of-Distribution Generalization}
\label{sec:}

\begin{figure*}[!th]
\centering
\begin{subfigure}{0.32\textwidth}
\centering
\includegraphics[width=\textwidth]{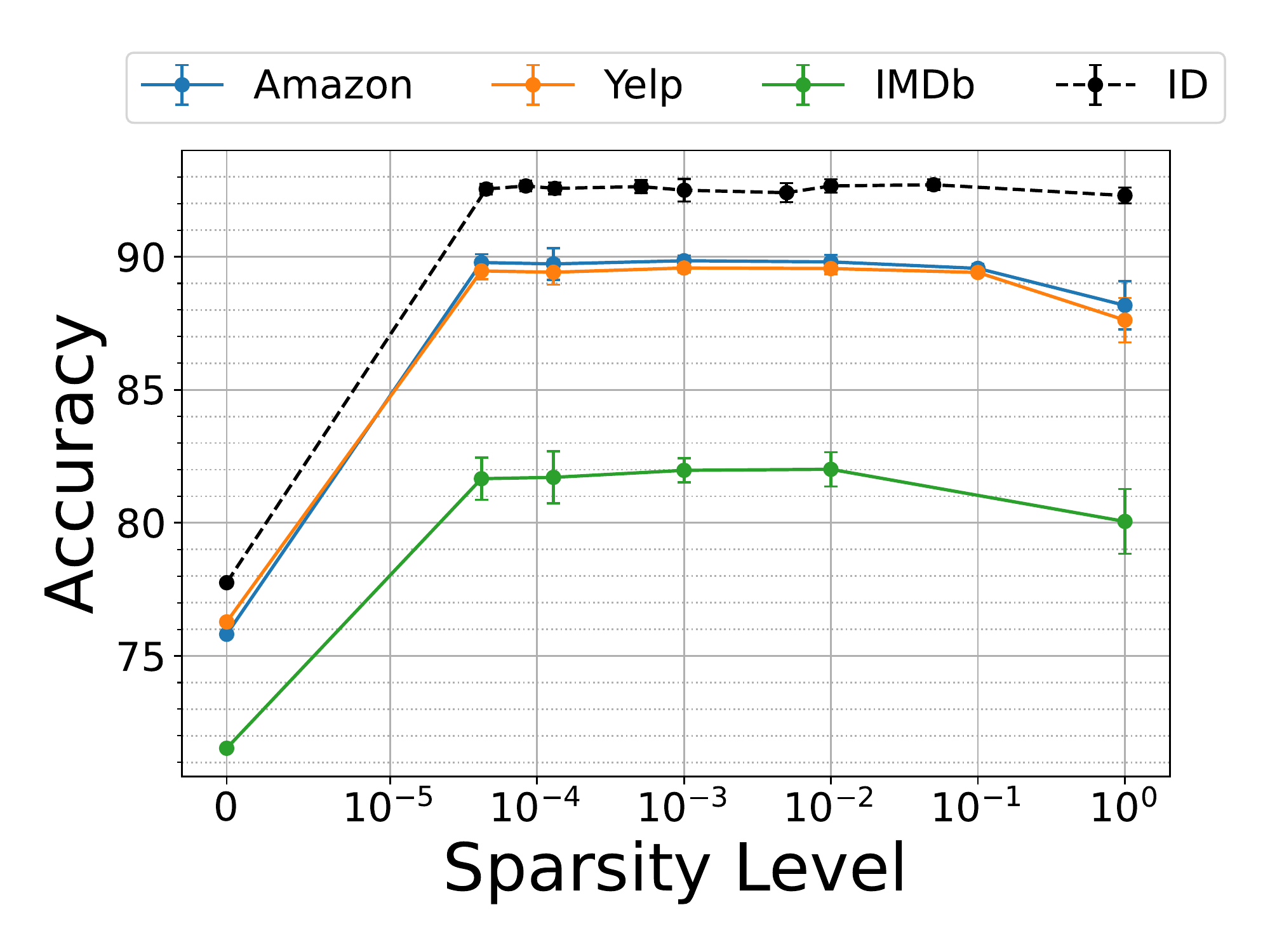}\caption{OOD performance of SST-2 FT model}
\label{fig:OOD_sst-2_4096}
\end{subfigure}\hfill
\begin{subfigure}{0.32\textwidth}
\centering
\includegraphics[width=\textwidth]{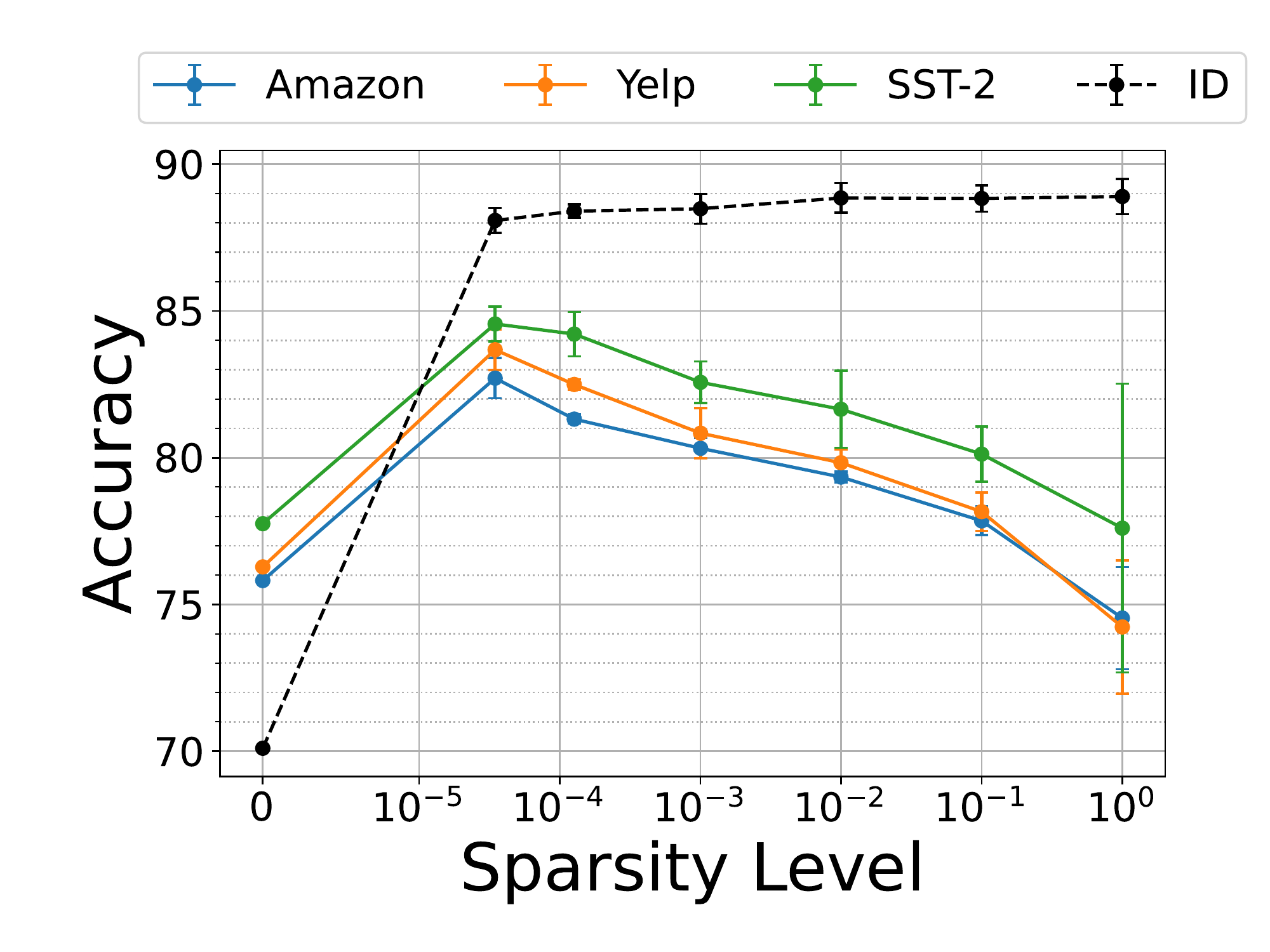} \caption{ OOD performance of MPQA FT model  }
\label{fig:OOD_mpqa_4096}
\end{subfigure}\hfill
\begin{subfigure}{0.32\textwidth}
\centering
\includegraphics[width=\textwidth]{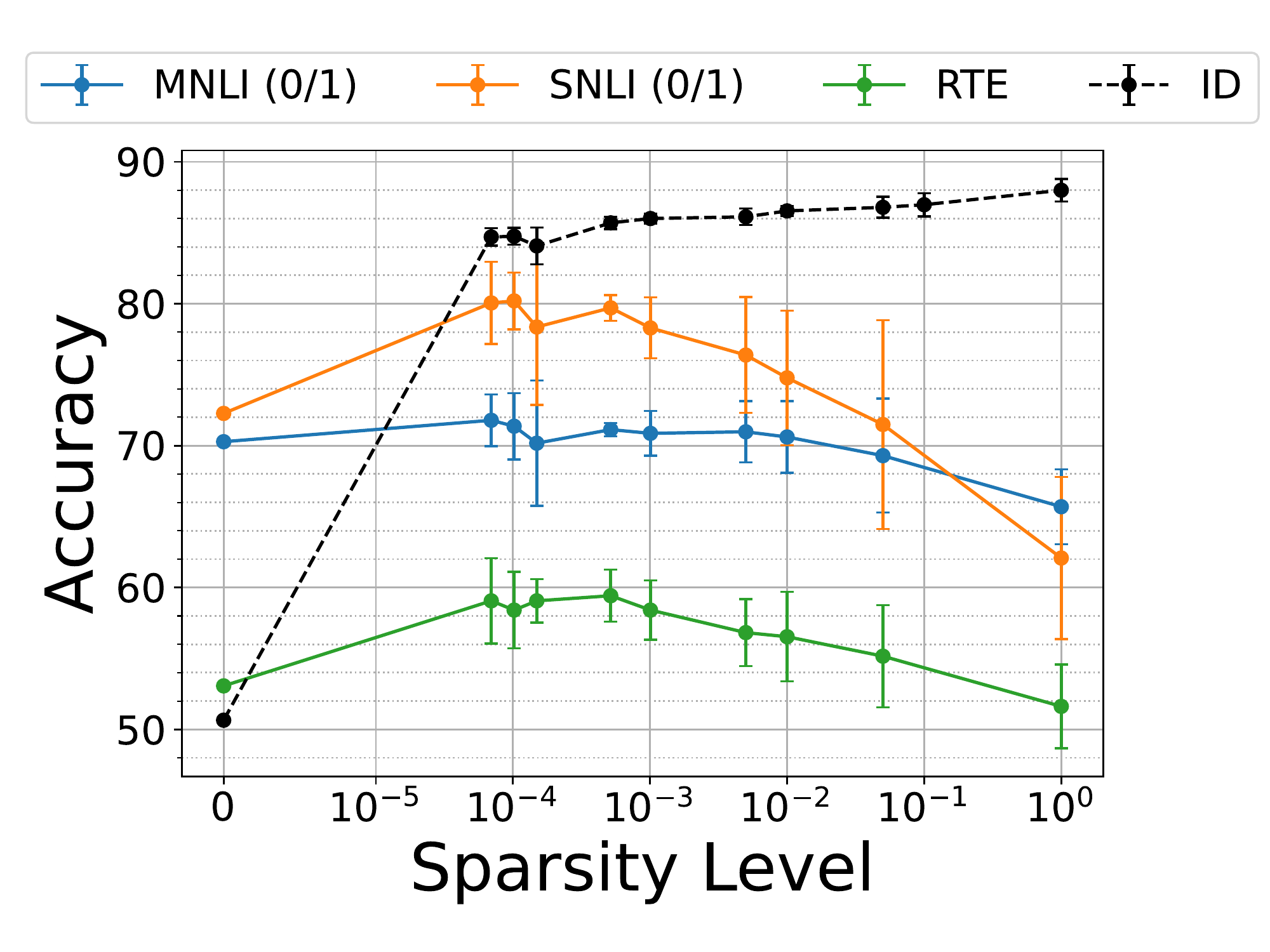} \caption{ OOD performance of QNLI FT model }
\label{fig:OOD_qnli_4096}
\end{subfigure}\hfill
\centering
\begin{subfigure}{0.32\textwidth}
\label{tab:OOD_expts_wisepatch_qnli}
\centering
\includegraphics[width=\textwidth]{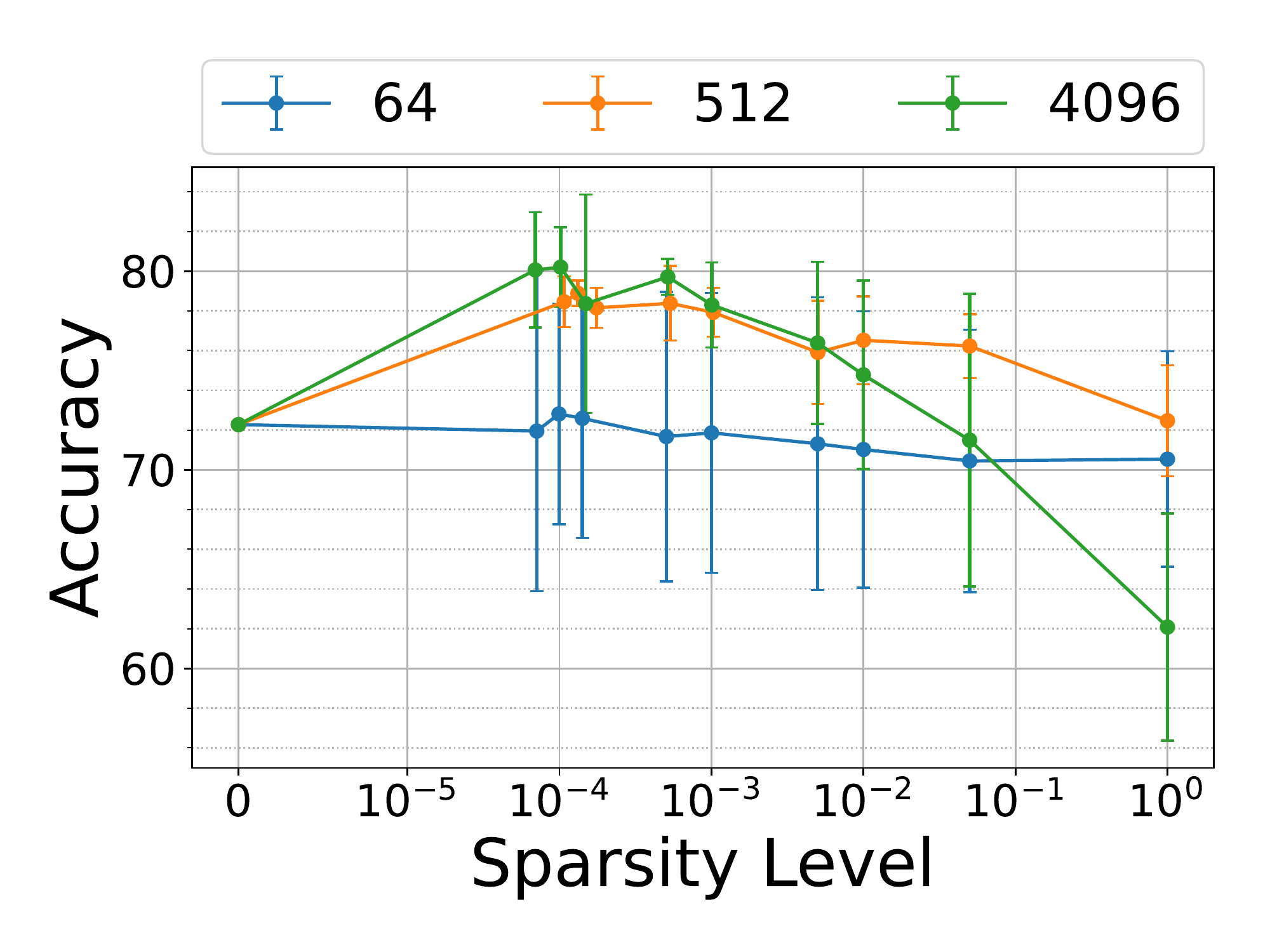}\caption{QNLI $\rightarrow$ SNLI (0/1) for varying $k$-shot}
\label{fig:OOD_qnli_multishot}
\end{subfigure}\hfill
\centering\hfill
\begin{subfigure}{0.32\textwidth}
\centering
\includegraphics[width=\textwidth]{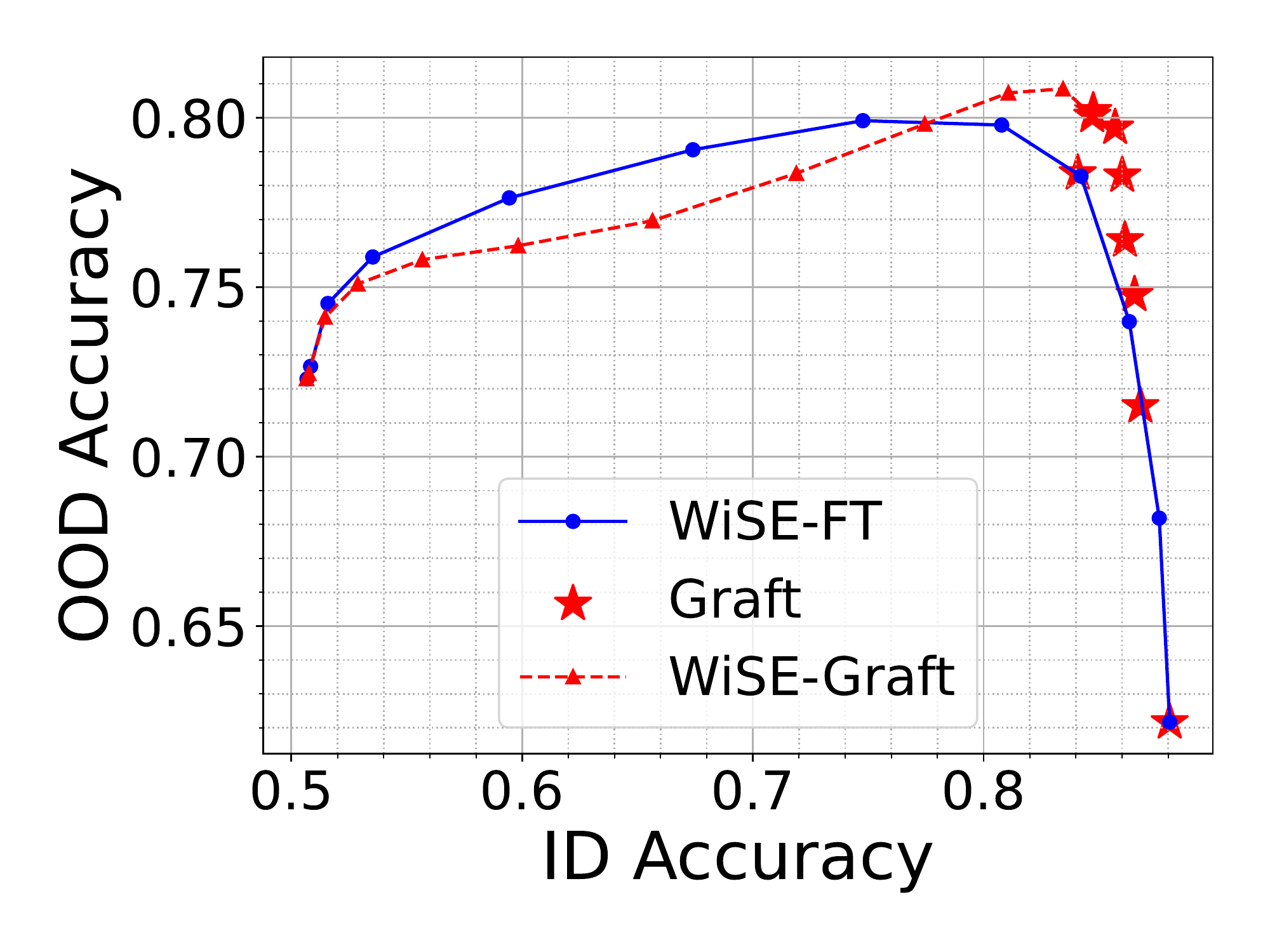}\caption{ID-OOD curves for QNLI $\rightarrow$ SNLI (0/1)}
\label{fig:Wise_patch_snli_qnli}
\end{subfigure}\hfill
\centering
\begin{subfigure}{0.32\textwidth}
\label{tab:OOD_expts_wisepatch_mpqa}
\centering
\includegraphics[width=\textwidth]{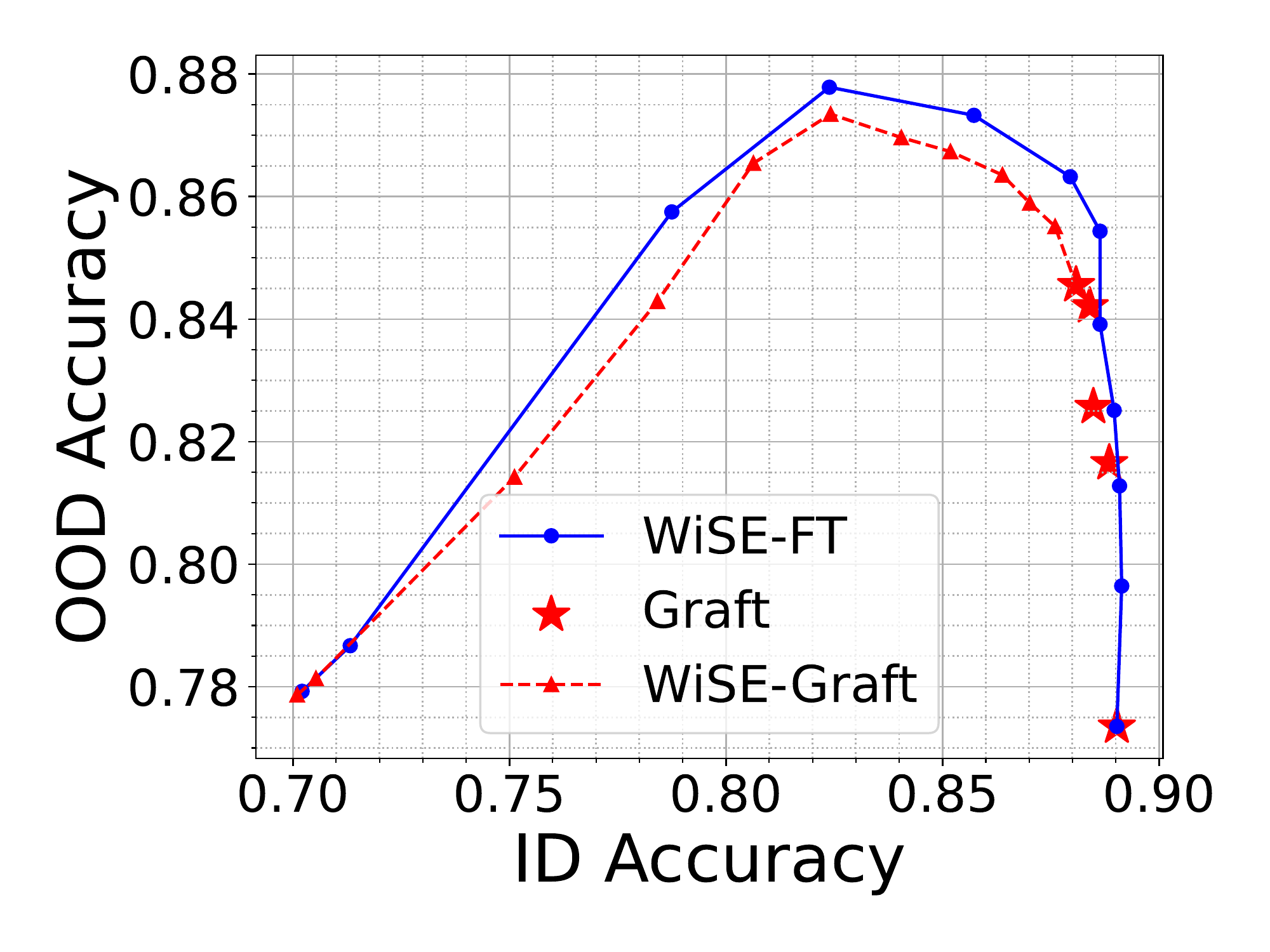}\caption{ ID-OOD curves for MPQA $\rightarrow$ SST-2 }
\label{fig:Wise_patch_mpqa_sst}
\end{subfigure}\hfill
\caption{Comparing the zero-shot OOD performance of the FT model and grafting in various settings. (b,c) We observe at least  a $5\%$ gap between the performance of the two, when the distribution shifts are large. (c) The gap gets worse as the number of available in-distribution samples increases. (e) For transfer in NLI task, the optimal (ID, OOD) point for WiSE-FT is (80.5, 79.6), and for grafting (84.2, 80.0). (f) For transfer in sentiment task, WiSE-FT on sparse grafted models (WiSE-Graft) gets a competitive ID-OOD curve. \ifthenelse{\boolean{arxiv}}{}{
\vspace{-0.1in}
} }
\label{fig:OOD_expts_4096-shot}
\end{figure*}

\looseness-1 Human skills extend at least a bit to new settings; e.g., skill at throwing a baseball should also lead to ability to throw a tennis ball. 
We evaluated in-distribution (ID) and out-of-distribution (OOD) accuracies for grafted models of varying sparsities  in \cref{fig:OOD_expts_4096-shot}. 
We find that grafted models may suffer a little on ID performance but match or significantly outperform vanilla fine-tuning  on OOD. This suggests that grafting has indeed captured ``core'' skills.

\looseness-1\textbf{Small distribution shifts.} When the distribution shift between tasks is intuitively small (e.g. SST-2 to IMDb or MNLI to SNLI) vanilla fine-tuned model itself is robust enough ---grafting provides little or no advantage. Similar  findings appear in \citet{hendrycks2020pretrained}.

\looseness-1\textbf{Large distribution shifts.} Sentiment analysis task MPQA uses text from news articles while SST-2/Yelp/Amazon uses  reviews.
We find that models fine-tuned on MPQA perform poorly when tested on SST-2 and Yelp. However, the grafted model for MPQA performs at least $5\%$ better than the fine-tuned model.
Similar results hold for NLI datasets. QNLI task consists of question/answer pairs whereas MNLI and SNLI\footnote{We consider only contradiction and entailment labels here.} have pairs of assertions as inputs. This distribution shift is enough to make vanilla fine-tuning on QNLI perform poorly on MNLI and SNLI, but the grafted model for QNLI again performs around $5\%$ better. 

\looseness-1\textbf{Comparison with WiSE-FT.} Often there is no magic bullet for doing well on both ID and OOD generalization.  For image data (with model pre-trained on CLIP), it was shown that WiSE-FT \citep{wortsman2022robust}, which linearly interpolates between $\paramft$ and $\parampre$, does best in the ID-OOD trade-off. 
\Cref{fig:OOD_expts_4096-shot} explores similar ideas for NLP tasks.
Model grafting is better than WiSE-FT for one ID-OOD pair, but the opposite is true for a different pair.
Applying the WiSE-FT idea on the grafted model (i.e. interpolating between grafted model and pre-trained model),  "WiSE Graft," gets competitive ID-OOD tradeoff to Wise FT.

\textbf{Comparison with parameter-efficient methods.} To investigate if the number of updated parameters is the key factor behind the better performance of sparse grafted models in terms of calibration and OOD generalization, we assess the calibration error and OOD accuracy of few parameter-efficient fine-tuning methods. We find that, at comparable sparsity levels, grafting exhibits better calibration and OOD performance. This suggests that the sparsity of the grafting region is not the only factor for its better OOD and calibration performance.  Furthermore, the OOD and calibration performance of grafting change gradually with decreasing sparsity.
Please see \cref{sec:peft} for more details.

\input{table_fig/OOD_retrain_stitch_mainpaper.tex}

%% file: table_fig/OOD_retrain_stitch_mainpaper.tex
\begin{table}[!t]
\centering
\caption{
Comparing the OOD performance of model grafting, graft re-training and BitFit. For NLI tasks, the OOD accuracy of graft re-training is $5\%$ worse than model grafting. BitFit completely fails on OOD tasks when the distribution shift is large.\vspace{0.1in}}
\label{tab:ood_retrain_stitch}
\small
\npdecimalsign{.}
\nprounddigits{1}
\begin{tabular}{l|ccc} \toprule
  \midrule
 OOD task & Grafting & Graft re-training & BitFit \\
 \midrule
 \multicolumn{4}{c}{ID task: SST-2} \\
 \midrule
 Yelp & 89.5 {\tiny (0.3)} & 88.9 {\tiny (1.0)} & 89.0 {\tiny (0.3)}\\
IMDb & 81.5 {\tiny (0.7)} & 81.2 {\tiny (1.4)} & 81.3 {\tiny (0.7)}\\
\midrule
    \multicolumn{4}{c}{ID task: QNLI} \\
\midrule
 MNLI(0/1) & 71.8 {\tiny (1.8)} & 67.0 {\tiny (2.3)} & 60.5 {\tiny (4.9)}\\
SNLI(0/1) & 80.1 {\tiny (2.9)} & 66.4 {\tiny (7.1)} & 57.4 {\tiny (8.0)}\\
 \bottomrule 
\end{tabular}
\npnoround
\vspace{-0.1in}
\end{table}

%% file: Generalization.tex
\subsection{Understanding Generalization for Fine-tuning}
\label{sec:generalization}
\ifthenelse{\boolean{arxiv}}{}{
\vspace{-0.05in}
}
\looseness-1Fine-tuning a vast pre-trained model on a small dataset seems iffy since we are finding the best model in a very large class of models $\paramclass$, and according to the classic understanding of generalization, the error could  be as high as $\sup_{\param\in\paramclass} \left|\testloss(\param) - \trainloss(\param) \right|$. This bound is too pessimistic for most deep learning settings \citep{nagarajan2019uniform}, including fine-tuning, since the training data can be easily fit perfectly in these settings.

\looseness-1Understanding generalization of the grafted model is more tractable because of the small size of 
the grafting region. Empirically we find that re-training on the grafted parameters fails to make  $\left|\testloss(\param) - \trainloss(\param) \right|$ higher than $1\%$ once the dataset has a few thousand datapoints, which can be formalized as a ``complexity parameter''. \Cref{app:generalization} explores this further using classical generalization theory.

%% file: MTContinual.tex
\ifthenelse{\boolean{arxiv}}{}{
\vspace{-0.1in}
}
\section{Multi-Task and Continual Learning}
\label{sec:MT_Continual}

\looseness-1Previous sections have shown that sparse grafts can localize skills when fine-tuning on a single task.
In this section, we test a stronger version of skill localization involving multiple tasks in the following settings: (i) the model is fine-tuned on many tasks together, (ii) continual learning, where the model is fine-tuned on one task at a time.

\input{table_fig/MT_exps.tex}

\ifthenelse{\boolean{arxiv}}{}{
\vspace{-0.1in}
}
\subsection{Multi-Task Learning}
\label{sec:MT}

\looseness-1We perform multi-task learning (MT) by fine-tuning a RoBERTa-base model with SGD on 8 different datasets ($4096$-shot setting for each) simultaneously. The datasets represent four different classes of tasks: NLI, sentiment analysis, paraphrasing, and classification.
Firstly, the resulting MT model achieves test accuracy comparable with the task-specific FT models, suggesting no {\em gradient interference} that is observed in some cases \citep{yu2020gradient}.
For skill localization, we learn task-specific sparse regions: for each task $i$, we optimize for $\loss_{i}$ (i.e., performance on task $i$) from \Cref{eq:opt_patch} using the MT model parameters as $\paramft$ and $\basepatch=\bm{0}$. (Note that grafted models for $i, j$ have the same value for parameters that are contained in both $\patch_i$ and $ \patch_j$.) Results are presented in \cref{tab:MT_expts_4096-shot_8tasks_mainpaper}.

\looseness-1We find skill localization continues to exist in multi-task models, and
 now also provides signal about task similarity (through region overlap) and affords interesting compositional properties (through the union of regions).

\looseness-1\textbf{Region overlap and task similarity.} \Cref{fig:overlap_patch_mt} shows that the patches for different tasks have very little overlap (defined as $\frac{|\patch_i \cap \patch_j|}{|\patch_j|}$ for tasks $i, j$).  However, similar tasks show slightly more overlap compared to other pairs, e.g. (SST-2, CR), and (SNLI, MNLI).
This is some evidence of skill-sharing across  similar tasks.

\looseness-1\textbf{Skill isolation and transfer.} In \Cref{fig:ind_patch_mt} we find that grafted models for a single task, which presumably isolate the skills for just that task, indeed only help that particular task and a few closely related tasks.
We measure the effect of task $i$ on task $j$ by grafting only the parameters in the region $\gamma_i$, and measuring the performance gain on task $j$ compared to the performance gain of the MT model.
For a task $t$, if $P_{ \patch, t }$ is the accuracy of the model grafted with $\patch$, $P_{\bm{0}, t}$ is the pre-trained accuracy, and $P_{\bm{1}, t}$ is the MT model accuracy, then relative performance gain of grafting region $\patch$ is
\begin{align}
    \mathrm{Rel}_{\patch, t} = 
    (P_{\patch, t} - P_{\bm{0}, t}) / (P_{\bm{1}, t} - P_{\bm{0}, t})
    \label{eq:rel_perf_gain}
\end{align}
\looseness-1We find that some similar pairs of tasks like (SST-2, CR), and (SNLI, MNLI) show transfer, i.e. grafting the region from one task helps with the other. Interestingly, for some tasks that are seemingly similar (e.g. QQP and MRPC) the effect seems to be asymmetric, i.e. $\patch_{\mathrm{MRPC}}$ helps with QQP, but $\patch_{\mathrm{QQP}}$ does not help with MRPC. Furthermore, we observe that, $\gamma_{\mathrm{QNLI}}$ helps with QQP paraphrasing, presumably because they both have questions in their inputs.

\looseness-1\textbf{Skill compositionality through region unions: } Since grafting for a single task  works for that task as well as related tasks, we ask a more ambitious question: 
\textit{Can grafting multiple regions lead to skill isolation for that subset of tasks?}
A priori one would guess ``No,''  because the grafting regions were independently trained on individual tasks without any compositionality requirements.
Surprisingly we find the answer is a qualified ``yes.''
\Cref{fig:union_patch_zeroshot} presents compositionality results for 5 groups of tasks.
For each group $G$, we take the union of regions $\patch_{G} = \cup_{i\in G} \gamma_i$ and evaluate the relative performance gains for tasks using the grafted model of $\patch_{G}$.
We indeed find that composing tasks for a subset in this way retains around $70\%$ 
of the accuracy gains for that subset and related tasks, but not for other tasks.
We tried slightly fine-tuning the union region $\patch_{G}$ to optimize the joint loss on tasks in $G$, i.e. $\sum_{i\in G} \loss_{i}(\patch)$, by only taking 10 gradient steps for quick adaptation.
\Cref{fig:union_patch_10steps} shows that this causes the accuracy gain on relevant tasks to be even higher (around $\sim 80$\%) without affecting gains for other tasks by much.
The emergence of this compositionality property, we believe, is very interesting and deserves further exploration.

\ifthenelse{\boolean{arxiv}}{
\looseness-1\textbf{AdamW MT training: } We also check skill localization in a model trained with AdamW (see \cref{tab:MT_expts_4096-shot_8tasks_1e-7_Adam} in the Appendix).
Interestingly, we find that the task-specific grafts show small overlap across tasks, and only perform well on similar tasks, indicating localization even in the absence of an explicit $\ell_1$ regularization during training.
This is in stark contrast to single task trained models, which had failed to show any skill localization without an explicit $\ell_1$ regularization (\cref{fig:adam_sgd_qnli,fig:adam_sgd_sst-2}).
We speculate that forcing the model to do well on multiple tasks together naturally encourages the model to localize skills.
}{}

\ifthenelse{\boolean{arxiv}}{}{
\vspace{-0.1in}
}

\subsection{Forget-free Continual Learning}
\label{sec:continual}

\looseness-1Continual learning aims to train a model sequentially, seeing one task at a time.
A frequent complication is {\em catastrophic forgetting} (see chapter 4 in \citet{chen2018lifelong}): training on new tasks can greatly hurt performance on earlier tasks. 
Here skill localization could help: once skills for previous tasks have been localized, we can freeze them and only update the rest of the net for new tasks. 
We use this localization idea, through our grafting procedure, to perform forget-free continual learning, i.e. without forgetting anything about previous tasks.

\input{table_fig/continual.tex}

\looseness-1The main idea is to only train the parameters in the grafting region for a task, that does not intersect with the grafting regions of the previously encountered tasks.
During inference, inspired by \citep{kang2022forget}, we only use the grafted model that takes the union of the grafting regions of the task and the previous tasks.
While this requires resetting parameters to pre-trained values for each evaluation, the total memory needed to retain all skills is propositional to $T s$ instead of $T d$, where $T$ is the total number of tasks, $s$ ($\sim 5000$) is the sparsity of the grafting regions and $d$ is the total number of parameters in the model ($\sim 100$M).
Preliminary experiments in \Cref{tab:continual} on a sequence of three tasks suggest a significant benefit of skill localization.
\Cref{sec:apx_continual} provides more details on training and evaluation procedures, and includes a discussion on why grafting helps in this case.
Other explorations of skill localization and grafting for continual learning is left for future work.

%% file: table_fig/MT_exps.tex
\begin{figure*}[!th]
\centering
\begin{subfigure}{0.4\textwidth}
\centering
\includegraphics[width=\textwidth]{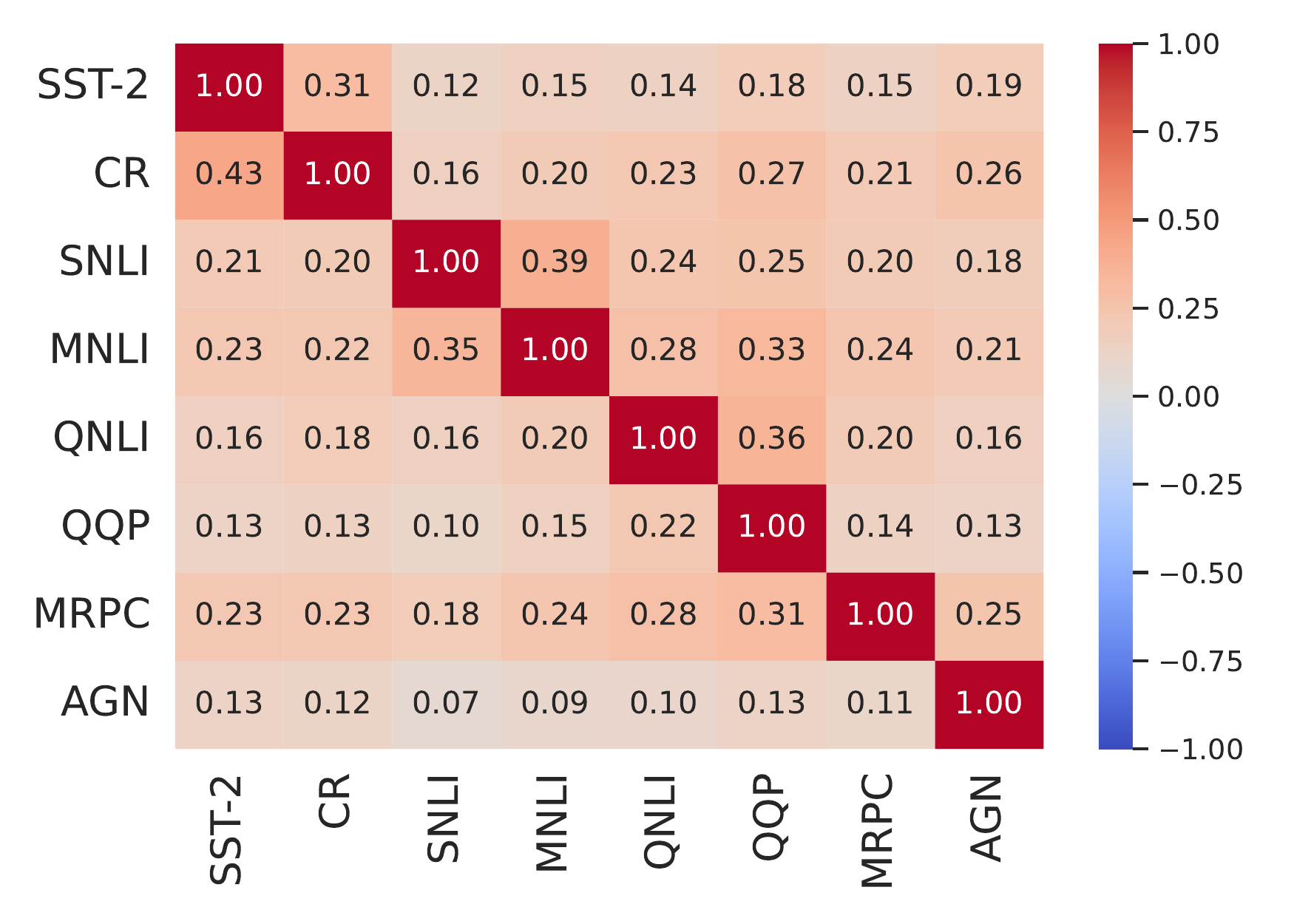}\caption{Overlap of task-specific grafting regions}
\label{fig:overlap_patch_mt}
\end{subfigure}
\begin{subfigure}{0.4\textwidth}
\centering
\includegraphics[width=\textwidth]{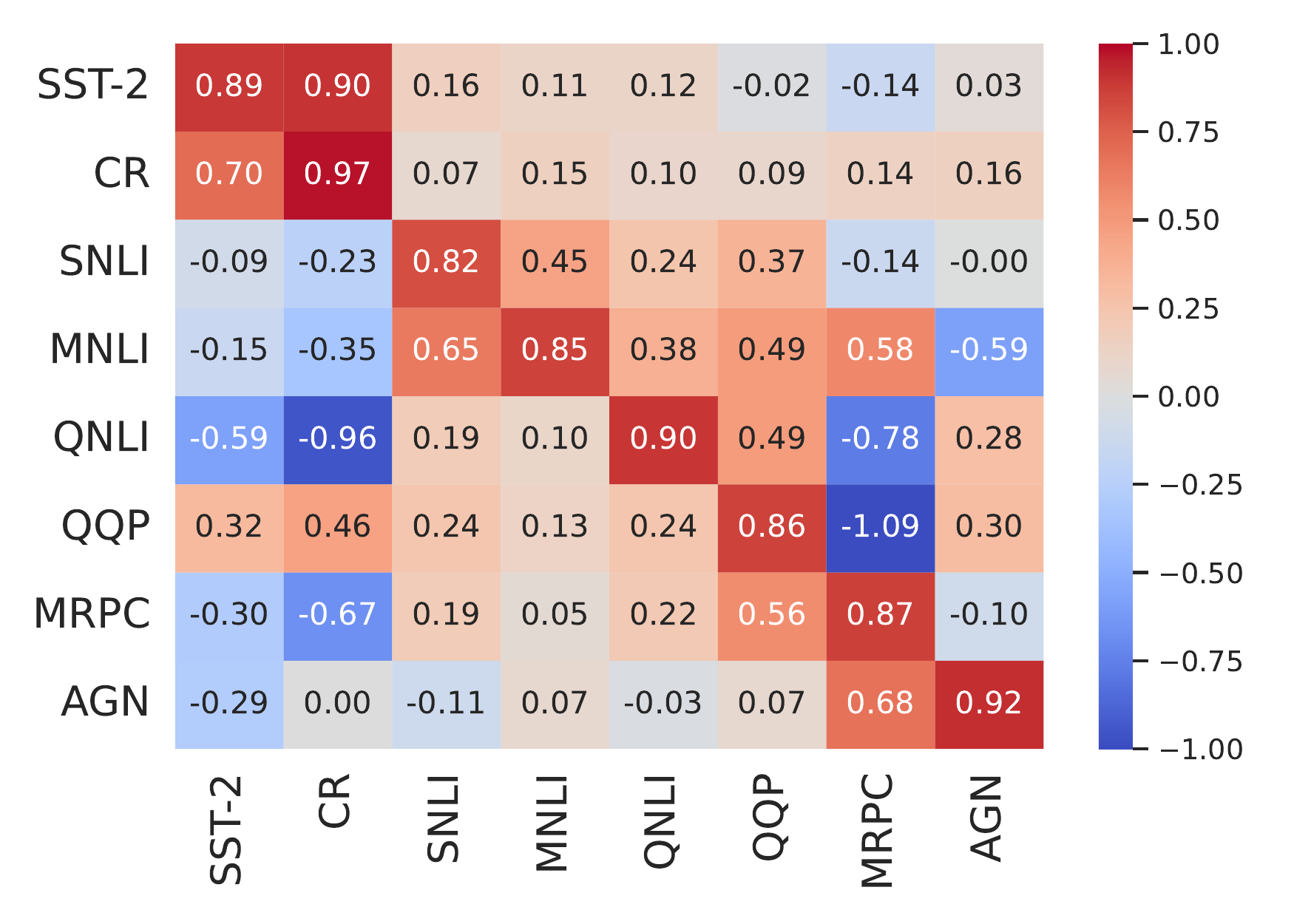}\caption{ Effect of task-specific grafted model on other tasks} \label{fig:ind_patch_mt}
\end{subfigure}
\hfill
\begin{subfigure}{0.43\textwidth}
\centering
\includegraphics[width=\textwidth]{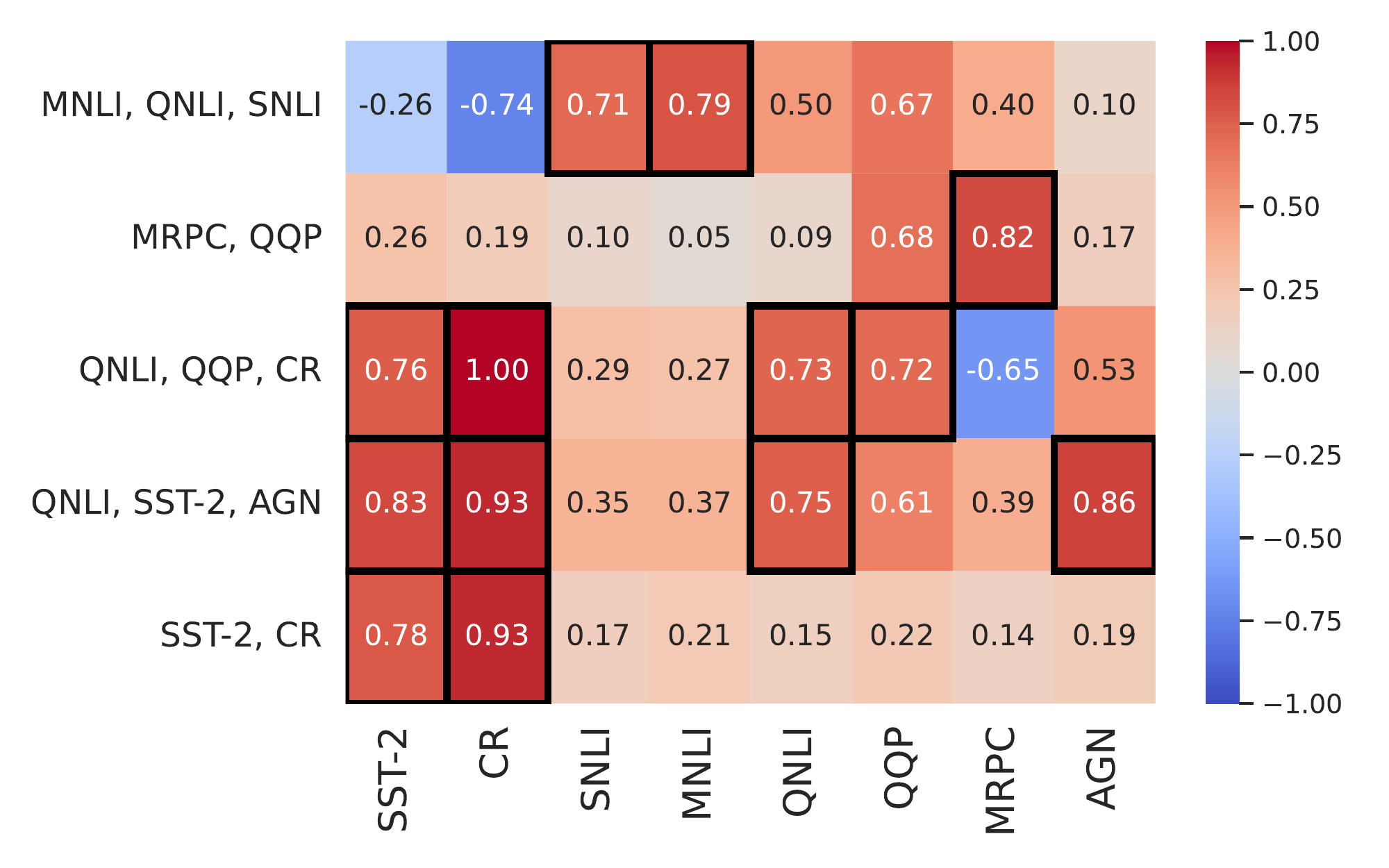} \caption{Effect of union grafting}
\label{fig:union_patch_zeroshot}
\end{subfigure}
\begin{subfigure}{0.43\textwidth}
\centering
\includegraphics[width=\textwidth]{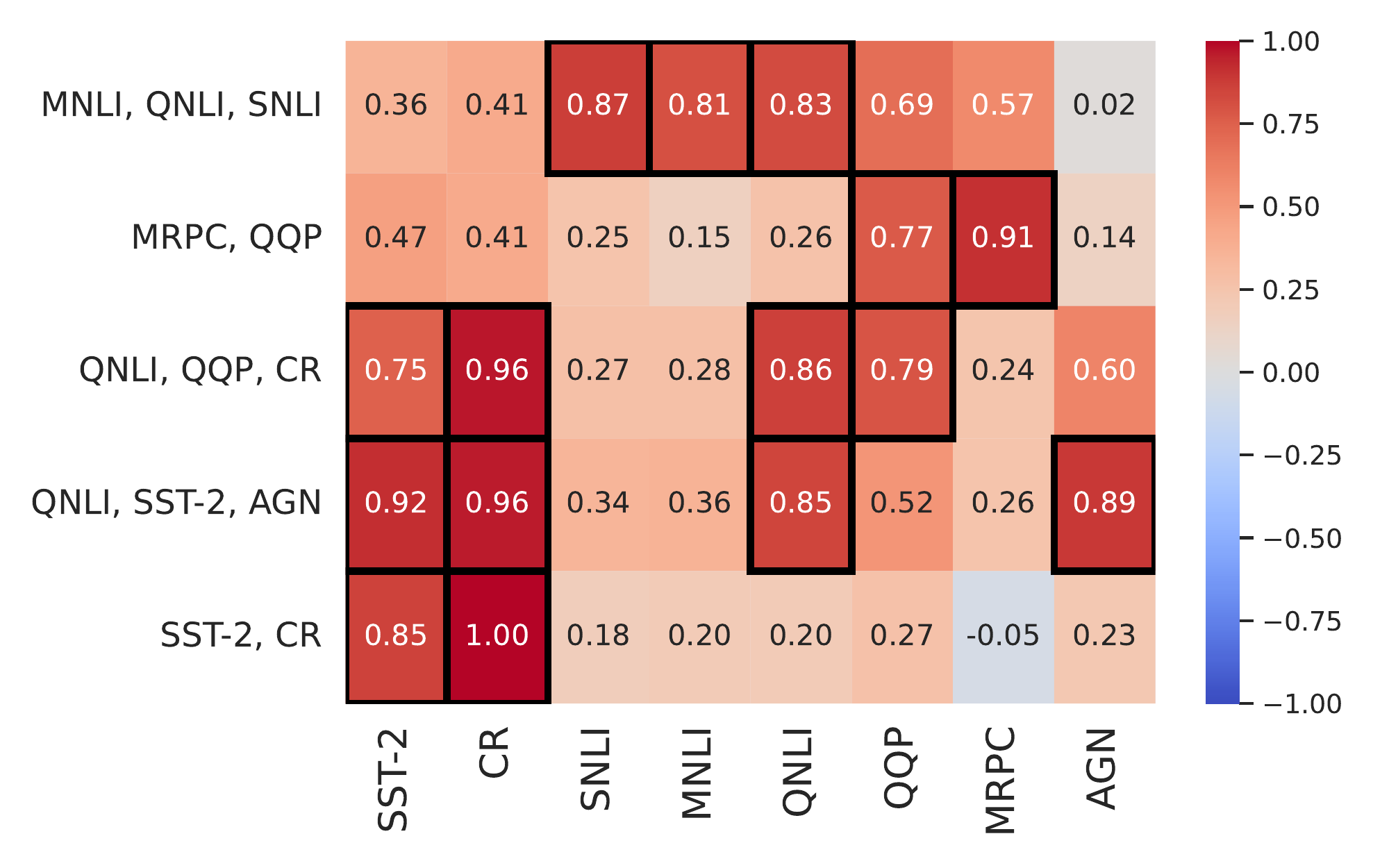}\caption{Effect of union grafting + purification}
\label{fig:union_patch_10steps}
\end{subfigure}
\caption{
Ablations on task-specific grafting region $\patch_{i}$ for task $i$, learned by optimizing $\loss_{i}$ on the MT model. \Cref{sec:MT} has details of experiments and the metrics being reported. 
\ifthenelse{\boolean{arxiv}}{
In all figures, we evaluate the effect of graft region of task in row $i$ on the task in row $j$.
Figure (a) measures the assymteric overlap in the regions defined as $\frac{|\patch_i \cap \patch_j|}{|\patch_j|}$ for tasks in row $i$ and column $j$.
Figure (b), (c), (d) evaluate the relative accuracy gain of task in column $j$ using the graft regions of task(s) in row $i$; refer to \Cref{eq:rel_perf_gain} for the precise expression.
}{}
\textbf{Observations:}   
(a) Similar tasks, like (SST-2, CR), and (SNLI, MNLI), show relatively higher overlap in grafting regions.
(b) The grafted model of a task only does well on itself and a few similar tasks.
(c) The grafted model with union of regions for a subset $G$ of tasks only does well on the tasks in $G$ and similar tasks (all values higher than $0.7$).
(d) Allowing a few steps of GD to purify the union of grafting regions improves the grafted model's performance on the desired set of tasks. \vspace{-0.1in}}
\label{tab:MT_expts_4096-shot_8tasks_mainpaper}
\end{figure*}

%% file: table_fig/continual.tex
\begin{table}[!t]
\centering
\caption{Continual learning on the sequence of tasks QNLI, AG news, SST-2. The naive continual FT leads to a $20\%$ drop in accuracy for QNLI, owing to catastrophic forgetting. Grafting continual FT (our procedure) can retain the performance on QNLI, while minimally affecting the performance of newer tasks. \vspace{0.1in}}
\label{tab:continual}
\footnotesize
\begin{tabular}{l|c|c|c} \toprule
  \midrule
 Method & QNLI & AGNews & SST-2 \\
 \midrule
 FT & 88.0 {\tiny (0.8)} & 93.1 {\tiny (0.1)} & 92.1 {\tiny (0.1)}\\
ContinualFT & 67.5 {\tiny (5.3)} & 87.6 {\tiny (2.5)} & 92.0 {\tiny (0.5)}\\
GraftingContinual & 86.5 {\tiny (0.7)} & 90.8 {\tiny (0.1)} & 92.5 {\tiny (0.3)}\\
 \bottomrule
\end{tabular}
 \vspace{-0.2in}
\end{table}

%% file: RelatedWorks.tex
\ifthenelse{\boolean{arxiv}}{}{
\vspace{-0.1in}
}
\section{Related Work}
\label{sec:related_work}

\looseness-1\noindent{\bf Knowledge/skills.} 
 \citet{li2022large} show that the feed-forward activations are sparse in large pre-trained models. \citet{dai2021knowledge} discover knowledge neurons in BERT, whose activations correlate with specific facts, whereas \citet{burns2022discovering} find latent knowledge in the internal representations of language models. Furthermore, \citet{meng2022locating,hase2023does} show that the language models localize knowledge in the feed-forward layers of the pre-trained models.
\citet{wang2022finding} find specific ``skill'' neurons highly predictive of the downstream task in soft prompt-tuning \citep{li2021prefix} of language models. \citet{xie-etal-2022-hidden} demonstrate that a layerwise analysis of hidden state variations in pre-trained language models can aid in the identification of an optimal subset of task-specific layers for fine-tuning purposes.

\looseness-1\noindent{\bf Parameter-efficient fine-tuning (PEFT).} 
The goal is to update very few parameters for efficient training.
\citet{houlsby2019parameter} only train a small set of trainable parameters added between layers of the network. \citet{gordon2020compressing} use an $\ell_0$ regularizer to update few layers during fine-tuning.
BitFit \citep{ben2022bitfit} only updates biases during FT, and performs comparably to vanilla FT. (Our graft regions have fewer parameters.) 
Grafting only retains updates for few parameters, but is not motivated by efficiency considerations. Leveraging grafting for better PEFT could be an interesting future direction.

\looseness-1\noindent{\bf Lottery ticket hypothesis} ~\citep{frankle2018lottery} asserts that a trained neural network can be re-trained using a small sub-network while setting other parameters to 0 and still reach the same performance. Lottery tickets for pre-trained language models are studied in~\citep{chen2020lottery,prasanna2020bert,liang-etal-2021-super}.
To our best knowledge, LTH results in sub-networks much denser than graft.

While \citet{gong-etal-2022-finding} claim to find a much sparser ``lottery ticket'' that transfers to different GLUE tasks, to the best of our understanding, they set parameters outside the ticket to the pre-trained value and not $0$. Their aim is to design a PEFT technique, whereas grafting aims to understand the mechanism of fine-tuning and finding {\em task-specific} regions in fine-tuned models. Since grafting is post-hoc, i.e. involves no further fine-tuning, it provides a lens for a more holistic evaluation of fine-tuning.

\looseness-1\noindent{\bf OOD generalization and distribution shifts.} 
\citet{diffenderfer2021winning,zhang2021can,liu2022win} show that re-trained lottery tickets can be more robust to the distribution shift. \citet{lee2022surgical} alleviate  distribution shift by only fine-tuning specific layers of the whole model. 
\citet{li2022large} show the efficacy of sparsifying the activations of feed-forward layers in the models.

\noindent{\bf Multi-task training.}
 \citet{misra2016cross} stitch networks from different tasks together, \citet{long2017learning} learn relations between tasks to enhance performance, and \citet{lu2017fully} apply adaptive weight-sharing between the networks for different tasks. Multi-task training in NLP is also studied in \citep{collobert2008unified,liu2016recurrent,gupta-etal-2016-table,lu2017fully,liu-etal-2019-multi}. These involve training additional task-specific parameters whereas our approach does not.
 Also, we attempt to understand the localization of skills within the model post-hoc.

\looseness-1\citet{saunshi2021a,wei2021pretrained,malladi2022kernel} mathematically study head-tuning, prompt-tuning and fine-tuning of language models for few-shot downstream tasks.

%% file: Conclusions.tex
\ifthenelse{\boolean{arxiv}}{}{
\vspace{-0.1in}
}
\section{Conclusions and future directions}
\label{sec:conclusions}

By successfully demonstrating the ability to do a sparse``graft'' of the skill on top of the pre-trained model, this paper makes a start on localizing newly acquired skills inside fine-tuned language models.
We hope our first-cut method will improve with further work, potentially yielding better understanding as well as applications in multi-task and continual learning, which we also begin to address in Section~\ref{sec:continual}. We hope these may yield new insights on how to compose skills, decompose the identified skill into finer skills, and give  applications to unlearning and federated learning.
One open problem for multi-task setting is a method to find, for any subset $S \subseteq \{1, \ldots, m\}$ of tasks, a model that does well on all tasks in $S$. (The naive method would train models for all $2^m$ subsets of tasks.) Our approach with finding task-specific regions and using their unions shows promise for small $m$.

\looseness-1\textbf{Acknowledgments.}
We thank: Danqi Chen for feedback on an earlier draft; Saurabh Garg for pointer to WiSE-FT;  Tianyu Gao and Mengzhou Xia for pointers on the prompt-based fine-tuning codebase; the anonymous reviewers for their helpful comments. This work is supported by funding from NSF, ONR, Simons Foundation, DARPA and SRC.



%% file: Appendix.tex
\newpage
\input{additional_details}

\section{Additional experiments}
\input{distribution}
\input{fullshot_results}
\input{compare_calibration_lora}
\input{comparison_withFISH}
\input{gpt_experiments}

\input{multi_task_appendix}

\input{generalization_theory}

\input{core_skills}

%% file: additional_details.tex
\clearpage
\section{Additional Experimental Details}

\subsection{Architecture Details on Fine-tuning}
\label{sec:optimization}
We fix the embeddings and the language model head of the pre-trained model during fine-tuning with prompts. We observe an improvement in the model's performance, when we fix the embedding layer while fine-tuning with SGD. This was also observed by \citet{kumar2022fine} for vision transformers. We fix the language model head, to stay consistent with the fact that  the weights of the language model head are an exact copy of the weights of the embedding layer in the pre-trained model. Furthermore, fixing the head for RoBERTa does as well as learning, but reduces the sizes of regions. For standard fine-tuning experiments, we fix only the embeddings of the pre-trained model. 

\subsection{Hyperparameter settings}
\label{sec:hyperparam}
For SGD, we follow the grid $\{2, 4, 8\}$ for batch size and $\{10^{-2}, 5 \times 10^{-3}, 10^{-3}\}$ for learning rate and apply a small weight decay of $10^{-4}$ on all the model parameters during training.
Following \cite{malladi2022kernel}, we train the model for $16 \times n \times k$ steps in $k$-shot tuning on a task with $n$ labels. However, to reduce training time for $4096$(and higher)-shot experiments, we stop training after $16 \times n \times 512$ steps and observe that the model converges to $100\%$ train accuracy during this time.
We fix the randomness of the model during FT by using seed $0$. In all our notations, a $k$-shot dataset for a task refers to a dataset with $k$ examples per classification class. For prompt-based fine-tuning, we do not use demonstrations in the context of each input example as proposed in \citet{gao2021making}.

\subsection{Calibration Error}
\label{sec:apx_calibration}

The ECE metric measures the difference between the confidence and accuracy of a model.
We use the implementation from \citet{guo2017calibration} and briefly describe the metric below.
Predictions are grouped into $M$ interval bins (each of size $1/M$)
and the accuracy of each bin is computed.
If $B_m$ is the set
of indices of samples whose prediction confidence falls into
the interval $I_m = [\frac{m-1}{M}, \frac{m}{M})$, then ECE is given by
\begin{align}
    \sum_{m=1}^{M} \frac{|B_m|}{n} | \mathrm{conf}(B_m) - \mathrm{acc} (B_m) |, \label{eq:calib}
\end{align}
where $\mathrm{acc}(B_m)$ and $\mathrm{conf}(B_m)$ denote the accuracy of the model and the average of all the predictions on the examples in the bin $B_m$ respectively.

One of the popular approaches to improve the calibration of a model is to learn a function on top of the logits of a trained model using a hold-out set. However, learning a function with $\epsilon$ ECE requires at least $\Omega(1/\epsilon^2)$ samples \cite{kumar2019verified}. This is difficult in few-shot settings, where we only have less than 100 examples to train with. Thus, grafting provides a direct way to improve the calibration of the model, with a minimal drop in performance.

%% file: distribution.tex
\ifthenelse{\boolean{arxiv}}
{
\input{table_fig/Distribution.tex}
}{}

\ifthenelse{\boolean{arxiv}}
{
\subsection{Distribution of graft parameters}  \label{sec:distribution}
We perform a closer analysis of the distribution of graft parameters for different downstream tasks in \cref{fig:layer_distribution_graft,fig:module_distribution_graft}.
Firstly we observe, in \Cref{fig:layer_distribution_graft}, that most of the graft region is concentrated in the middle layers for most tasks; AG News is an exception.
Furthermore, a closer look into the distribution of the graft parameters in \Cref{fig:module_distribution_graft} reveals an interesting pattern.
Most of the graft parameters are concentrated in three parameter types: (1) Value parameters of the attention module, (2) the first layer of the feed-forward module, and (3) LayerNorm parameters.
This observation could potentially provide a deeper understanding of the mechanism of fine-tuning and the role of pre-training.
The LayerNorm parameters in the grafting region are uniformly distributed across layers for all tasks, as evident in the bottom right of \Cref{fig:layer_distribution_graft}.
The layer-wise distribution of the value parameters and the first layer parameters of the feedforward module  in the grafting region show varying patterns among different tasks.
}{}

\ifthenelse{\boolean{arxiv}}{}{
\vspace{-0.1in}
}

%% file: table_fig/Distribution.tex
\ifthenelse{\boolean{arxiv}}{

\begin{figure}[!t]
    \centering
    \begin{subfigure}{0.24\textwidth}
    \centering
    \includegraphics[width=\textwidth]{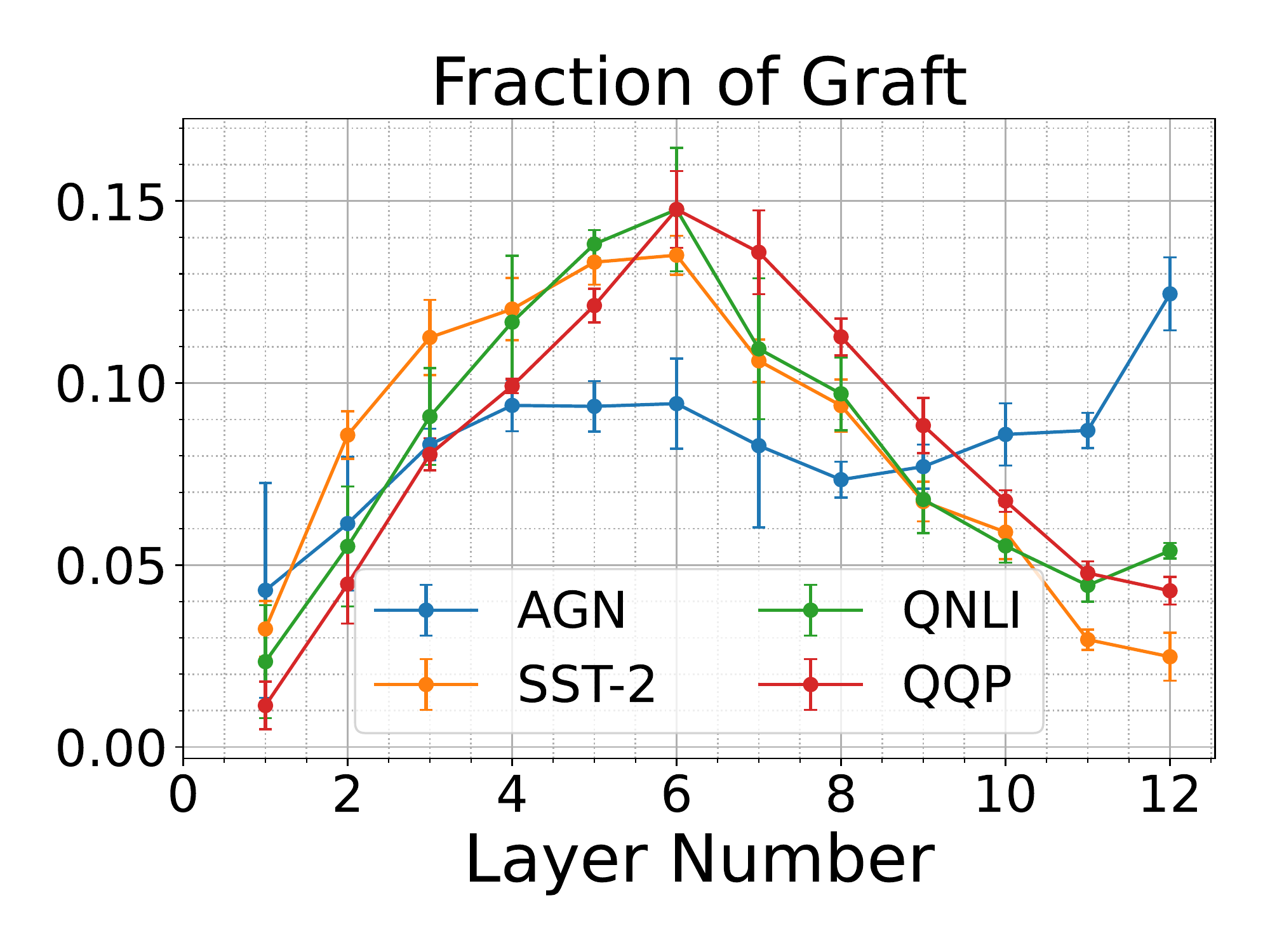}
    \end{subfigure}\hfill
    \begin{subfigure}{0.24\textwidth}
    \centering
    \includegraphics[width=\textwidth]{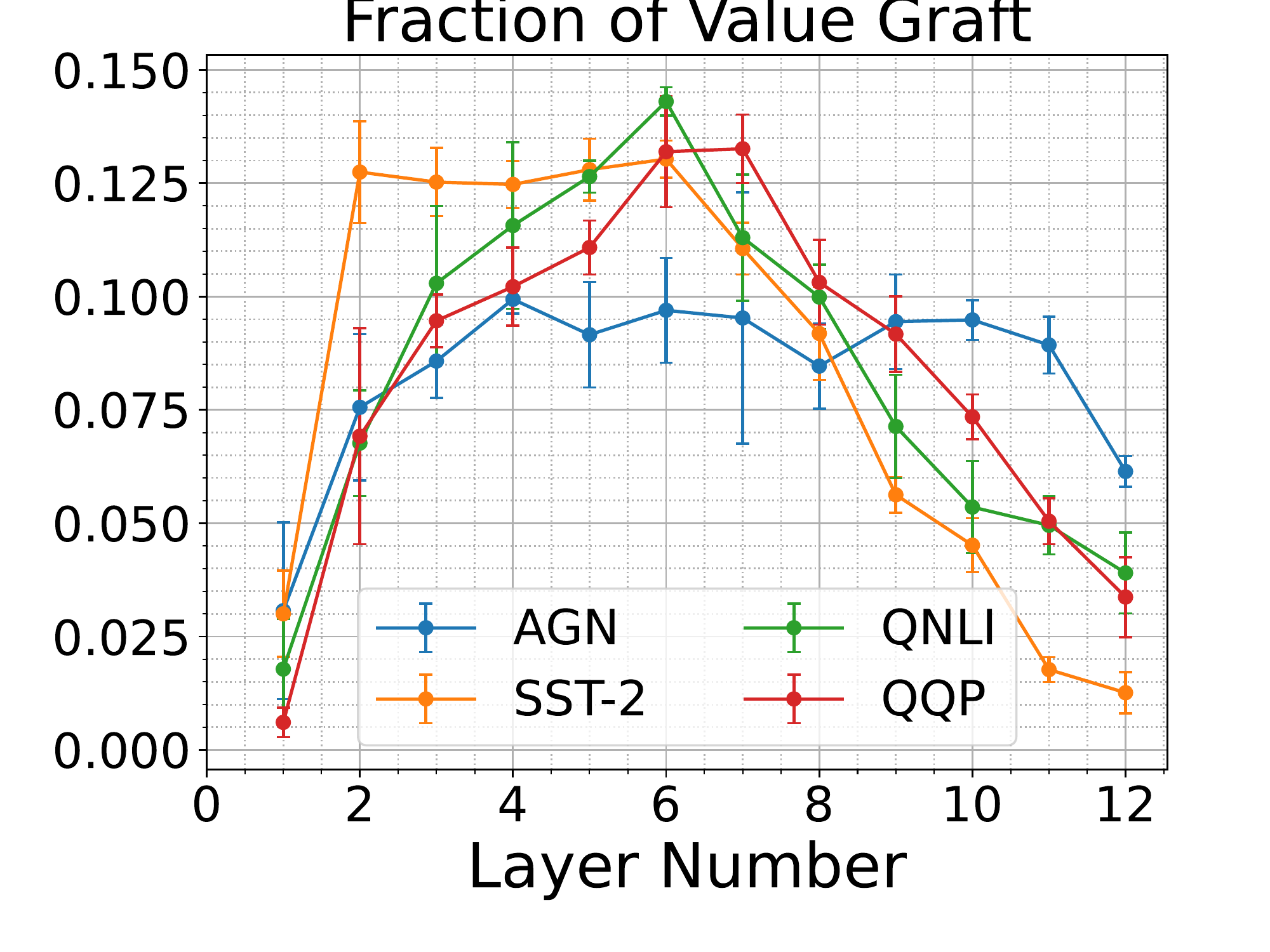}
    \end{subfigure}\hfill
    \begin{subfigure}{0.24\textwidth}
    \centering
    \includegraphics[width=\textwidth]{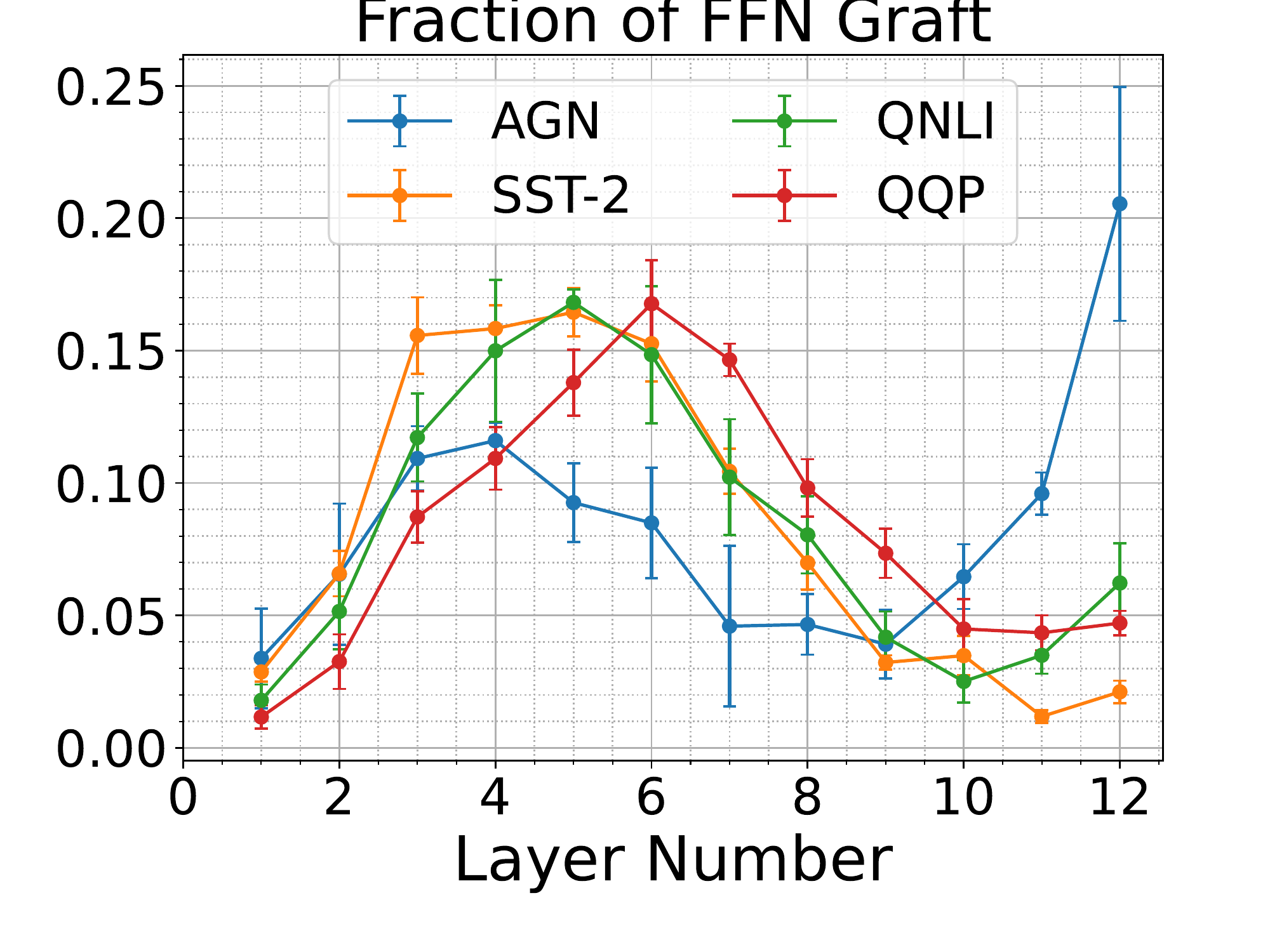}
    \end{subfigure}\hfill
    \begin{subfigure}{0.24\textwidth}
    \centering
    \includegraphics[width=\textwidth]{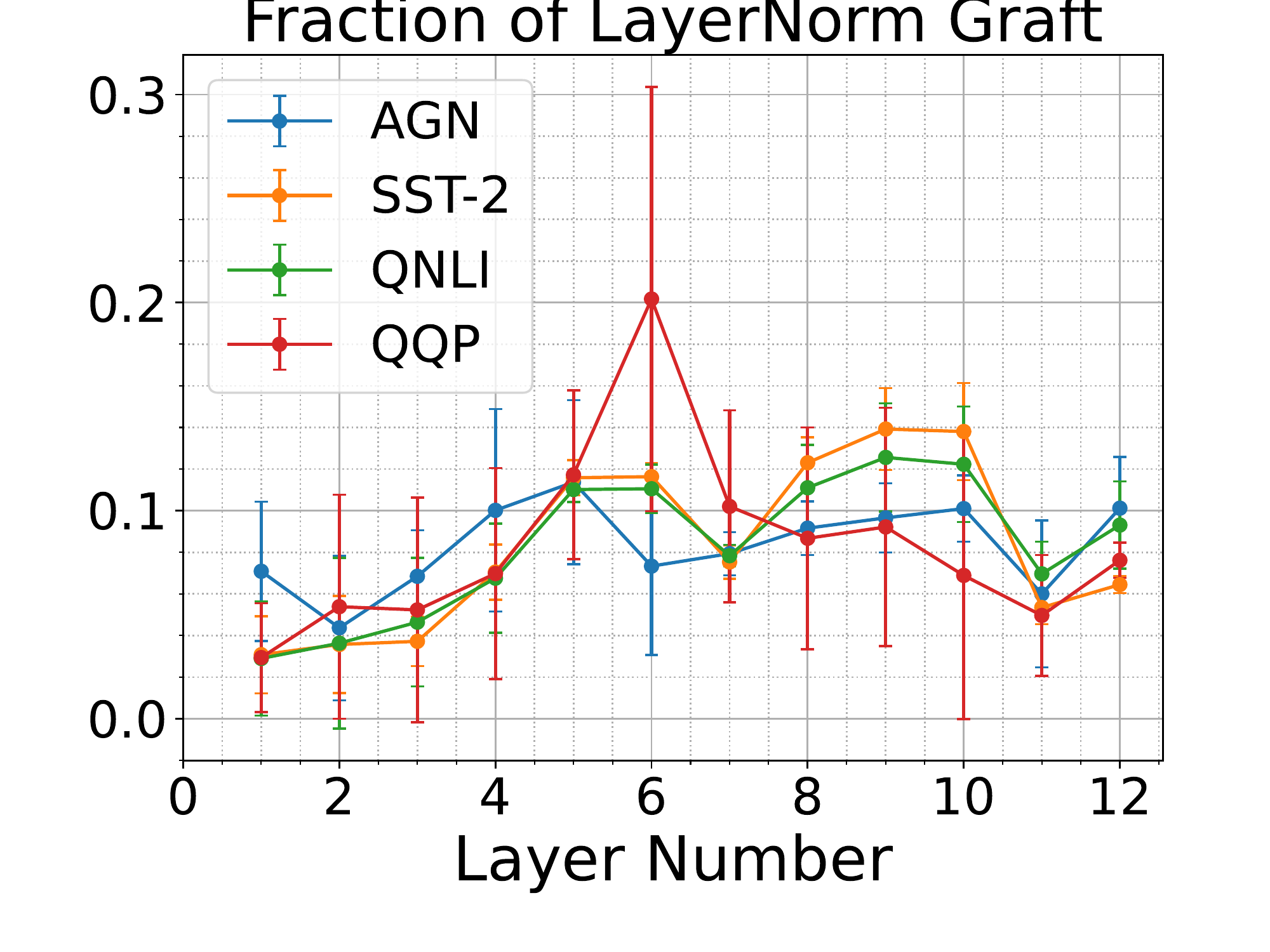}
    \end{subfigure}\hfill
    \caption{ Distribution of graft parameters in different regions of the model across layers. (TL) All parameters, (TR) Value parameters of attention module, (BL) first layer parameters of the feedforward module, (BR) LayerNorm parameters (feedforward and attention combined). For most of the tasks, the graft parameters are concentrated more in the middle layers. 
    }
    \label{fig:layer_distribution_graft}
\end{figure}

\begin{figure}[!t]
    \centering
    \begin{subfigure}{0.24\textwidth}
    \centering
    \includegraphics[width=\textwidth]{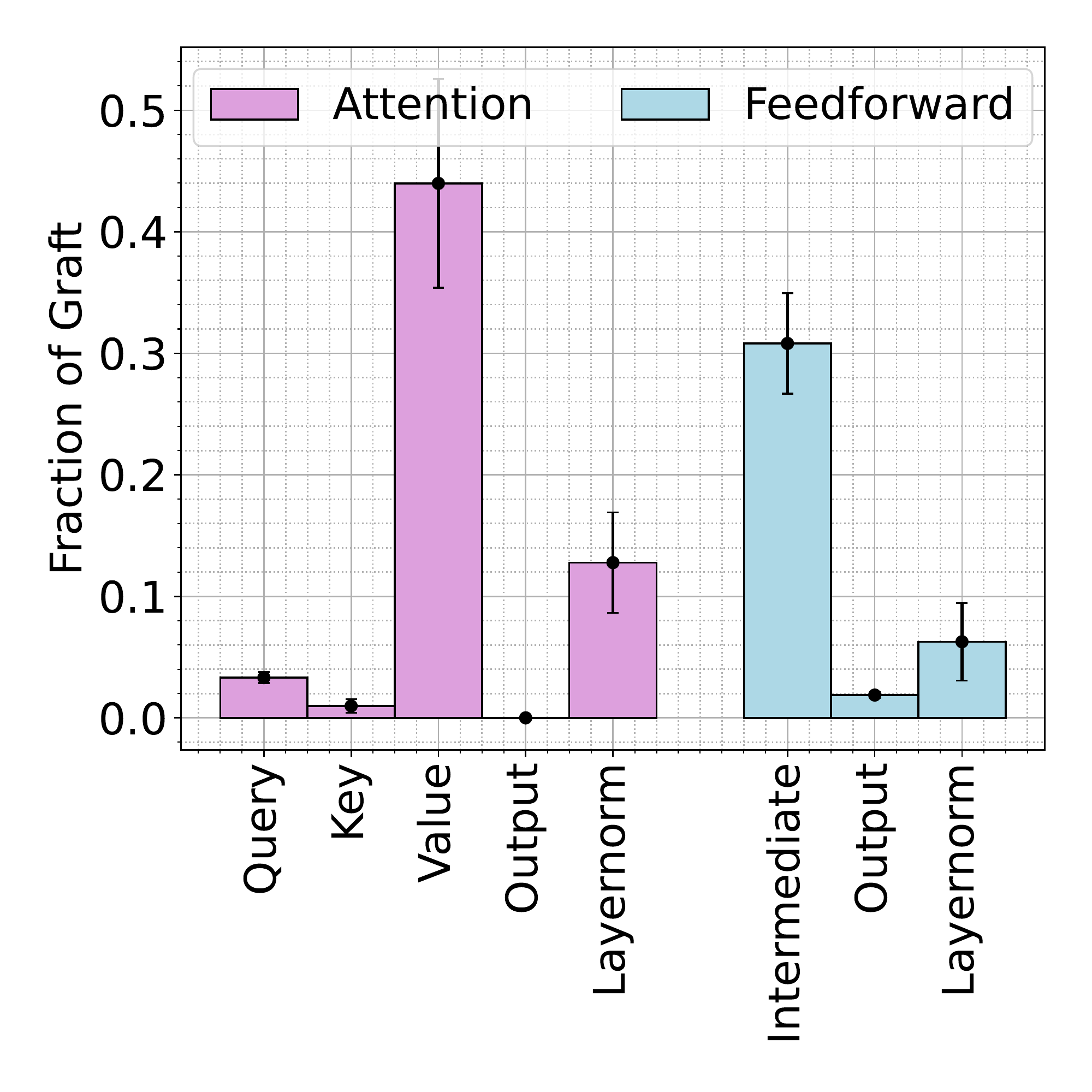}\caption { SST-2, 4096-shot}
    \end{subfigure}\hfill
    \begin{subfigure}{0.24\textwidth}
    \centering
    \includegraphics[width=\textwidth]{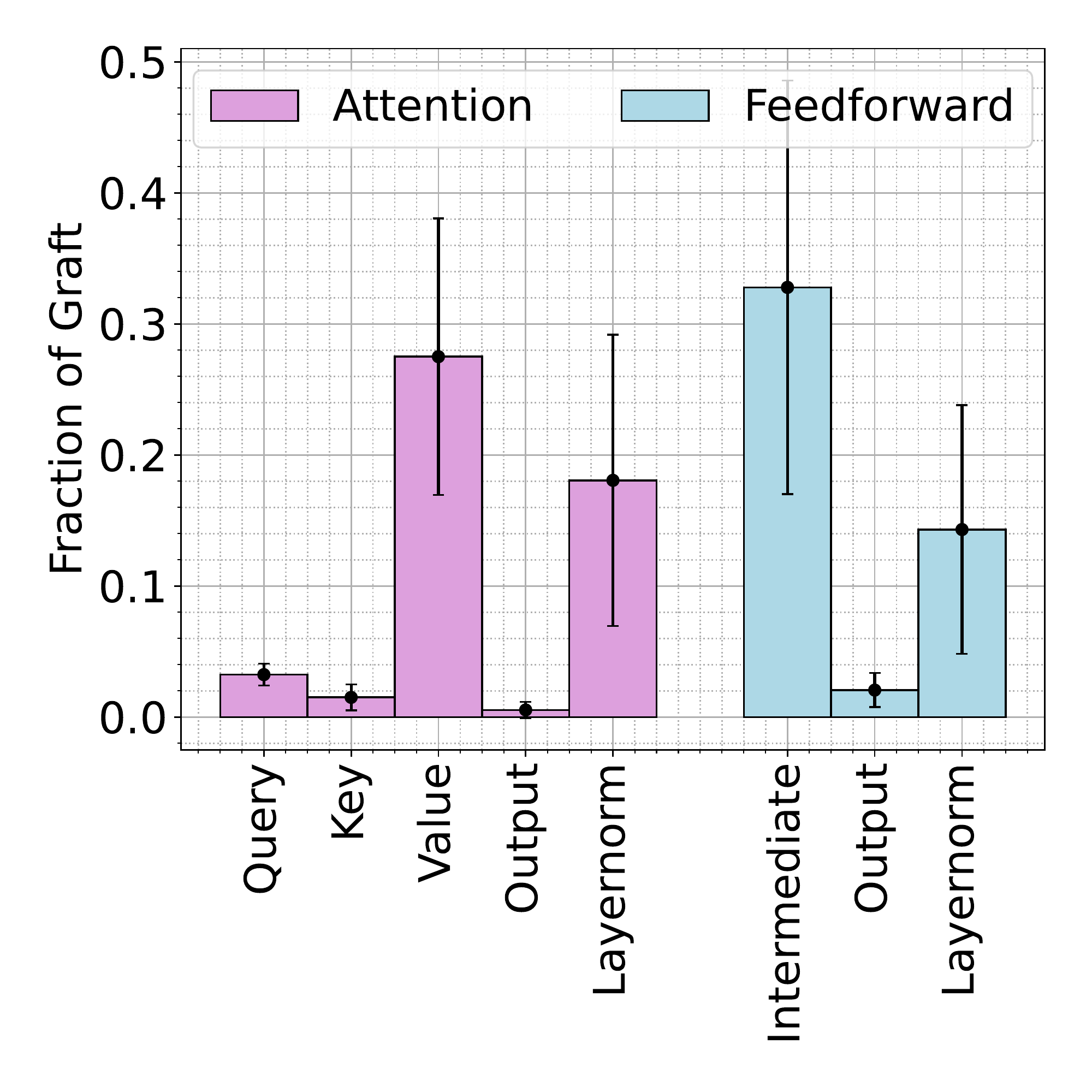}\caption { AG News, 4096-shot}
    \end{subfigure}\hfill
    \begin{subfigure}{0.24\textwidth}
    \centering
    \includegraphics[width=\textwidth]{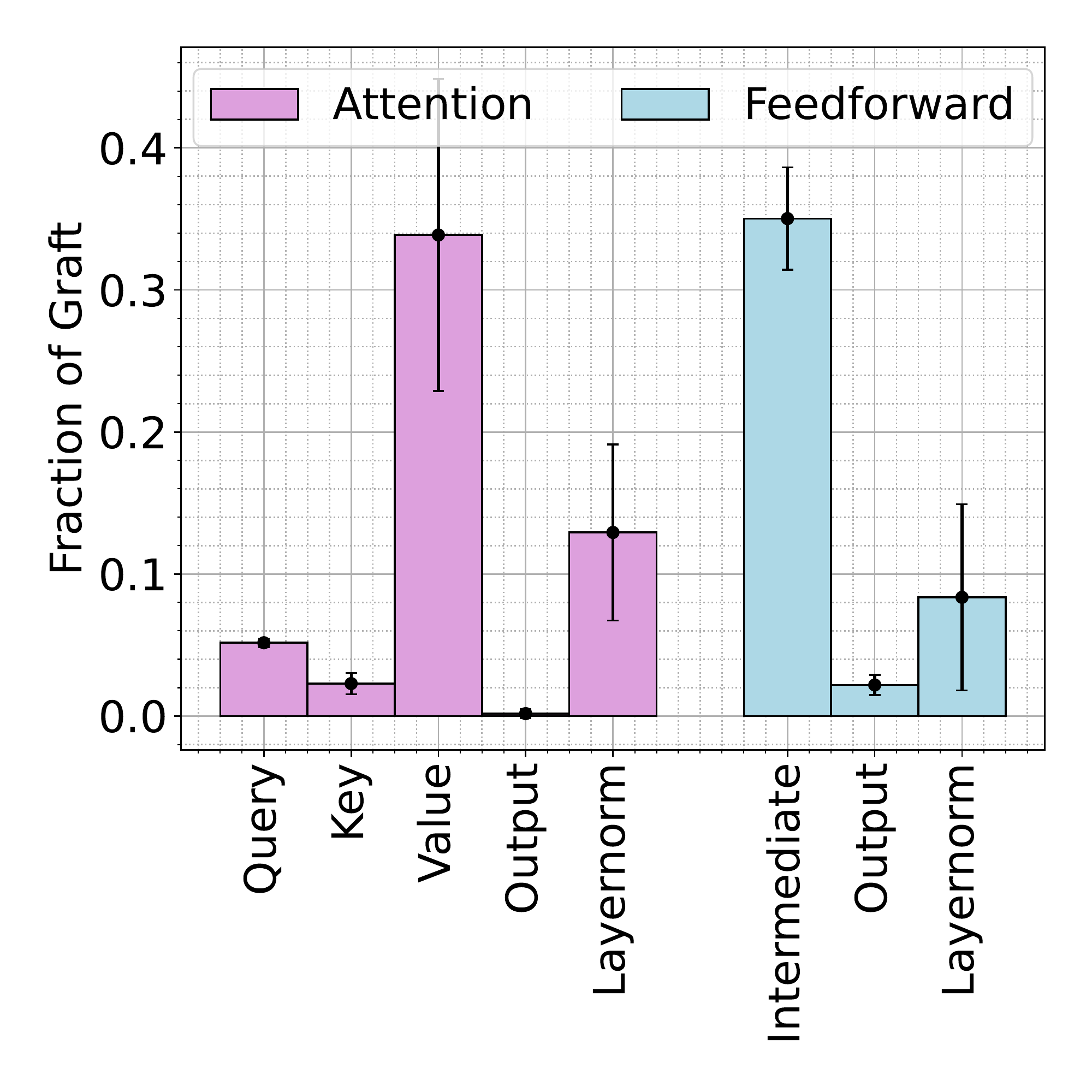}\caption { QNLI, 4096-shot}
    \end{subfigure}\hfill
    \begin{subfigure}{0.24\textwidth}
    \centering
    \includegraphics[width=\textwidth]{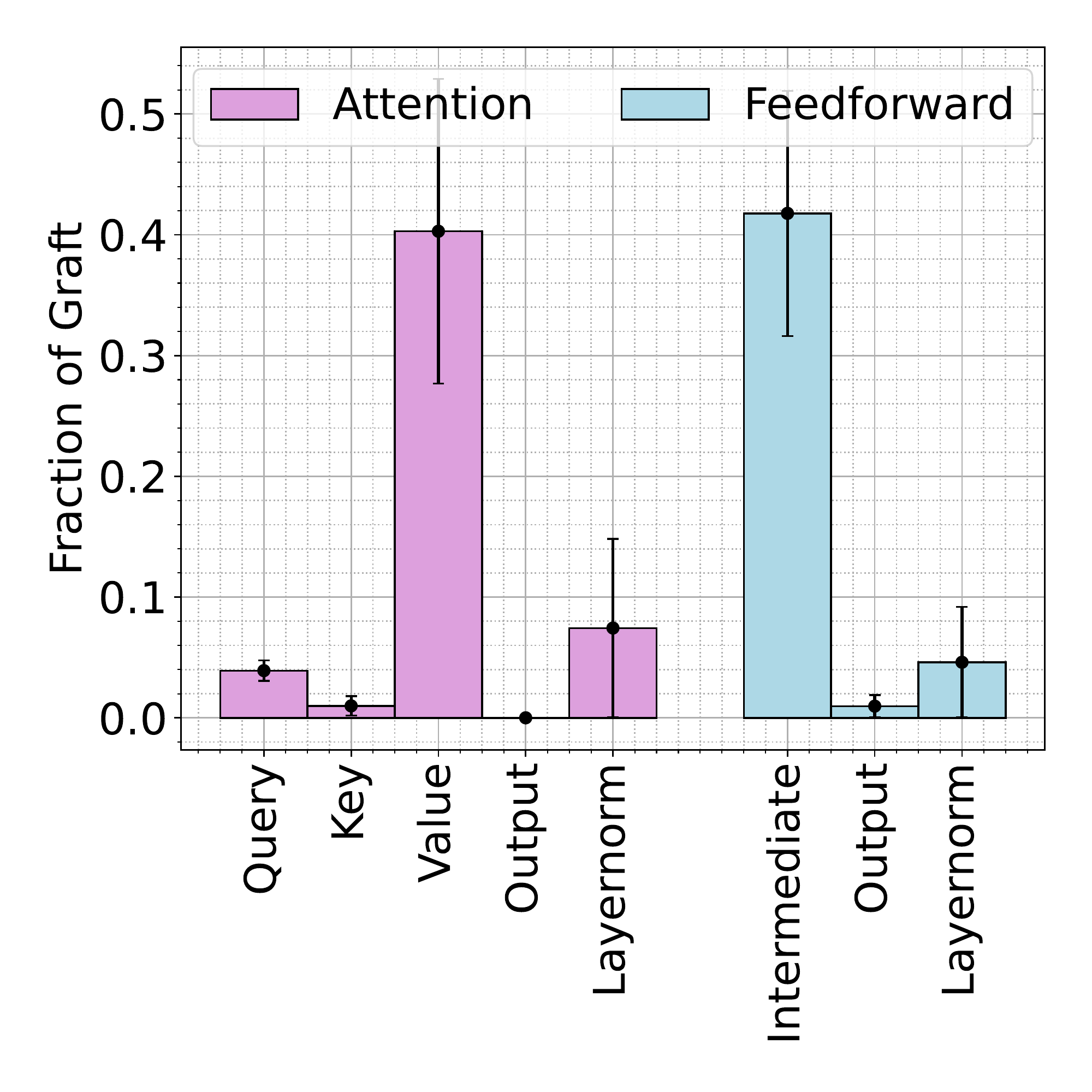}\caption { QQP, 4096-shot}
    \end{subfigure}\hfill
    \caption{Distribution of graft parameters in attention and feedforward modules. For feedforward module, Intermediate denotes its first layer and Output denotes its second layer. Most of the graft parameters are concentrated in the Value parameters of the attention module, the first layer of the feedforward module, and the LayerNorm parameters. 
    \vspace{-0.2in}
    }
    \label{fig:module_distribution_graft}
\end{figure}

}
{}

%% file: fullshot_results.tex
\subsection{Grafts with a larger training data}\label{sec:fullshot}
\input{table_fig/fullshot_graft}

Most experiments in the previous sections perform fine-tuning in the few-shot or mid-shot settings.
Here we verify that skill localization via grafting also happens in full-shot models.
We repeat the experiments in \cref{tab:roberta_patch_perf} for a few datasets with almost their entire training data, as we allow $50k$ training examples per classification
class. We make a random $95\%-5\%$ split of the training set to have a validation set for hyperparameter tuning. Since all the sentiment analysis tasks (MPQA, SST-2, CR, MR) considered in our paper have at most $10k$ training examples, we merge them to form one Sentiment dataset. For training models, we follow the same hyperparameter tuning procedure from \cref{sec:hyperparam}. With  $\sim 10 \times$ increase in training examples, we also increase the number of epochs during mask training by $10\times$, while using the same set of hyperparameters as given in \cref{sec:patch_properties}.

We observe that grafted models using $0.03\%$ sparse grafting regions can achieve $>95$\% of the final model performance, while also showing an absolute improvement of $~4.5$\% in calibration error. With a $10\times$ increase in the number of training examples, the necessary grafting region however doesn't increase by $10\times$. This indicates that the grafting region doesn't increase proportionally with the number of training samples, which points to some form of task-specific intrinsic dimensionality for fine-tuning. This is left for future exploration. 

%% file: table_fig/fullshot_graft.tex
\begin{table*}[htbp]
\centering
\caption{\looseness-1 We consider a larger training set for each downstream task from \cref{tab:roberta_patch_perf}, where we allow $50k$ training examples per classification class. The Sentiment dataset represents the combination of MR, CR, MPQA, and SST-2. The main findings are (1) Grafted models with $0.03\%$ sparse grafting regions can retrieve $>95\%$ of the FT accuracy while being better calibrated than the original model itself.
For single-sentence tasks, the grafted model shows only a $0.3\%$ drop in accuracy but an improvement of $6\%$ in the calibration error. Similarly for two-sentence tasks, the grafted model shows a $2.8\%$ drop in accuracy with an improvement of $4.0\%$ in the calibration error.\vspace{0.1in}}
\label{tab:fullshot_roberta}
%
\begin{tabular}{l|cc|cc|cc|cc} 
\toprule
      & \multicolumn{2}{c|}{FT} & \multicolumn{2}{c|}{0.03\% Sparse Graft} & \multicolumn{2}{c|}{0.06\% Sparse Graft} & \multicolumn{2}{c}{0.12\% Sparse Graft} \\
  \midrule
 Dataset & Acc. & ECE & Acc. & ECE  & Acc. & ECE & Acc. & ECE  \\
 \midrule
 Sentiment & 89.2 & 9.3 & 89.5 & 2.7 & 89.6 & 3.3 & 89.4 & 3.6\\
 AG News & 92.8 & 6.1 & 91.9 & 0.8 & 91.8 & 0.9 & 91.9 & 0.8\\
 \midrule
 QNLI & 91.8 & 5.5 & 88.9 & 1.2 & 89.2 & 1.2 & 89.6 & 1.3 \\
 SNLI & 88.6 & 5.1 & 85.9 & 1.4 & 86.2 & 1.3 &  86.3 & 1.4 \\
 \bottomrule
\end{tabular}
\end{table*}

%% file: compare_calibration_lora.tex
\subsection{OOD and calibration comparisons with parameter-efficient methods}\label{sec:peft}
\input{table_fig/compare_lora}

Since calibration and OOD generalization for sparse grafted models are better than fully fine-tuned models, a natural question is whether the number of parameters being updated is the main reason for this good performance.
To test this hypothesis, we perform two sets of experiments, (1) evaluate the calibration error and OOD accuracy of parameter-efficient fine-tuning methods, and (2) evaluate these metrics for grafted models at different levels of sparsity.

\paragraph{Parameter-efficient methods.} We report the calibration and the OOD performance of three parameter efficient methods, LoRA~\citep{hu2022lora} and two adapter-based methods~\citep{houlsby2019parameter,pfeiffer-etal-2021-adapterfusion} in \cref{tab:compare_peft,tab:compare_peft_ood}. We do not change the hyperparameter setting for these techniques from their corresponding papers, i.e. we train these methods with AdamW optimizer and follow the corresponding papers for hyperparameter tuning. We report at different sparsity levels, to allow a fairer comparison across the methods.

We observe that at similar sparsity levels, grafting shows a better calibration and OOD performance with a small accuracy drop, compared to the parameter-efficient methods under consideration.
This shows that the sparsity of the grafting region is not the only reason behind its better OOD and calibration performance.
Other factors like optimization algorithm and the grafting procedure could also be responsible and a deeper exploration of various factors is left for future work. 

\paragraph{Grafts with different sparsity.} In \cref{tab:compare_peft,tab:compare_peft_ood}, we show the OOD and calibration performance of grafting at different levels of sparsity. They show a smooth decrease with the sparsity of the grafting region, with the grafted model's performance increasing smoothly. Thus for grafting, the calibration error and OOD generalization correlates with number of the size of the graft.

%% file: table_fig/compare_lora.tex
\begin {table*}[!t]
\centering
\caption { \looseness-1  Comparing the accuracy (${}^\dagger$F1) in \% and the calibration error (ECE) for model grafting, LoRA \cite{hu2022lora}, and two adapter methods \cite{pfeiffer-etal-2021-adapterfusion,houlsby2019parameter} at different sparsity levels. Results are in the 4096-shot setting.  The major observations are: (a) At similar sparsity levels ($0.05-0.1$\%), grafting achieves an improvement of $4.6$\% in  calibration error while being only $1.7$\% lower in accuracy, implying parameter count is not the only reason behind the calibration results for grafting.  (b) For grafting,  calibration error, and accuracy change smoothly with the grafting region sparsity.\vspace{0.1in} }
\label {tab:compare_peft}
\footnotesize
\begin {tabular} {l|c|cc|cc|cc|cc|cc} \toprule
  \midrule
    & &  \multicolumn {2} {c|} {SST-2} &  \multicolumn {2} {c|} {AGNews} &  \multicolumn {2} {c|} {QNLI} &  \multicolumn {2} {c|} {QQP${}^\dagger$} & \multicolumn {2} {c} {Average}  \\
  \midrule
 \multicolumn {1} {c|} {} & \multicolumn {1} {c|} {\makecell{Rank/\\Sparsity}} & Acc. & ECE & Acc. & ECE & Acc. & ECE & F1 & ECE & Acc./F1 & ECE \\
 \midrule 
 \multirow {3} {*} {\rotatebox[origin=c]{90}{LoRA}} & \cellcolor{gray!10} 1  (0.05\%) & \cellcolor{gray!10} 92.8  {\tiny  (0.5)} & \cellcolor{gray!10} 6.2 {\tiny (0.7)}  &	\cellcolor{gray!10} 93.1 {\tiny (0.2)} & \cellcolor{gray!10}  5.1 {\tiny (0.4)} &\cellcolor{gray!10}  87.6  {\tiny (0.3)} & \cellcolor{gray!10}  8.0 {\tiny (1.9)}	& \cellcolor{gray!10} 79.4 {\tiny (0.2)} & \cellcolor{gray!10} 9.9 {\tiny (2.8)} & \cellcolor{gray!10} 88.2 & \cellcolor{gray!10} 7.3 \\
 & 2  (0.1\%) & 93.0 {\tiny (0.2)} & 5.7 {\tiny (0.8)}	& 93.0 {\tiny (0.3)} & 5.1  {\tiny (0.9)} &	87.5  {\tiny (0.4)} & 9.0 {\tiny (2.4)}	& 79.2 {\tiny (0.6)} & 13.7  {\tiny (3.6)} & 88.2 & 8.4 \\
 & 8  (0.4\%) & 	93.4 {\tiny (0.4)} & 5.8 {\tiny (0.9)} &	93.1 {\tiny (0.1)} & 5.7 {\tiny (0.6)}	& 87.6 {\tiny (0.3)} & 9.7 {\tiny (2.1)}	& 79.7 {\tiny (0.3)} & 14.3 {\tiny (2.3)} & 88.5 & 8.9  \\
 \midrule
 \multirow {4} {*} {\rotatebox[origin=c]{90}{\makecell{Houlsby\\ et al.}}}  & 1  (0.025\%) & 92.9  {\tiny (0.5)}  & 6.3  {\tiny (1.0)}  &  93.2  {\tiny (0.2)}  & 5.5  {\tiny (0.9)}   & 88.1  {\tiny (0.6)} & 10.2  {\tiny (1.7)} & 79.7  {\tiny (0.4)} & 13.6  {\tiny (3.0)} & 88.5 & 8.9 \\
  & \cellcolor{gray!10} 2  (0.05\%) & \cellcolor{gray!10}  92.9  {\tiny (0.6)}  & \cellcolor{gray!10}  5.8  {\tiny (1.1)}  &  \cellcolor{gray!10} 93.0  {\tiny (0.4)}  & \cellcolor{gray!10} 5.4  {\tiny (1.4)}  & \cellcolor{gray!10} 88.0  {\tiny (0.4)} & \cellcolor{gray!10} 9.8  {\tiny (1.8)} & \cellcolor{gray!10} 79.3  {\tiny (02)}  & \cellcolor{gray!10} 13.7  {\tiny (2.8)} & \cellcolor{gray!10} 88.3 & \cellcolor{gray!10} 8.7 \\
 & 16  (0.4\%) & 93.2 {\tiny (0.3)} & 6.2 {\tiny (0.9)} &	93.3 {\tiny (0.2)} & 5.5 {\tiny (0.8)} &	88.4 {\tiny (0.8)} & 11.3 {\tiny (0.6)} &	80.0 {\tiny (0.3)} & 15.1 {\tiny (1.3)} & 88.7 & 9.5 \\
  & 64  (1.6\%) & 92.6 {\tiny (0.6)} & 6.9 {\tiny (0.8)} &	93.3 {\tiny (0.1)} & 6.1 {\tiny (0.5)} &	88.7 {\tiny (0.2)} & 10.3 {\tiny (1.6)} &	80.1 {\tiny (0.3)} & 15.8 {\tiny (0.2)} &  88.7 & 9.8 \\
 \midrule
 \multirow {4} {*} {\rotatebox[origin=c]{90}{\makecell{Pfeiffer\\ et al.}}}   & 1  (0.025\%) & 92.8  {\tiny (0.6)}  & 6.4  {\tiny (1.0)}  &  93.2  {\tiny (0.2)}  & 5.4  {\tiny (0.9)}  & 88.0  {\tiny (0.7)} & 10.1  {\tiny (1.8)} & 79.7  {\tiny (0.4)} & 13.7  {\tiny (3.1)}  & 88.4 & 8.9 \\
 & \cellcolor{gray!10} 2  (0.05\%) & \cellcolor{gray!10} 93.1  {\tiny (0.4)}  &  \cellcolor{gray!10} 6.0  {\tiny (1.4)}  & \cellcolor{gray!10} 93.0  {\tiny (0.2)}  & \cellcolor{gray!10} 4.5  {\tiny (0.6)}  &  \cellcolor{gray!10} 87.4  {\tiny (0.4)} &\cellcolor{gray!10}  7.7  {\tiny (2.5)}  & \cellcolor{gray!10} 78.9  {\tiny (0.2)} & \cellcolor{gray!10} 12.5  {\tiny (4.2)}  & \cellcolor{gray!10} 88.1 & \cellcolor{gray!10}  7.7 \\
 & 16  (0.4\%) & 92.5 {\tiny (0.4)} & 6.2 {\tiny (0.8)} &	93.1 {\tiny (0.2)} & 5.1 {\tiny (0.8)} &	87.8 {\tiny (0.4)} & 9.5 {\tiny (1.6)} &	79.6 {\tiny (0.4)} & 13.5 {\tiny (2.2)} &  88.3 & 8.6 \\
 & 64  (1.6\%) & 93.1 {\tiny (0.5)} & 6.7 {\tiny (0.5)} &	93.1 {\tiny (0.1)} & 5.8 {\tiny (0.7)} &	87.9 {\tiny (0.3)} & 9.9 {\tiny (2.2)} &	79.6 {\tiny (0.4)} & 14.9 {\tiny (3.5)} & 88.5 & 9.4\\
 \midrule
 \multirow {5} {*} {\rotatebox[origin=c]{90}{Grafting}} & 0.01\% & 92.4 {\tiny (0.1)} & 3.2 {\tiny (0.4)} &	91.1 {\tiny (0.9)} & 0.9 {\tiny (0.2)} &	84.7 {\tiny (0.6)} & 1.0  {\tiny (0.3)} &	76.3 {\tiny (0.4)} & 3.5  {\tiny (0.7)} & 86.1 & 2.2 \\
 & \cellcolor{gray!10}   0.05\% & \cellcolor{gray!10} 92.5  {\tiny (0.5)} & \cellcolor{gray!10} 4.0  {\tiny (0.9)} & \cellcolor{gray!10} 89.9 {\tiny(0.5)} &  \cellcolor{gray!10} 0.9 {\tiny(0.1)} & \cellcolor{gray!10} 86.0  {\tiny (0.4)} & \cellcolor{gray!10} 1.9 {\tiny (0.8)} & \cellcolor{gray!10} 77.6 {\tiny(0.4)} &  \cellcolor{gray!10} 3.9 {\tiny(0.8)} & \cellcolor{gray!10} 86.5 & \cellcolor{gray!10} 2.7 \\
 & 0.1\% & 92.5 {\tiny (0.4)}  & 4.4  {\tiny (0.7)} & 90.8 {\tiny (0.1)} &  1.1 {\tiny (0.3)}  & 86.0 {\tiny (0.4)}  & 2.2 {\tiny (0.9)}  & 78.2 {\tiny (0.0)} & 4.5 {\tiny (1.3)} & 86.9 & 3.1 \\
 & 1\% & 92.3 {\tiny (0.3)}    & 4.6 {\tiny (0.7)}  & 91.0 {\tiny (0.3)} & 1.2 {\tiny (0.2)} & 86.5 {\tiny (0.4)} &  3.1 {\tiny (1.1)} & 78.7 {\tiny (0.2)} & 5.4 {\tiny (1.7)} & 87.2 &  3.6 \\
 & 100\% & 92.3 {\tiny (0.3)} & 7.4 {\tiny (0.3)} &	92.7 {\tiny (0.4)} & 6.8 {\tiny (0.3)} &	88.0 {\tiny (0.8)} & 10.2 {\tiny (0.0)} &	79.6 {\tiny (0.6)} & 10.1 {\tiny (4.2)} & 88.2 &  8.6 \\
 \bottomrule
\end {tabular}
\vspace {-0.1in}
\end {table*}

\begin {table*}[!t]
\centering
\caption { \looseness-1 Comparing the zero-shot OOD performance  for model grafting, LoRA \cite{hu2022lora}, and two adapter methods \cite{pfeiffer-etal-2021-adapterfusion,houlsby2019parameter} at different sparsity levels. Results are in the 4096-shot setting.   The major observations are: (a) At similar sparsity levels ($0.05-1$\%), grafting achieves an improvement of $7.9$\% in OOD performance while being a $1.0$\% lower in ID accuracy, implying parameter count is not the only reason behind the OOD results for grafting.  (b) For grafting,  the OOD and ID accuracy change smoothly with the grafting region sparsity.\vspace{0.1in} }
\label {tab:compare_peft_ood}
\footnotesize
\begin {tabular} {l|c|ccc|ccc|cc} \toprule
  \midrule
    & &  \multicolumn {3} {c|} {ID task: MPQA} &  \multicolumn {3} {c|} {ID task: QNLI} & & \\
  \midrule
 \multicolumn {1} {c|} {Method} & \multicolumn {1} {c|} {Rank/Sparsity} & MPQA & Yelp & SST-2 & QNLI & MNLI (0/1) & SNLI (0/1)  & Avg. ID & Avg. OOD \\
 \midrule 
 \multirow {3} {*} {LoRA} & \cellcolor{gray!10} 1  (0.05\%) & \cellcolor{gray!10} 89.1 {\tiny (0.7)} & \cellcolor{gray!10} 79.8 {\tiny (4.2)} & \cellcolor{gray!10} 64.7  {\tiny (5.7)} & \cellcolor{gray!10} 87.6 {\tiny (0.3)}  & \cellcolor{gray!10} 57.2  {\tiny (4.4)} & \cellcolor{gray!10} 57.9 {\tiny (4.8)}  & \cellcolor{gray!10} 88.3 & \cellcolor{gray!10} 64.9 \\
 & 2  (0.1\%) & 89.1 {\tiny (0.4)} &  74.3 {\tiny (7.7)} & 65.5 {\tiny (5.0)} &  87.5 {\tiny (0.4)} & 54.7  {\tiny (3.7)} & 54.0 {\tiny (4.3)} & 88.3	& 62.1 \\
 & 8  (0.4\%) & 89.2  {\tiny (0.7)} &	83.9  {\tiny (1.5)} & 79.6  {\tiny (3.5)} & 87.6  {\tiny (0.3)} & 64.2  {\tiny (3.9)} &	68.5  {\tiny (4.9)} & 88.4 & 74.1 \\
 \midrule
 \multirow {4} {*} {\makecell{Houlsby\\ et al.}}  & 1  (0.025\%) & 88.6 {\tiny(0.7)}  & 81.6 {\tiny(1.8)} &  74.8 {\tiny(2.0)} & 88.1 {\tiny(0.6)} & 58.4 {\tiny(3.8)} & 57.8 {\tiny(7.2)} & 88.4  & 68.2 \\
  & \cellcolor{gray!10} 2  (0.05\%)  & \cellcolor{gray!10} 88.7 {\tiny(0.6)} &  \cellcolor{gray!10} 83.0 {\tiny(0.9)} &  \cellcolor{gray!10} 77.5 {\tiny(4.4)} &  \cellcolor{gray!10} 88.0 {\tiny(0.4)}  & \cellcolor{gray!10} 60.6 {\tiny(0.7)} & \cellcolor{gray!10}  62.7 {\tiny(3.8)} & \cellcolor{gray!10} 88.4 & \cellcolor{gray!10} 71.0 \\
 & 16  (0.4\%)  & 88.1  {\tiny (0.9)}	 & 82.7  {\tiny (2.3)} &	77.9  {\tiny (3.1)} & 88.4  {\tiny (0.8)} & 61.9  {\tiny (5.3)}	& 61.8  {\tiny (7.8)} & 88.3 & 71.1 \\
  & 64  (1.6\%) & 88.1  {\tiny (0.6)}	& 80.7  {\tiny (3.3)} & 75.5  {\tiny (4.8)} & 88.7  {\tiny (0.2)} &	63.1  {\tiny (3.2)} &	63.2  {\tiny (5.1)} & 88.4 & 70.6\\
 \midrule
 \multirow {4} {*} {\makecell{Pfeiffer\\ et al.}}   & 1  (0.025\%)  & 88.6 {\tiny(1.0)} & 77.0 {\tiny(2.2)} &  71.6 {\tiny(2.4)} & 88.0 {\tiny(0.7)}  & 56.9 {\tiny(5.4)} &  58.3 {\tiny(8.3)} & 88.3 & 66.0 \\
 & \cellcolor{gray!10} 2  (0.05\%)  & \cellcolor{gray!10} 88.6 {\tiny(0.9)} & \cellcolor{gray!10}  77.1 {\tiny(6.2)} & \cellcolor{gray!10}  70.0 {\tiny(7.1)}  & \cellcolor{gray!10} 87.4 {\tiny(0.4)}  & \cellcolor{gray!10} 61.5 {\tiny(3.9)} & \cellcolor{gray!10} 63.8 {\tiny(5.5)}  &  \cellcolor{gray!10} 88.0 & \cellcolor{gray!10} 68.1 \\
 & 16  (0.4\%) & 88.4  {\tiny (0.5)} & 80.3  {\tiny (1.4)}	& 75.9  {\tiny (2.1)} & 87.8  {\tiny (0.4)} &	66.5  {\tiny (1.8)}	& 68.1  {\tiny (5.3)} & 88.1 &   71.2 \\
 & 64  (1.6\%) & 88.2  {\tiny (0.9)} & 78.2  {\tiny (6.4)}	& 75.4  {\tiny (6.8)} & 87.9  {\tiny (0.3)} &	66.3  {\tiny (2.7)} & 66.7  {\tiny (7.9)}  &  88.1 & 71.7 \\
 \midrule
 \multirow {5} {*} {Grafting } & 0.01\% & 88.1  {\tiny (0.4)} & 83.7  {\tiny (0.7)} & 84.6  {\tiny (0.6)} & 84.7  {\tiny (0.6)}	& 71.8  {\tiny (1.8)} & 80.1  {\tiny (2.9)} & 86.4 & 80.1 \\
 & \cellcolor{gray!10}  0.05\% & \cellcolor{gray!10} 88.8 {\tiny (0.5)}  &\cellcolor{gray!10}  81.6 {\tiny (0.6)} & \cellcolor{gray!10}  83.1  {\tiny (1.1)} & \cellcolor{gray!10} 86.0  {\tiny (0.4)}   & \cellcolor{gray!10} 71.1 {\tiny (0.5)}  & \cellcolor{gray!10} 79.7 {\tiny (0.9)} & \cellcolor{gray!10} 87.4 & \cellcolor{gray!10} 78.9\\
 & 0.1\% & 88.5  {\tiny (0.5)} & 80.8 {\tiny (0.9)} &  82.6 {\tiny (0.5)}  & 86.0  {\tiny (0.4)}  & 70.9 {\tiny (1.57)}   & 78.3 {\tiny (2.2)} &  87.3 & 78.2 \\
 & 1\% & 88.9 {\tiny (0.5)} &  79.8 {\tiny (0.7)} &  81.7 {\tiny (1.3)} & 86.5  {\tiny (0.4)}  & 70.6 {\tiny (2.5)}  &  74.8 {\tiny (4.7)} &  87.7 & 76.7 \\
 & 100\% & 88.9  {\tiny (0.6)} & 74.2  {\tiny (2.3)} &	77.6  {\tiny (4.9)} & 88.0  {\tiny (0.8)}	& 65.7  {\tiny (2.6)} & 62.1  {\tiny (5.7)} &  88.5 & 69.9 \\
 \bottomrule
\end {tabular}
\vspace {-0.1in}
\end {table*}

%% file: comparison_withFISH.tex
\subsection{Overlap with FISH~\cite{sung2021training} mask}

\citet{sung2021training} proposed a parameter-efficient technique called Fisher-Induced Sparse uncHanging (FISH) Mask training.  The authors select 
a sparse set of parameters containing the largest Fisher information  in the pre-trained model  to fine-tune for a specific task and the FISH mask represents the selected parameters. We compute the overlap of our grafting regions with the FISH mask at different sparsity levels in \cref{tab:fish_overlap}.
If $\patch_{FISH}$ denotes the FISH mask, and $\patch$ denotes the grafting region given by \cref{eq:opt_patch}, then overlap score is defined as the intersection over union measure, i.e. 
($\nicefrac{ \abs{ \patch_{FISH} \cap \patch } }  {  \abs{ \patch_{FISH} \cup \patch }  }$).
Furthermore, we report the performances of the grafted models formed with the FISH mask and grafted models with a random mask as a baseline. Our observations can be summarized as follows: (a) There is a minimal overlap ($<10$\%) between the FISH mask and our grafting regions. (b) Sparse FISH masks do not effectively serve as suitable grafting regions, while the performance of grafted models using FISH masks improves with decreasing sparsity. This shows that full model fine-tuning need not use the top parameters with the largest Fisher information at pre-training.

\begin{table*}[!htbp]
\centering
\caption{Measuring overlap between FISH mask \cite{sung2021training} and model grafting at different sparsity levels.  If $\patch_{FISH}$ denotes the FISH mask, and $\patch$ denotes our grafting region (computed using with \cref{eq:opt_patch}), then overlap is defined by 
($\nicefrac{ \abs{ \patch_{FISH} \cap \patch } }  {  \abs{ \patch_{FISH} \cup \patch }  }$). Furthermore, we report the accuracy  (${}^\dagger$F1) in \% for our grafted models, grafted models with FISH mask, and grafted models with random grafting regions. Our observations are: (a) overlap is $<10$\% between the FISH mask and our grafting regio, and(b) There is an average 10\% gap between the performance of our grafted models and grafted models with FISH mask at 0.01\% sparsity, which reduces with decreasing sparsity levels. \vspace{0.1in}  }
\label{tab:fish_overlap}
\begin{tabular}{l|c|c|cc|cc|cc} \toprule
 Dataset & Sparsity & Overlap (in \%) & \multicolumn{2}{c|}{Graft} & \multicolumn{2}{c|}{Graft with FISH mask} & \multicolumn{2}{c}{Random Graft} \\
 \midrule
 & & & Acc/F1 & ECE & Acc/F1 & ECE & Acc/F1 & ECE \\
 \midrule
 \multirow{3}{*}{SST-2} & 0.01\% & 9.2 {\tiny(1.3)} &  92.4  {\tiny(0.1)} & 3.1 {\tiny(0.4)} & 90.8 {\tiny(0.3)} &  8.5 {\tiny(0.6)}  & 77.8 {\tiny(0.1)} & 4.7 {\tiny (0.1)} \\
 & 0.1\% & 4.1 {\tiny(1.7)} & 92.5 {\tiny(0.4)} & 4.4 {\tiny (0.7)}  & 91.5 {\tiny(0.3)} & 8.1 {\tiny(0.4)}   & 77.8 {\tiny(0.2)} & 4.8 {\tiny (0.4)} \\
 & 1\%   & 1.9 {\tiny(0.5)} & 92.3 {\tiny(0.3)} & 4.6 {\tiny (0.7)} & 91.6 {\tiny(0.2)} & 8.0 {\tiny(0.5)}  & 78.7 {\tiny(0.5)} & 4.5 {\tiny (0.5)} \\
 \midrule
 \multirow{3}{*}{AGNews} & 0.01\% & 7.6 {\tiny(3.3)}  &  91.1 {\tiny(0.9)}  & 0.9 {\tiny(0.2)}  & 80.3 {\tiny(5.8)}  & 5.1 {\tiny(1.6)} & 65.8 {\tiny (0.1)} & 6.8 {\tiny (0.1)}\\
 & 0.1\% & 1.8  {\tiny(0.2)} & 90.8 {\tiny(0.1)}  & 1.1 {\tiny(0.3)} &  82.4 {\tiny(4.8)} &  5.1 {\tiny(2.7)} &  66.0 {\tiny (0.2)} & 6.7 {\tiny (0.1)}\\
 & 1\%   & 1.1 {\tiny(0.1)}  & 91.0 {\tiny(0.3)}  & 1.2 {\tiny(0.2)}  &  71.8 {\tiny(13.2)} &  7.9 {\tiny(6.0)} & 67.8 {\tiny (0.7)} & 6.7 {\tiny (0.3)}\\
 \midrule
 \multirow{3}{*}{QNLI} & 0.01\% & 8.1 {\tiny(3.1)} & 84.7 {\tiny(0.6)}  &  1.0 {\tiny(0.3)} & 65.1 {\tiny(8.9)}  & 22.5 {\tiny(11.8)} & 50.7 {\tiny(0.0)} & 34.5 {\tiny (0.0)}\\
 & 0.1\% & 4.8 {\tiny(2.0)} & 86.0 {\tiny(0.4)}  &  2.2 {\tiny (0.9)} & 81.3 {\tiny(5.4)}  & 12.7 {\tiny(3.5)} & 50.7 {\tiny(0.1)} & 34.5 {\tiny (0.2)}  \\
 & 1\%   &  1.9 {\tiny(0.6)} & 86.5  {\tiny(0.4)}  & 3.1 {\tiny (1.1)} & 78.9 {\tiny(12.2)}  & 16.6 {\tiny(10.4)} & 50.7 {\tiny(0.1)} & 35.2 {\tiny (0.3)} \\
 \midrule
 \multirow{3}{*}{QQP}${}^\dagger$ & 0.01\% & 11.4 {\tiny(5.9)} &  76.3 {\tiny(0.4)}  & 3.5 {\tiny(0.7)} & 57.0 {\tiny(10.5)} & 16.9 {\tiny(4.9)} & 51.6 {\tiny (0.0)} & 36.2 {\tiny (0.0)}\\
 & 0.1\% & 6.6 {\tiny(4.7)} & 78.2  {\tiny(0.0)}   & 4.5 {\tiny(1.3)} & 73.8 {\tiny(1.2)} & 14.3 {\tiny(1.0)} & 51.6 {\tiny (0.1)} & 36.1 {\tiny (0.2)}\\
 & 1\% & 2.3 {\tiny(1.1)}  & 78.7  {\tiny(0.2)}  & 5.4 {\tiny(1.7)} & 75.9 {\tiny(0.2)} & 14.1 {\tiny(0.6)} & 51.8 {\tiny (0.2)} & 35.6 {\tiny (0.2)}\\
 \bottomrule
\end{tabular}
\end{table*}

%% file: gpt_experiments.tex
\ifthenelse{\boolean{arxiv}}
{
\subsection{Experiments with auto-regressive language modeling}
\label{sec:}

Given the tremendous recent success of auto-regressive models~\citep{brown2020language}, we fine-tune a pre-trained GPT-2 (small) model \citep{radford2019language} on GLUE tasks using prompts from \citet{holtzman2021surface}. 
Firstly, we find that skill localization is possible for GPT-2 based fine-tuning as well, albeit requiring denser regions with $0.05$\% parameters as opposed to $0.01$\% required by the RoBERTa model; \Cref{tab:gpt_patch_perf_64_4096} summarizes the results.
Overall the performance of GPT-2 fine-tuned models is worse than a similarly sized RoBERTa model, which is consistent with prior work. GPT-2 requiring denser regions and having worse generalization is consistent with our connection between sparsity and generalization from \Cref{sec:generalization}.
\input{table_fig/gpt_patch_single}


}{}

%% file: table_fig/gpt_patch_single.tex
\ifthenelse{\boolean{arxiv}}{

\begin{table*}[htbp]
\centering
\caption{\looseness-1 We fine-tune GPT-2 models (small) \cite{radford2019language} on different downstream tasks using prompts from \cite{holtzman2021surface}. FT on GPT-2 (small) performs worse than RoBERTa-base in both 64-shot and 4096-shot settings. In fact, we observe that FT gives close to random guessing performance on most two sentence tasks in the 64-shot setting. For each downstream task, we learn a grafting region $\patch$ using our optimization procedure in \cref{eq:opt_patch}. The grafting regions for all tasks have sparsity at most $0.05\%$ ($<42500$ parameters).
We report test accuracy (${}^\dagger$F1) and the calibration error using ECE of the fine-tuned model and the grafted model for each task. The main findings are (1) The grafted model can retrieve $>95\%$ of the FT accuracy while being better calibrated than the original model itself.
For single-sentence tasks (4096-shot) the grafted model shows only a $1.2\%$ drop in accuracy but an improvement of $3.2\%$ in the calibration error. Similarly for two-sentence tasks, the grafted model shows a $3.4\%$ drop in accuracy with an improvement of $7.3\%$ in the calibration error. \vspace{0.1in}}
\label{tab:gpt_patch_perf_64_4096}
\footnotesize
\begin{tabular}{l|cc|cc|cc|cc|c} 
\toprule
  & \multicolumn{4}{c|}{$64$-shot} & \multicolumn{5}{c}{$4096$-shot}  \\
  \midrule
      & \multicolumn{2}{c|}{FT} & \multicolumn{2}{c|}{Graft} & \multicolumn{2}{c|}{FT} & \multicolumn{2}{c|}{Graft} & \\
  \midrule
 Dataset & Acc. & ECE & Acc. & ECE  & Acc. & ECE & Acc. & ECE & Agreement \\
 \midrule
 \multicolumn{10}{c}{Single sentence tasks} \\
 \midrule
 SST-2 & 85.0 {\tiny (0.9)} & 11.8 {\tiny (2.0)} & 84.8 {\tiny (0.8)} & 7.3 {\tiny (2.1)} & 90.6 {\tiny (0.6)} & 5.8 {\tiny (1.7)} & 88.8 {\tiny (0.5)} & 3.4 {\tiny (0.8)} & 93.6 {\tiny (0.9)}\\
CR & 87.4 {\tiny (1.0)} & 11.1 {\tiny (1.1)} & 87.5 {\tiny (0.8)} & 7.4 {\tiny (0.9)} & 87.3 {\tiny (1.5)} & 11.2 {\tiny (1.3)} & 88.5 {\tiny (0.6)} & 5.4 {\tiny (1.6)} & 92.8 {\tiny (1.6)}\\
MR & 80.2 {\tiny (2.0)} & 17.0 {\tiny (2.0)} & 80.5 {\tiny (1.6)} & 11.0 {\tiny (1.4)} & 86.7 {\tiny (0.2)} & 6.0 {\tiny (2.2)} & 86.0 {\tiny (0.4)} & 2.0 {\tiny (0.2)} & 91.6 {\tiny (1.0)}\\
MPQA & 85.2 {\tiny (1.2)} & 11.9 {\tiny (1.3)} & 85.5 {\tiny (0.5)} & 6.7 {\tiny (1.9)} & 88.4 {\tiny (0.6)} & 7.6 {\tiny (2.1)} & 88.1 {\tiny (0.2)} & 3.7 {\tiny (1.4)} & 93.7 {\tiny (1.3)}\\
AGNews & 87.2 {\tiny (0.8)} & 9.6 {\tiny (0.5)} & 84.3 {\tiny (0.8)} & 6.0 {\tiny (0.8)} & 92.2 {\tiny (0.4)} & 2.8 {\tiny (0.9)} & 88.6 {\tiny (0.9)} & 1.6 {\tiny (0.3)} & 91.9 {\tiny (0.7)}\\
Subj & 89.5 {\tiny (1.1)} & 7.5 {\tiny (0.4)} & 85.0 {\tiny (2.2)} & 2.8 {\tiny (1.3)} & 95.8 {\tiny (0.2)} & 3.3 {\tiny (0.3)} & 93.5 {\tiny (0.3)} & 1.5 {\tiny (0.6)} & 95.2 {\tiny (0.5)}\\
\midrule
 \multicolumn{10}{c}{Two sentence tasks} \\
 \midrule
 QNLI & 53.9 {\tiny (3.3)} & 30.5 {\tiny (13.0)} & 54.1 {\tiny (3.2)} & 18.4 {\tiny (7.3)} & 81.9 {\tiny (0.3)} & 6.1 {\tiny (4.8)} & 79.8 {\tiny (0.2)} & 2.5 {\tiny (0.9)} & 87.8 {\tiny (1.6)}\\
SNLI & 56.0 {\tiny (3.2)} & 33.4 {\tiny (4.3)} & 50.4 {\tiny (3.1)} & 23.3 {\tiny (4.4)} & 80.7 {\tiny (0.6)} & 8.2 {\tiny (2.7)} & 76.3 {\tiny (0.5)} & 1.7 {\tiny (0.7)} & 84.0 {\tiny (1.2)}\\
MNLI & 45.1 {\tiny (3.5)} & 41.3 {\tiny (2.1)} & 42.2 {\tiny (5.0)} & 30.7 {\tiny (3.1)} & 72.3 {\tiny (0.8)} & 12.2 {\tiny (4.5)} & 67.6 {\tiny (0.6)} & 4.3 {\tiny (2.6)} & 80.1 {\tiny (1.2)}\\
RTE & 51.8 {\tiny (3.5)} & 36.8 {\tiny (4.5)} & 51.9 {\tiny (2.9)} & 23.8 {\tiny (2.5)} & 66.7 {\tiny (1.5)} & 27.3 {\tiny (2.3)} & 65.7 {\tiny (1.5)} & 16.8 {\tiny (1.8)} & 79.6 {\tiny (2.9)}\\
MRPC$^{\dagger}$ & 81.1 {\tiny (0.2)} & 13.2 {\tiny (7.5)} & 78.5 {\tiny (1.7)} & 6.3 {\tiny (3.1)} & 85.2 {\tiny (0.5)} & 15.8 {\tiny (3.5)} & 80.3 {\tiny (1.5)} & 8.3 {\tiny (2.2)} & 82.9 {\tiny (2.2)}\\
QQP$^{\dagger}$ & 56.8 {\tiny (0.7)} & 25.6 {\tiny (7.9)} & 54.6 {\tiny (1.8)} & 16.1 {\tiny (5.9)} & 76.1 {\tiny (0.2)} & 13.5 {\tiny (2.4)} & 73.0 {\tiny (0.5)} & 5.9 {\tiny (1.9)} & 87.4 {\tiny (1.1)}\\
 \bottomrule
\end{tabular}
\end{table*}

}{}

%% file: multi_task_appendix.tex
\section{Multi-task and Continual Learning}

\subsection{Training details for multi-task training.}

Each dataset was downsampled to $4096$-shot and SGD was used to fine-tune the model on all the tasks (with batches being picked from a randomly chosen dataset), but with $8$ times as many iterations. Using same hyperparameter grid as before, we select the model that performs the best on average on the validation sets of each of these tasks.

\subsection{Additional Multi-task Experiments}
\input{table_fig/MT_appendix.tex}

\textbf{Varying $\basepatch$.} We conduct further experiments on SGD trained MT models in \cref{tab:MT_expts_4096-shot_8tasks_1e-5} and \cref{tab:MT_expts_4096-shot_8tasks_1e-4}. In these experiments, we vary the initialization $\basepatch$ for the task specific graft regions in \cref{eq:opt_patch}. Increasing the size of the $\basepatch$ leads to increasing overlap among the task-specific grafts. However, we still observe a similar pattern, albeit dimishing, as \cref{tab:MT_expts_4096-shot_8tasks_mainpaper} for the effect of the task specific grafts on other tasks.

\textbf{AdamW optimizer.} We also conduct experiments on MT models trained with AdamW in \cref{tab:MT_expts_4096-shot_8tasks_1e-7_Adam} and \cref{tab:MT_expts_4096-shot_8tasks_1e-7_Adam_l1}. Interestingly, we observe skill localization in the model, without an explicit use of $\ell_1$ regularization. This is in stark contrast to single task models (\cref{fig:adam_sgd_qnli,fig:adam_sgd_sst-2}), where an explicit $\ell_1$ regularization was necessary to get sparse grafts.

\subsection{Details for Continual Learning}
\label{sec:apx_continual}

Formally, let task $t_n$ represent the task arrives at order $n$ in the sequence of tasks $t_1, t_2, \cdots, t_n, \cdots$. Suppose we have identified $\patch_{t_n}$ as the grafting region for the task $t_n$ from an independent run of the model on the task $t_n$. Then, when $t_n$ arrives, we train the parameters of the model in the region $\patch_{t_n} \setminus \cup_{i < n} \patch_{t_i}.$ When we are asked to infer for task $t_n$, we construct the grafted model using $\cup_{i \le n} \patch_{t_i}$.

As a simple experiment, we train a model on QNLI, AG news, and SST-2, with each dataset represented by  a 4096-shot training set, in sequential order. After the model has been trained on SST-2, we check the model's performance on QNLI and AG news. We compare the performance of the model when trained with the SGD algorithm naively, and when it is trained with our continual learning algorithm  in \cref{tab:continual}. We observe that our proposed  continual method can retain the performance on QNLI, while having a minimal effect on the performance of other tasks.

The main advantage of our proposed method is that we can retain performance by memorizing the grafting region and the arrival sequence number for each task. This method does not lead to a blowup of the memory required, since each parameter in the memory can be indexed by the sequence number of the task it belongs to. However, the major disadvantage of this method is that we need to create the grafted model for a task when we need to infer for the particular task. Thus, an important question that we aim to explore for future work is whether we can use the concept of grafting to get a single model that does well in continual learning.

%% file: table_fig/MT_appendix.tex
\begin{figure*}[!t]
\centering
\begin{subfigure}{0.49\textwidth}
\centering
\includegraphics[width=\textwidth]{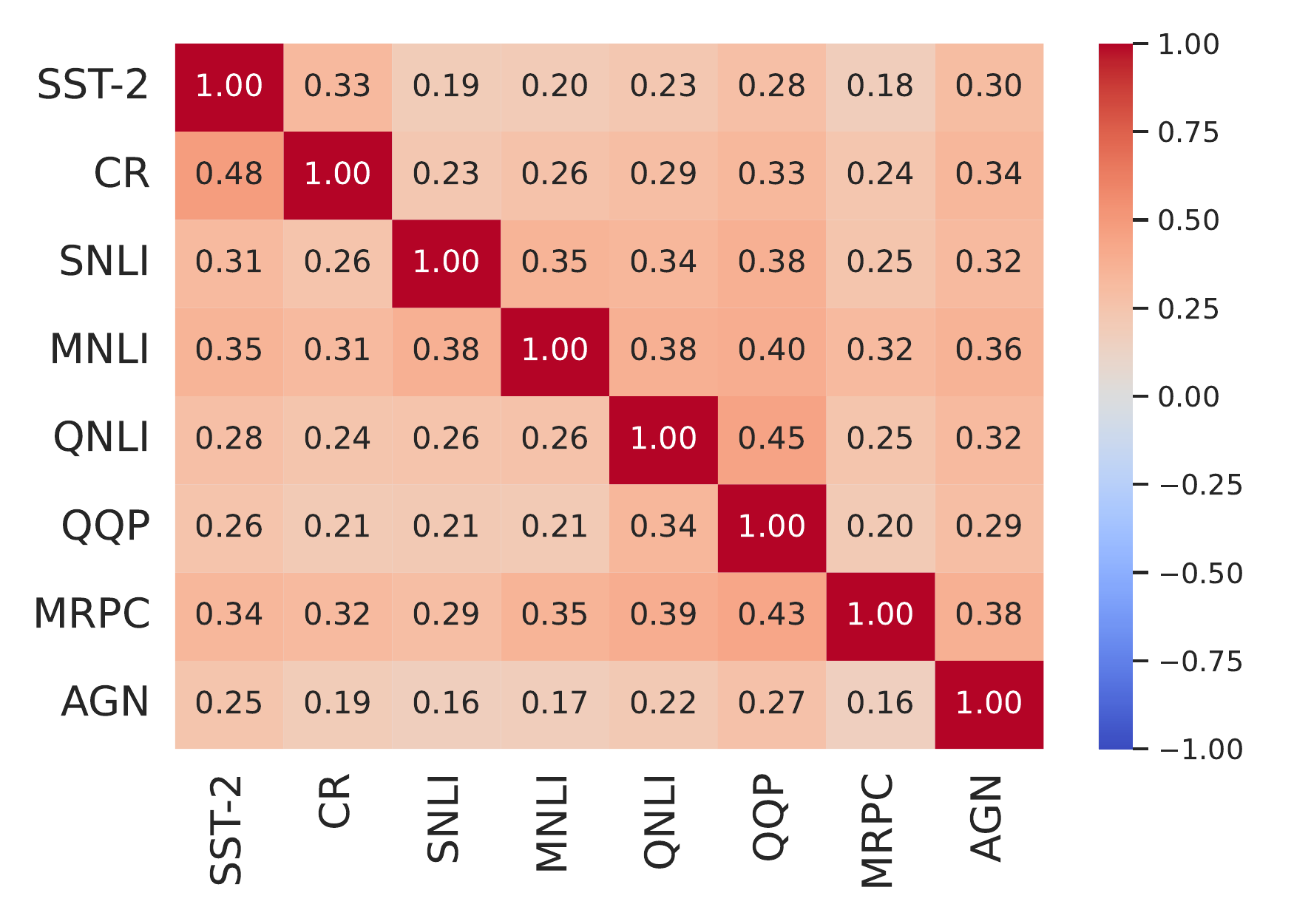}\caption{Overlap of task-specific grafting regions}
\end{subfigure}\hfill
\begin{subfigure}{0.49\textwidth}
\centering
\includegraphics[width=\textwidth]{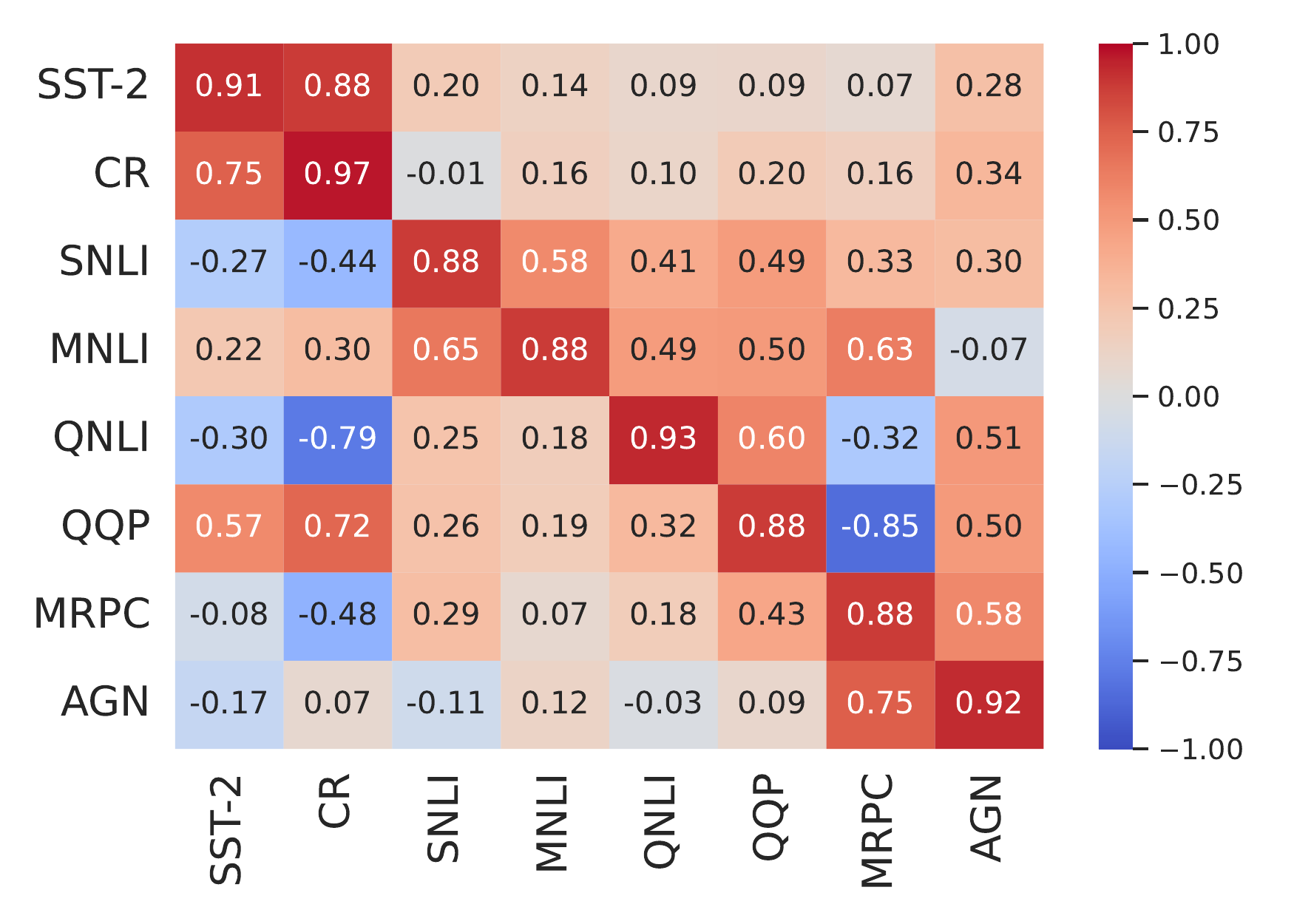}\caption{ Effect of task-specific grafted model on other tasks}
\end{subfigure}
\caption{Ablations on task-specific grafting region $\patch_{i}$ for task $i$, learned by optimizing $\loss_{i}$ on the MT model, trained with SGD. Here, we learn the task specific grafting regions $\patch_\task$ by solving $\loss_\task$ on the MT model with $\basepatch$ set to the top-$10^{-5}$ moving parameters in the model.
\ifthenelse{\boolean{arxiv}}{
In all figures, we evaluate the effect of graft region of task in row $i$ on the task in row $j$.
Figure (a) measures the assymteric overlap in the regions defined as $\frac{|\patch_i \cap \patch_j|}{|\patch_j|}$ for tasks in row $i$ and column $j$.
Figure (b) evaluates the relative accuracy gain of task in column $j$ using the graft regions of task in row $i$; refer to \Cref{eq:rel_perf_gain} for the precise expression.
\textbf{Observations:}}{}
We have similar observations as \cref{tab:MT_expts_4096-shot_8tasks_mainpaper}. However, we have an increasing amount of intersection between the task-specific graftings, as $\basepatch$ is common among all the tasks for optimization.}
\label{tab:MT_expts_4096-shot_8tasks_1e-5}
\end{figure*}

\begin{figure*}[!t]
\centering
\begin{subfigure}{0.49\textwidth}
\centering
\includegraphics[width=\textwidth]{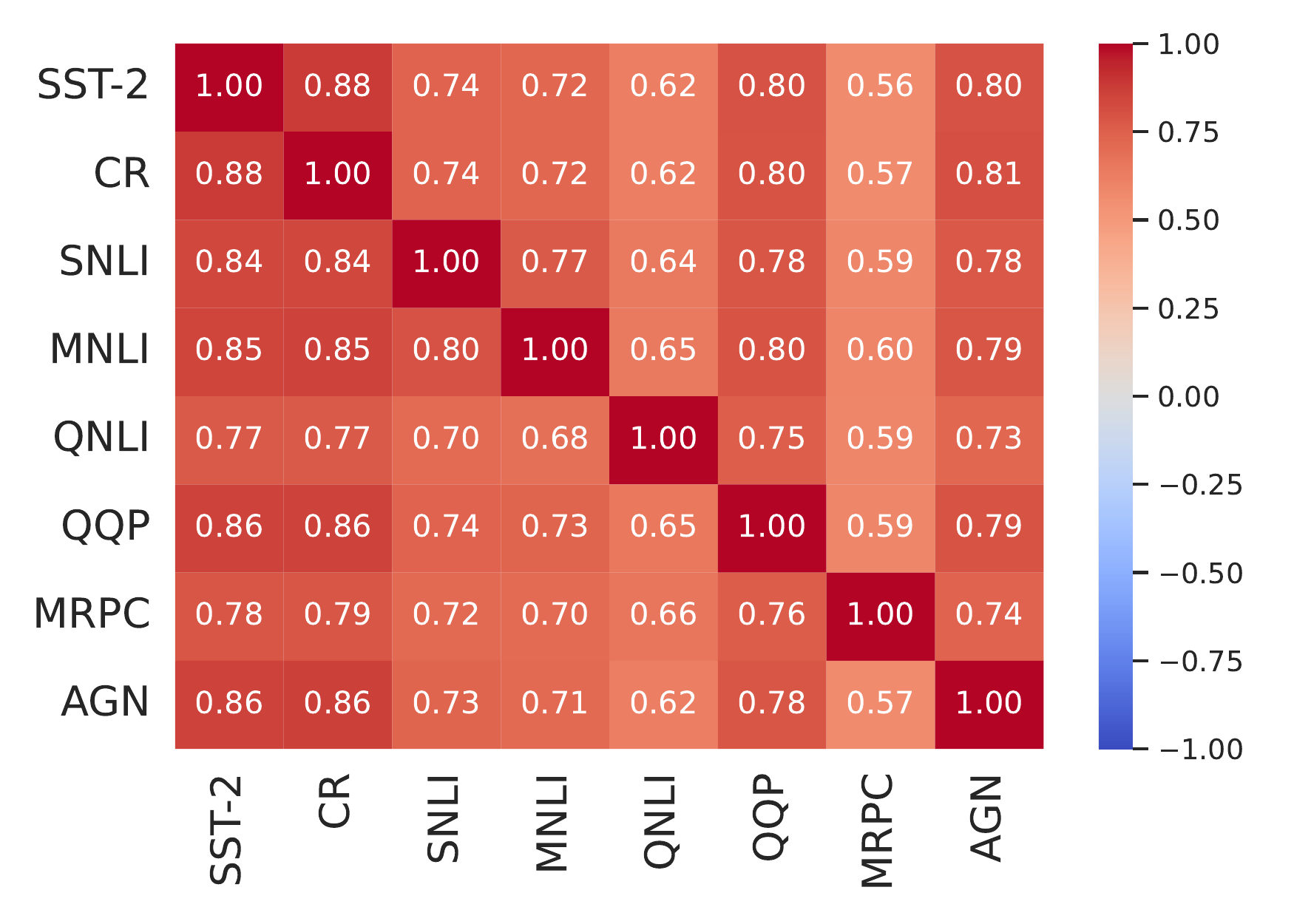}\caption{Overlap of task-specific grafting regions}
\end{subfigure}\hfill
\begin{subfigure}{0.49\textwidth}
\centering
\includegraphics[width=\textwidth]{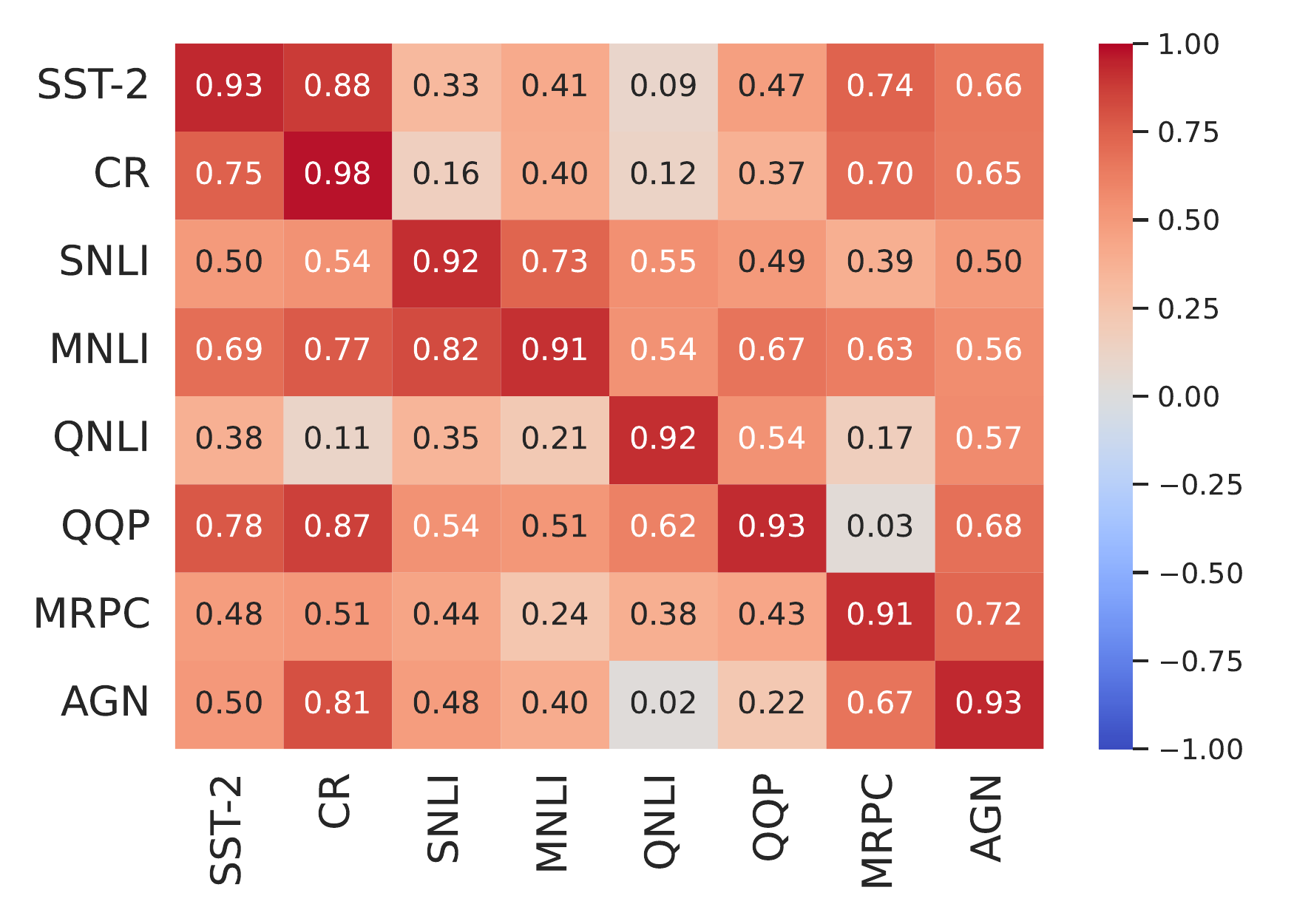}\caption{ Effect of task-specific grafted model on other tasks}
\end{subfigure}
\caption{Ablations on task-specific grafting region $\patch_{i}$ for task $i$, learned by optimizing $\loss_{i}$ on the MT model, trained with SGD. Here, we learn the task specific grafting regions $\patch_\task$ by solving $\loss_\task$ on the MT model with $\basepatch$ set to the top-$10^{-4}$ moving parameters in the model. 
\ifthenelse{\boolean{arxiv}}{
In all figures, we evaluate the effect of graft region of task in row $i$ on the task in row $j$.
Figure (a) measures the assymteric overlap in the regions defined as $\frac{|\patch_i \cap \patch_j|}{|\patch_j|}$ for tasks in row $i$ and column $j$.
Figure (b) evaluates the relative accuracy gain of task in column $j$ using the graft regions of task in row $i$; refer to \Cref{eq:rel_perf_gain} for the precise expression.
\textbf{Observations:}}{}
We have similar observations as \cref{tab:MT_expts_4096-shot_8tasks_mainpaper}. However, we have an increasing amount of intersection between the task-specific graftings, as $\basepatch$ is common among all the tasks for optimization.}
\label{tab:MT_expts_4096-shot_8tasks_1e-4}
\end{figure*}

\ifthenelse{\boolean{arxiv}}{
\begin{figure*}[!t]
\centering
\begin{subfigure}{0.49\textwidth}
\centering
\includegraphics[width=\textwidth]{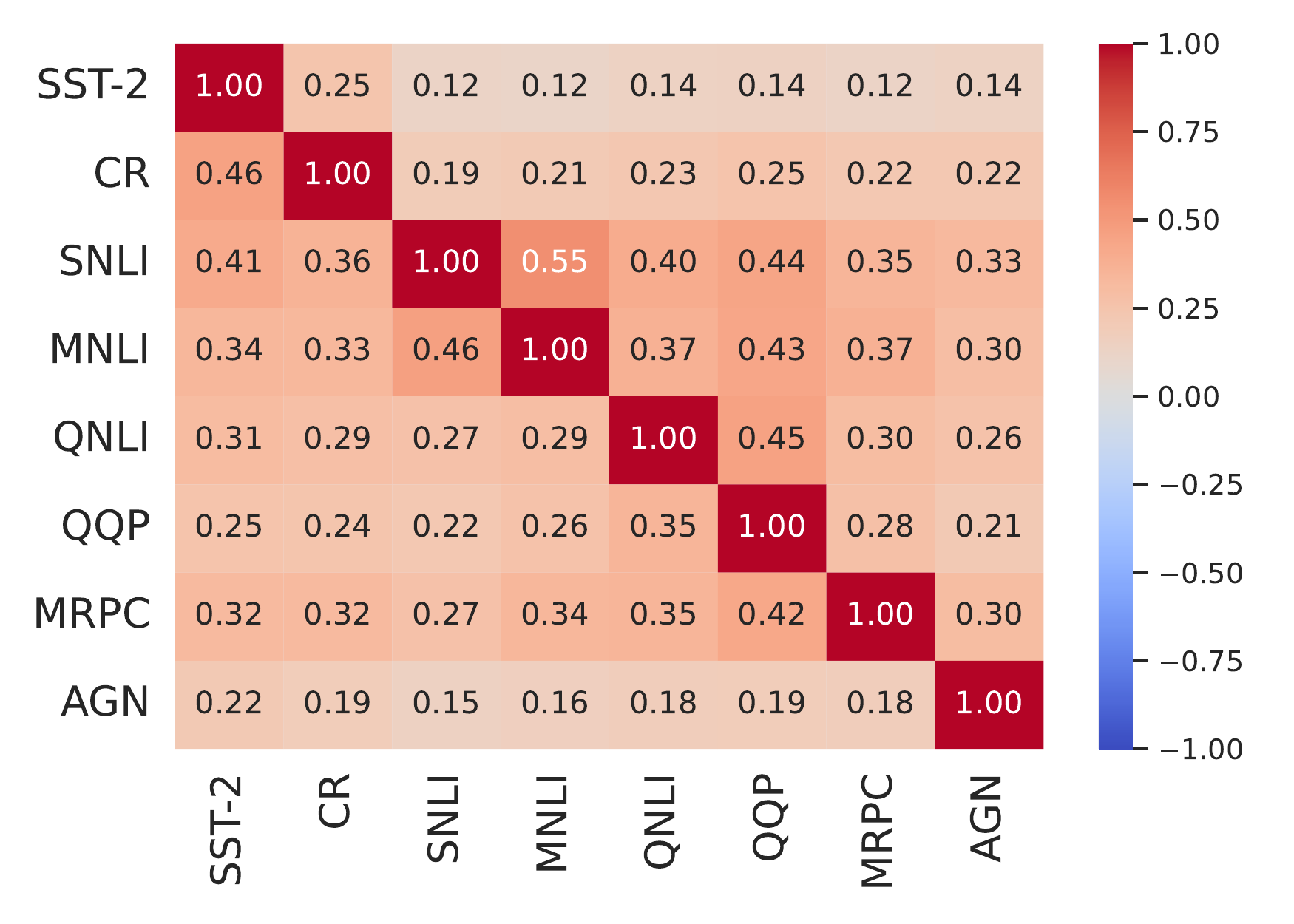}\caption{Overlap of task-specific grafting regions}
\end{subfigure}\hfill
\begin{subfigure}{0.49\textwidth}
\centering
\includegraphics[width=\textwidth]{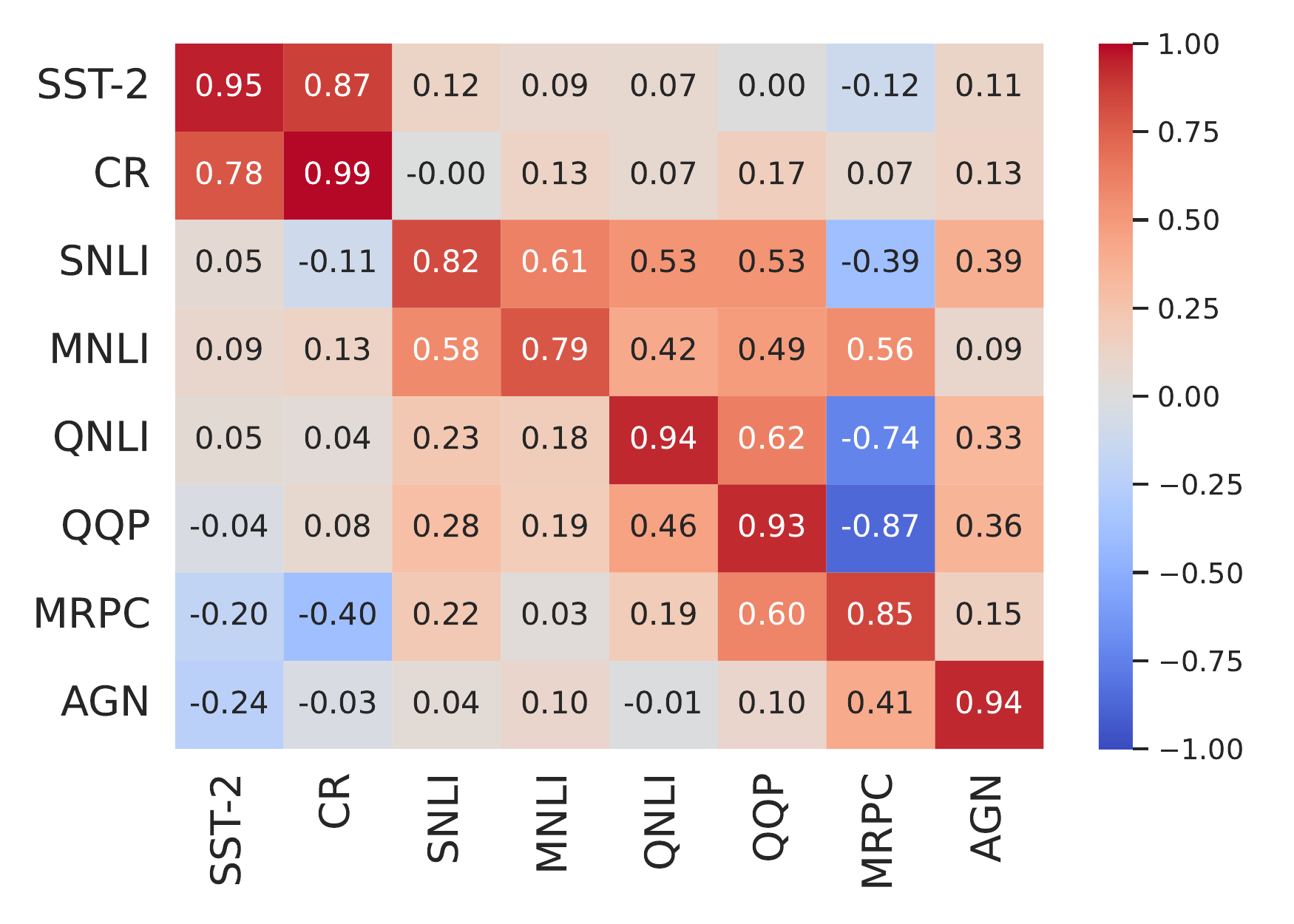}
\caption{ Effect of task-specific grafted model on other tasks}
\end{subfigure}
\caption{Ablations on task-specific grafting region $\patch_{i}$ for task $i$, learned by optimizing $\loss_{i}$ on the MT model, trained with AdamW. Here, we learn the task specific grafting regions $\patch_\task$ by solving $\loss_\task$ on the MT model with $\basepatch$ set to $0$. In all figures, we evaluate the effect of graft region of task in row $i$ on the task in row $j$.
Figure (a) measures the assymteric overlap in the regions defined as $\frac{|\patch_i \cap \patch_j|}{|\patch_j|}$ for tasks in row $i$ and column $j$.
Figure (b) evaluates the relative accuracy gain of task in column $j$ using the graft regions of task in row $i$; refer to \Cref{eq:rel_perf_gain} for the precise expression.
\textbf{Observations:}
Interestingly, we observe that AdamW shows better skill localization in the MT model, as opposed to task-specific models (\cref{fig:adam_sgd_qnli,fig:adam_sgd_sst-2}). }
\label{tab:MT_expts_4096-shot_8tasks_1e-7_Adam}
\end{figure*}

\begin{figure*}[!t]
\centering
\begin{subfigure}{0.49\textwidth}
\centering
\includegraphics[width=\textwidth]{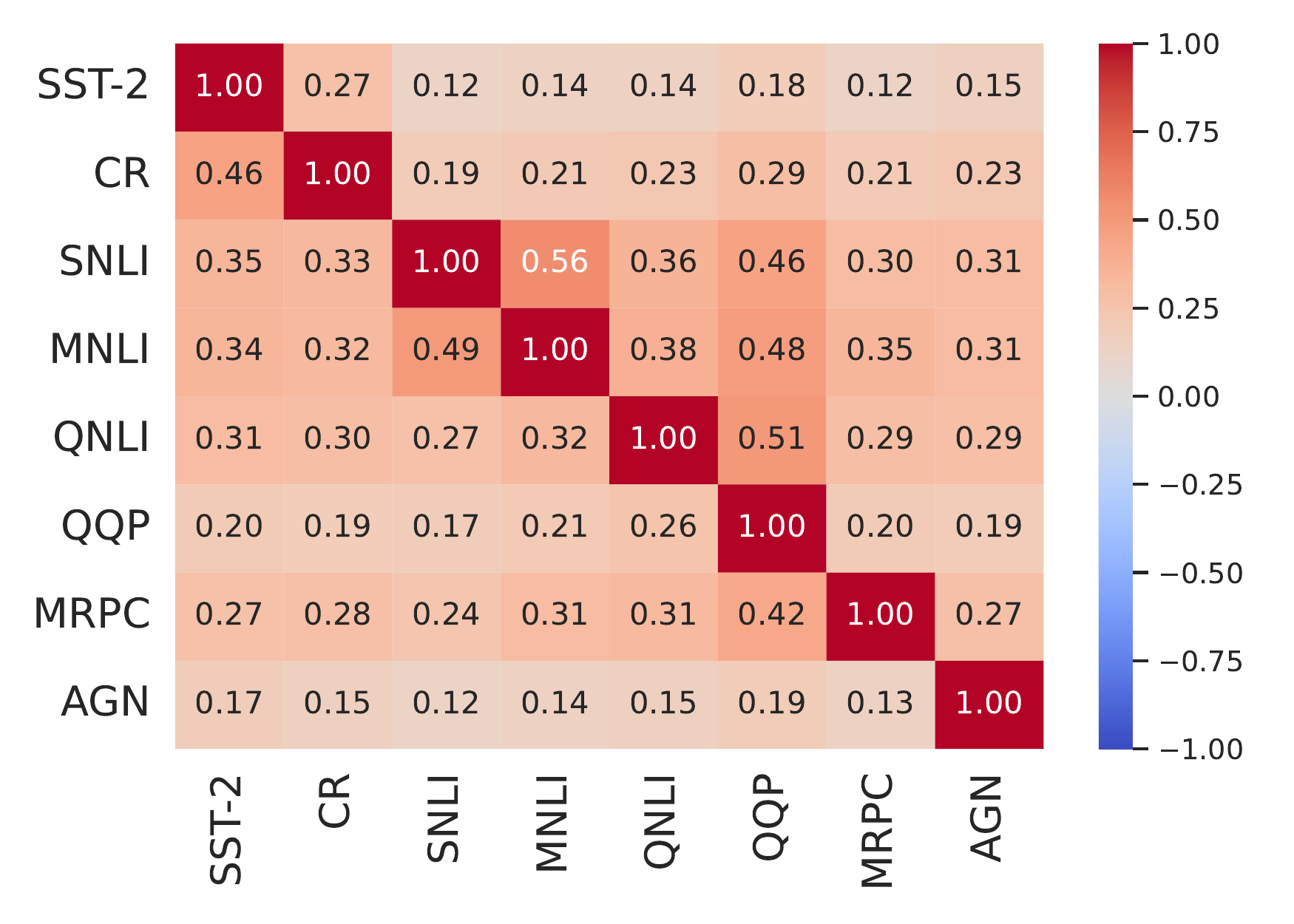}\caption{Overlap of task-specific grafting regions}
\end{subfigure}\hfill
\begin{subfigure}{0.49\textwidth}
\centering
\includegraphics[width=\textwidth]{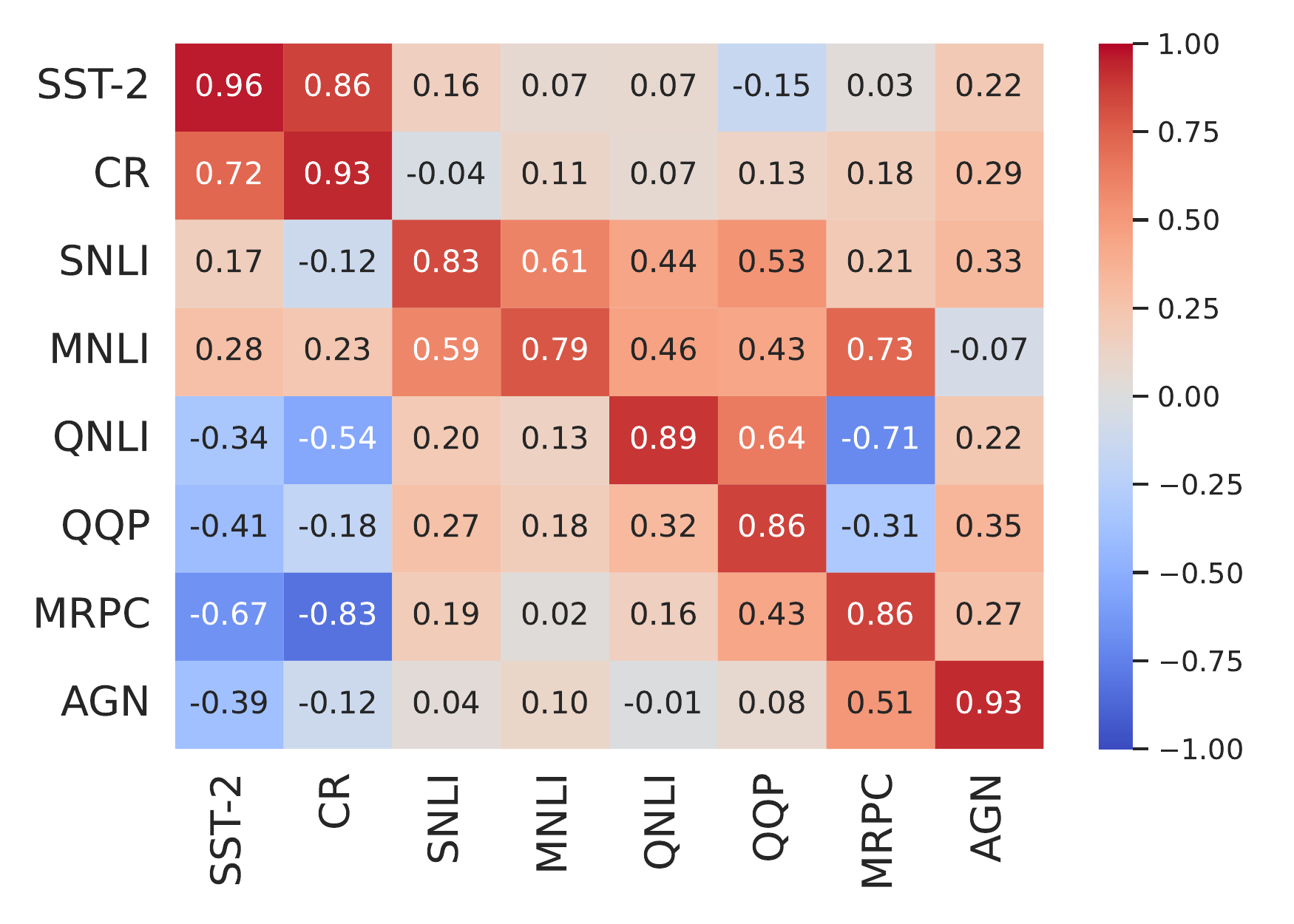}\caption{ Effect of task-specific grafted model on other tasks}
\end{subfigure}
\caption{Ablations on task-specific grafting region $\patch_{i}$ for task $i$, learned by optimizing $\loss_{i}$ on the MT model, trained with AdamW with an explicit $0.001$-$\ell_1$ regularization on the movement of the parameters during fine-tuning. Here, we learn the task specific grafting regions $\patch_\task$ by solving $\loss_\task$ on the MT model with $\basepatch$ set to $0$.
\ifthenelse{\boolean{arxiv}}{
In all figures, we evaluate the effect of graft region of task in row $i$ on the task in row $j$.
Figure (a) measures the assymteric overlap in the regions defined as $\frac{|\patch_i \cap \patch_j|}{|\patch_j|}$ for tasks in row $i$ and column $j$.
Figure (b) evaluates the relative accuracy gain of task in column $j$ using the graft regions of task in row $i$; refer to \Cref{eq:rel_perf_gain} for the precise expression.
\textbf{Observations:}}{}
We find that adding an explicit $\ell_1$ regularization doesn't affect or improve the skill localization of AdamW in the MT model, observed in \cref{tab:MT_expts_4096-shot_8tasks_1e-7_Adam}. }
\label{tab:MT_expts_4096-shot_8tasks_1e-7_Adam_l1}
\end{figure*}

}{}

%% file: generalization_theory.tex
\section{Discussions on Generalization Theory}\label{app:generalization}

\looseness-1 
Classical tools like uniform convergence (UC) seemingly fail to explain generalization in deep learning \citep{nagarajan2019uniform} since they bound the generalization error\footnote{This is a rough picture just to present the high-level idea.} as
\begin{align*}
    \testloss(\paramft) &\lesssim \inf_{\param^{\star}\in\paramclass} \testloss(\param^{\star}) + \sup_{\param\in\paramclass} \left|\testloss(\param) - \trainloss(\param) \right|
\end{align*}
\looseness-1The second term is further upper bounded by $\frac{\gC\left(\paramclass\right)}{\sqrt{n}}$ where $\gC(\paramclass)$ denotes a complexity measure for the class (e.g. Rademacher complexity).
This approach necessarily fails for vanilla fine-tuning because the class $\paramclass$ is very large (100M of parameters) and expressive enough to perfectly fit the training set with a large train-test gap, thus giving vacuous bounds -- fine-tuning on QNLI in the 4096-shot setting get $100$\% training and $\sim88$\% test accuracy, making the second term $\ge12$\%..

One way to get around this vacuous-ness of UC is to use properties of the training algorithm (here fine-tuning) to transform the learned model into a ``surrogate model'' that belongs to a much simpler class of functions, for which uniform convergence is more like to succeed, e.g. low rank compressions \citep{arora2018stronger}, denoising classifiers \citep{negrea2020defense}.
The existence of sparse graft models with fine-tuning precisely gives us a surrogate model class that allows for a tighter UC-based bound instead.
For a grafting region $\patch$ the relevant UC bound will looks as follows:
\begin{align*}
    \sup_{\param\in\paramclass} \left|\testloss(\patchparamfn{\param}(\patch)) - \testloss(\patchparamfn{\param}(\patch))\right|
\end{align*}
The sparsity of the region $\patch$ can be further leveraged to upper bound this UC term by $\tilde{\gO}\left(\sqrt{\frac{s}{n}}\right)$ using standard arguments, where $s = \|\patch\|_{0}$, $n$ is the training set size and $\tilde{\gO}$ hides logarithmic factors. Besides the theoretical benefit of the sparsity, we also empirically find that the train-test gap of the graft model using just $\sim5000$ parameters is small ($\sim1$\%) since a sparse graft model is unable to perfectly fit the training sets even in the $4096$-shot setting.
Crucially this small train-test gap comes at a very small impact on the expressivity, i.e. there exist sparse grafting models that can do very well on the task.
Leveraging these observations, we show a generalization bound for a sparse re-training based procedure inspired by model grafting.
In fact, our results hold generally for any parameter efficient fine-tuning method that only updates a small subset of parameters, e.g. BitFit \citep{ben2022bitfit}.

\subsection{Generalization Bound for Graft Re-training and Parameter-Efficient FT}

We first clarify some notations. Let $\patchparam(\patch)$ denote the graft model learned after the fine-tuning procedure, $\param(\patch)$ to denote the model re-trained on $n$ samples by only updating the parameters in $\patch$, $\param^*(\patch)$ to denote the model with best test performance while only changing the parameters in $\patch$, and $\param^*$ to denote the model with the best test performance. For simplicity, we also assume that the loss $\gL$ is bounded by $[0,1]$. To bound the complexity of the function class, we assume that each parameter can only take the value from $q$ numbers, and the quantized network has performance close to the original network. This assumption is very close to the definition of compressible classifier in \citet{arora2018stronger}. Formally, we have
\begin{assumption}\label{ass:quantization}
    For any model $\param$, we define $\bar q(\param)$ to be the model that quantized every parameter into the $q$ given values. Then there exist $\varepsilon_1 > 0$ such that for all $\param$ and any data point $x$, we have
    \[|\gL(x;\bar q(\param)) - \gL(x;\param)| \le \varepsilon_1.\]
\end{assumption}
\textit{\textbf{Remark.}} We only need this condition for quantization of the parameters in $\patch$ and not necessarily all parameters. Furthermore, fixed precision floating point numbers are anyway quantized in practice, with $q=2^{32}$ for $32$-bit floats.

We denote $\Theta_n$ to be the number of different regions $\patch$ that can be found from \Cref{eq:opt_patch} when trained on different training datasets from the same task.
While $\Theta_n$ can be trivially bounded by $d^s$ where $d$ is the parameters in the model and $s = \|\patch\|_0$, empirically $\Theta_n$ should be small ($\log \Theta_n \ll s$) since the regions $\patch$ we find for different training sets of the same task have large overlap.
Additionally, for the case of parameter-efficient fine-tuning algorithms where a fixed set of parameters are chosen to be updated (e.g. BitFit~\citep{ben2022bitfit}), this term $\Theta_n$ is $1$.

Then we have the following theorem for the generalization bound of the re-trained graft model $\param(\patch)$ on $n$ samples. 

\begin{theorem}\label{thm:generalization-retrained-formal}
Under \Cref{ass:quantization}, we have
    \begin{align*}
        &\testloss(\param(\patch)) - \testloss(\param^*) \\&\le \underbrace{2\sup_{\tilde\param}\left|\testloss(\bar q(\tilde\param(\patch))) - \trainloss(\bar q(\tilde\param(\patch))) \right|}_{\text{variance term}} \\& + \underbrace{\testloss(\param^*(\patch)) - \testloss(\param^*)}_{\text{bias term}} + \underbrace{4\varepsilon_1}_{\text{quantization error}}.
    \end{align*}
    Moreover, with probability at least $1-\delta$, the variance term can be bounded by
    \begin{align*}
    &
    2\sup_{\tilde\param}\left|\testloss(\bar q(\tilde\param(\patch))) - \trainloss(\bar q(\tilde\param(\patch)))\right| \\& \le 2\sqrt{\frac{s\log q + \log \Theta_n + \log(1/\delta)}{n}}.\end{align*}
\end{theorem}

The first term captures the variance of the re-trained graft model, which can be further bounded by $\tilde O(\sqrt{s/n})$ in our setting using uniform convergence over all possible grafting regions $\patch$, since the total number of the grafted region can be bounded as $\Theta_n \le d^s$.
Thus the sparser the graft region, the more sample efficient fine-tuning will be.
Empirically, because the (re-trained) grafting model has a very small train-test gap in practice, we know that the term $\testloss(\param(\patch)) - \trainloss(\param(\patch))$ is small. Note that $\param(\patch)$ is the re-trained model on the grafted region $\patch$ using the training data, so so one would expect the model $\param(\patch)$ to have nearly the largest train-test gap compared to all grafted models. This is because, intuitively, the solution found by ERM should not have better generalization compared to other solutions.
This leads us to believe that for all models on grafted regions $\tilde\param(\patch)$, the train-test gap is also small, which means that the variance term $\sup_{\tilde\param}\left|\testloss(\bar q(\tilde\param(\patch))) - \trainloss(\bar q(\tilde\param(\patch))) \right|$ should be small empirically.
In practice, the train-test gap for re-trained models is of the order of $1$\%.

The second term $(\testloss(\param^*(\patch)) - \testloss(\param^*))$ actually characterizes the capacity of the grafted region $\patch$. If we let the sparsity level goes to $1$ (the full model), this term becomes $0$ since it is as expressive as the whole model, and if the graft $\patch$ is too sparse, this term may be large.
Thus if there exists a good localized way to do well on the given task, we expect this term to be small.

\textit{\textbf{Remark.}} It is important to note that setting $\patch$ to be the all ones vector in the above result, i.e. including all parameters, will reduce to the naive uniform convergence bound for fine-tuning the entire network.
In this case $s=d$ (all parameters), and thus the bias term is exactly 0. However the variance term, in practice and in theory, is far from being small because fine-tuning the entire network can fit the training set perfectly, leading to the variance term being at least as large as $\sim10$\%.
This shows the utility of using model grafting as a "surrogate" to show much tighter generalization guarantees.

\Cref{thm:generalization-retrained-formal} gives an end-to-end bound for the re-training of the grafted model $\param(\patch)$, and also provides us with the freedom to balance the bias and variance.
In fact this bound also works for any parameter-efficient fine-tuning regime like BitFit.
However the sparsity level of model grafting is much lower than previous parameter efficient methods.

\input{table_fig/Track_stitch.tex}
\subsection{Generalization Bound for Grafted Models}

The previous discussion works for re-training a graft model $\param(\patch)$. In order to get a generalization bound for the graft model $\patchparam(\patch)$ without re-training (the graft model obtained directly after the fine-tuning), we need to add the following assumption.

\begin{assumption}\label{ass:nearly-erm}
    We assume that the graft model $\patchparam(\patch)$ is nearly ERM on $n$ training samples among all grafted models, i.e., there exist $\varepsilon_{2} > 0$ such that
    \[\trainloss(\patchparam(\patch)) \le \trainloss(\param(\patch)) + \varepsilon_2.\]
\end{assumption}

Although this assumption seems a little bit strong at first glance, it is actually validated by our empirical findings. 
If we further have \Cref{ass:nearly-erm}, we can directly have the generalization bound for the grafted model $\patchparam(\patch)$ from \Cref{thm:generalization-retrained-formal}.

\begin{theorem}\label{thm:generalization-graft-formal}
    Under \Cref{ass:quantization} and \Cref{ass:nearly-erm}, we have
    \begin{align*}
        &\testloss(\patchparam(\patch)) - \testloss(\param^*) \\&\le \underbrace{2\sup_{\tilde\param}\left|\testloss(\bar q(\tilde\param(\patch))) - \trainloss(\bar q(\tilde\param(\patch))) \right|}_{\text{variance term}} \\& + \underbrace{\testloss(\param^*(\patch)) - \testloss(\param^*)}_{\text{bias term}} + \underbrace{4\varepsilon_1 + \varepsilon_2}_{\text{error term}}.
    \end{align*}
    Moreover, with probability at least $1-\delta$, the first term can be bounded by
    \begin{align*}
        &2\sup_{\tilde\param}\left|\testloss(\bar q(\tilde\param(\patch))) - \trainloss(\bar q(\tilde\param(\patch))) \right| \\&\le 2\sqrt{\frac{s\log q + \log \Theta_n + \log(2/\delta)}{n}}.
    \end{align*}
\end{theorem}

\subsection{Proof of \Cref{thm:generalization-retrained-formal}}

First, we can decompose the term we interested $\testloss(\param(\patch)) - \testloss(\param^*)$ into seven terms.

\begin{align*}
    & \testloss(\param(\patch)) - \testloss(\param^*)\\ &\le (\testloss(\param(\patch)) - \testloss(\bar q(\param(\patch)))) \\
    & +\left(\testloss(\bar q(\param(\patch))) - \trainloss(\bar q(\param(\patch)))\right) \\
    & + (\trainloss(\bar q(\param(\patch))) - \trainloss(\param(\patch)))\\
    &+ (\trainloss(\param(\patch)) - \trainloss(\param^*(\patch)))\\
    & + (\trainloss(\param^*(\patch)) - \trainloss(\bar q(\param^*(\patch)))) \\
    &  + (\trainloss(\bar q(\param^*(\patch))) - \testloss(\bar q(\param^*(\patch)))) \\
    &  + (\testloss(\bar q(\param^*(\patch))) - \testloss(\param^*)).
\end{align*}

Now we bound each term one by one. For the first term $\testloss(\param(\patch)) - \testloss(\bar q(\param(\patch)))$, it is bounded by $\varepsilon_1$ because of \Cref{ass:quantization}. Similarly, the third term $\trainloss(\bar q(\param(\patch))) - \trainloss(\param(\patch))$ is also bounded by $\varepsilon_1$. As for the fourth term $\trainloss(\param(\patch)) - \trainloss(\param^*(\patch))$, it is at most $0$ since $\param(\patch)$ is the re-trained model by only updating the parameters in the region $\patch$ on the samples. As for the fifth term $(\trainloss(\param^*(\patch)) - \trainloss(\bar q(\param^*(\patch))))$ and the seventh term $(\testloss(\bar q(\param^*(\patch))) - \testloss(\param^*))$, they are also bounded by $\varepsilon_1$ according to \Cref{ass:quantization}.

For the second term $\left(\testloss(\bar q(\param(\patch))) - \trainloss(\bar q(\param(\patch)))\right)$ and the sixth term $(\trainloss(\bar q(\param^*(\patch))) - \testloss(\bar q(\param^*(\patch))))$, both of them can be bounded by
\[\sup_{\tilde\param}\left|\testloss(\bar q(\tilde\param(\patch))) - \trainloss(\bar q(\tilde\param(\patch))) \right|.\]

Thus, we show that under \Cref{ass:quantization}, we have
    \begin{align*}
       & \testloss(\patchparam(\patch)) - \testloss(\param^*) \\&\le \underbrace{2\sup_{\tilde\param}\left|\testloss(\bar q(\tilde\param(\patch))) - \trainloss(\bar q(\tilde\param(\patch))) \right|}_{\text{variance term}} \\&+ \underbrace{\testloss(\param^*(\patch)) - \testloss(\param^*)}_{\text{bias term}} + \underbrace{4\varepsilon_1 + \varepsilon_2}_{\text{error term}}.
    \end{align*}
Now we bound the variance term. Note that this can be bounded by standard generalization theory: because the number of possible models for $\bar q(\param(\patch))$ is bounded by $\Theta_n\cdot q^s$, after applying for the union bound, we have with probability at least $1-\delta$,
\begin{align*}
    &\sup_{\tilde\param}\left|\testloss(\bar q(\tilde\param(\patch))) - \trainloss(\bar q(\tilde\param(\patch))) \right| \\& \le \sqrt{\frac{s\log q + \log \Theta_n + \log(1/\delta)}{n}},
\end{align*}
and we complete the proof of \Cref{thm:generalization-retrained-formal}.

%% file: table_fig/Track_stitch.tex
\begin{figure*}[!htbp]
    \centering
    \begin{subfigure}{0.24\textwidth}
    \centering
    \includegraphics[width=\textwidth]{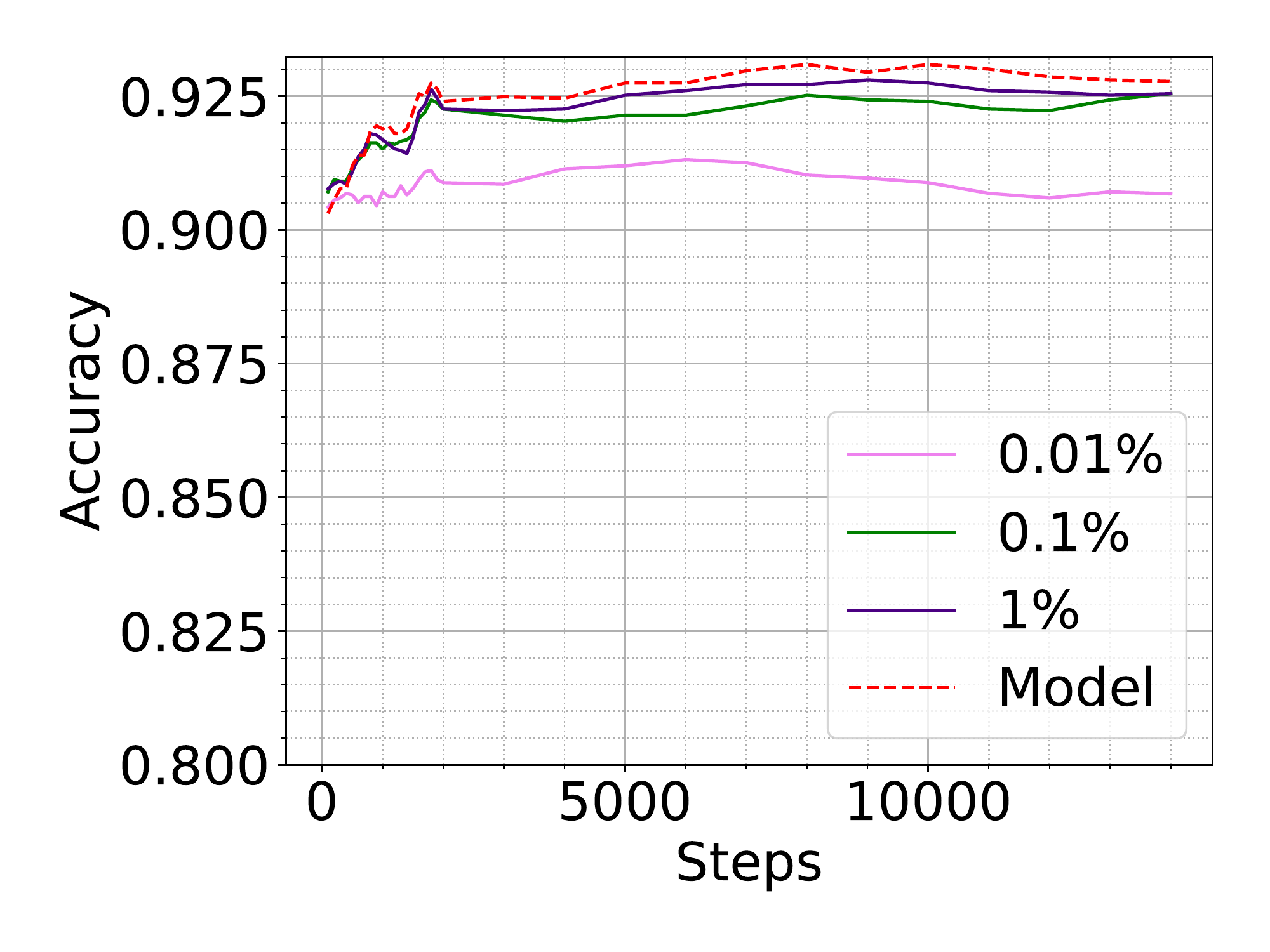}\caption {SST-2, 4096-shot. Checkpoint selected at step 500.}
    \end{subfigure}\hfill
    \begin{subfigure}{0.24\textwidth}
    \centering
    \includegraphics[width=\textwidth]{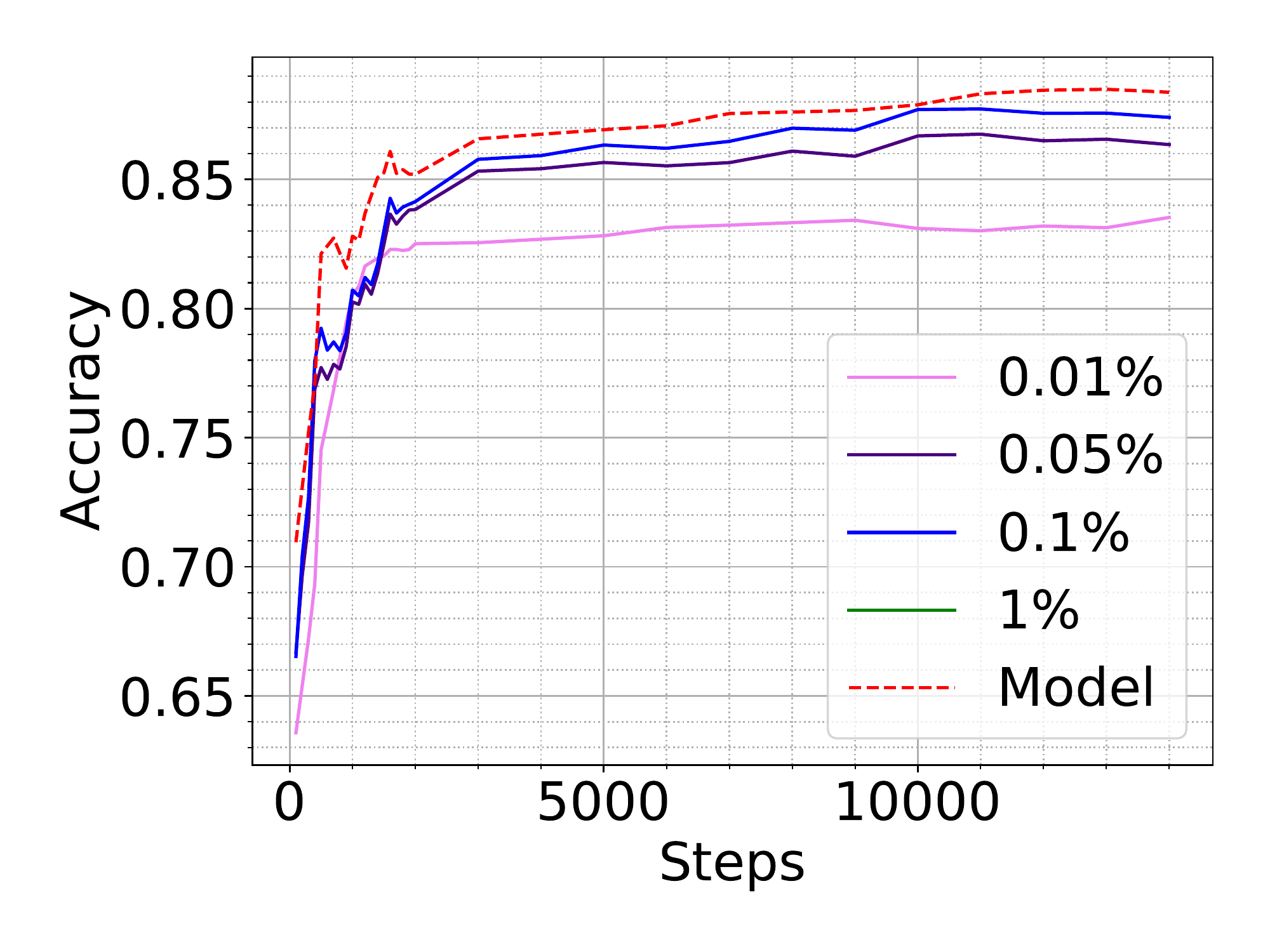}\caption { QNLI, 4096-shot. Checkpoint selected at step 2000.}
    \end{subfigure}\hfill
    \begin{subfigure}{0.24\textwidth}
    \centering
    \includegraphics[width=\textwidth]{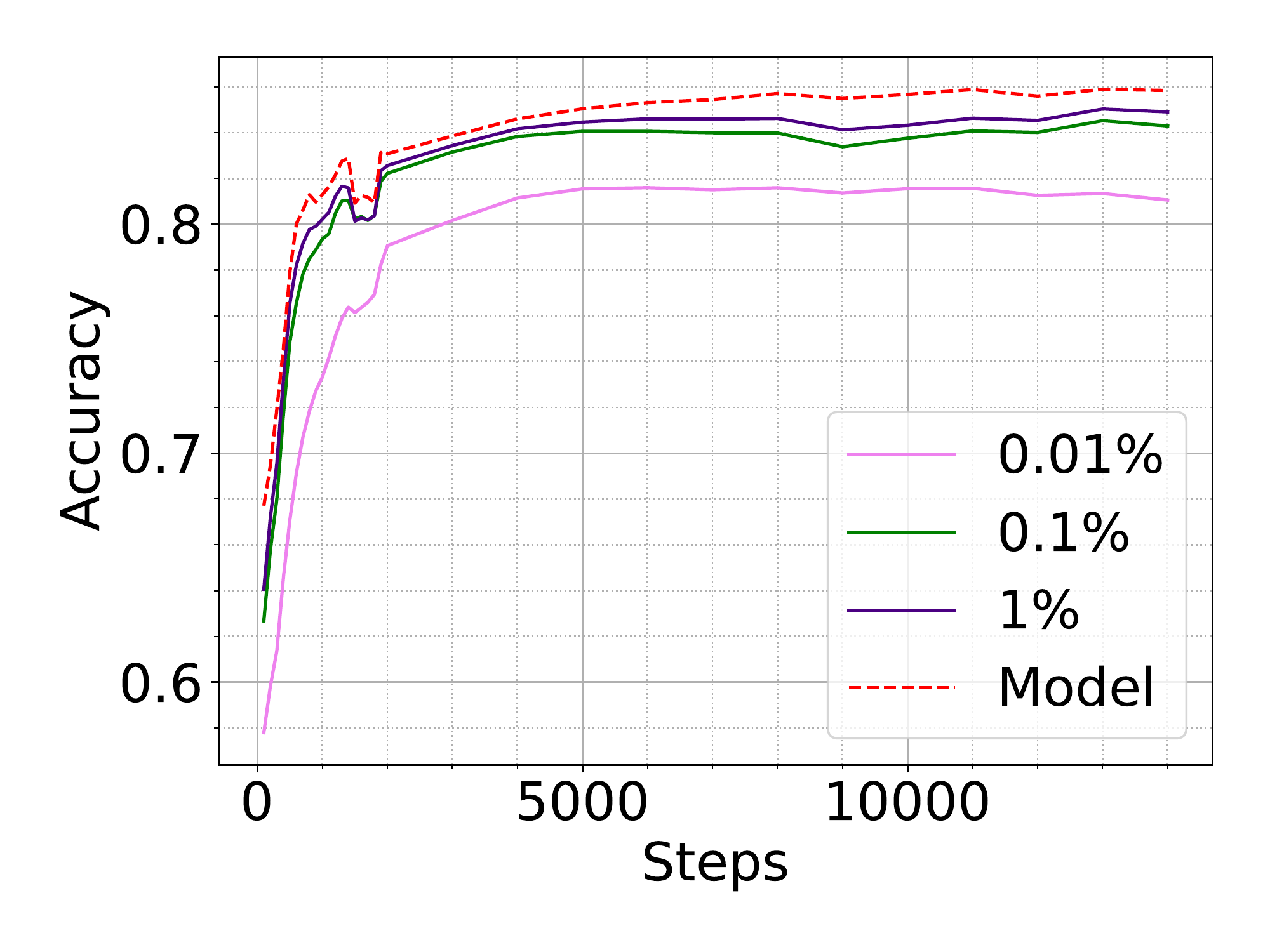} \caption { SNLI, 4096-shot. Checkpoint selected at step 3000. }
    \end{subfigure}\hfill
    \centering
    \begin{subfigure}{0.24\textwidth}
    \centering
    \includegraphics[width=\textwidth]{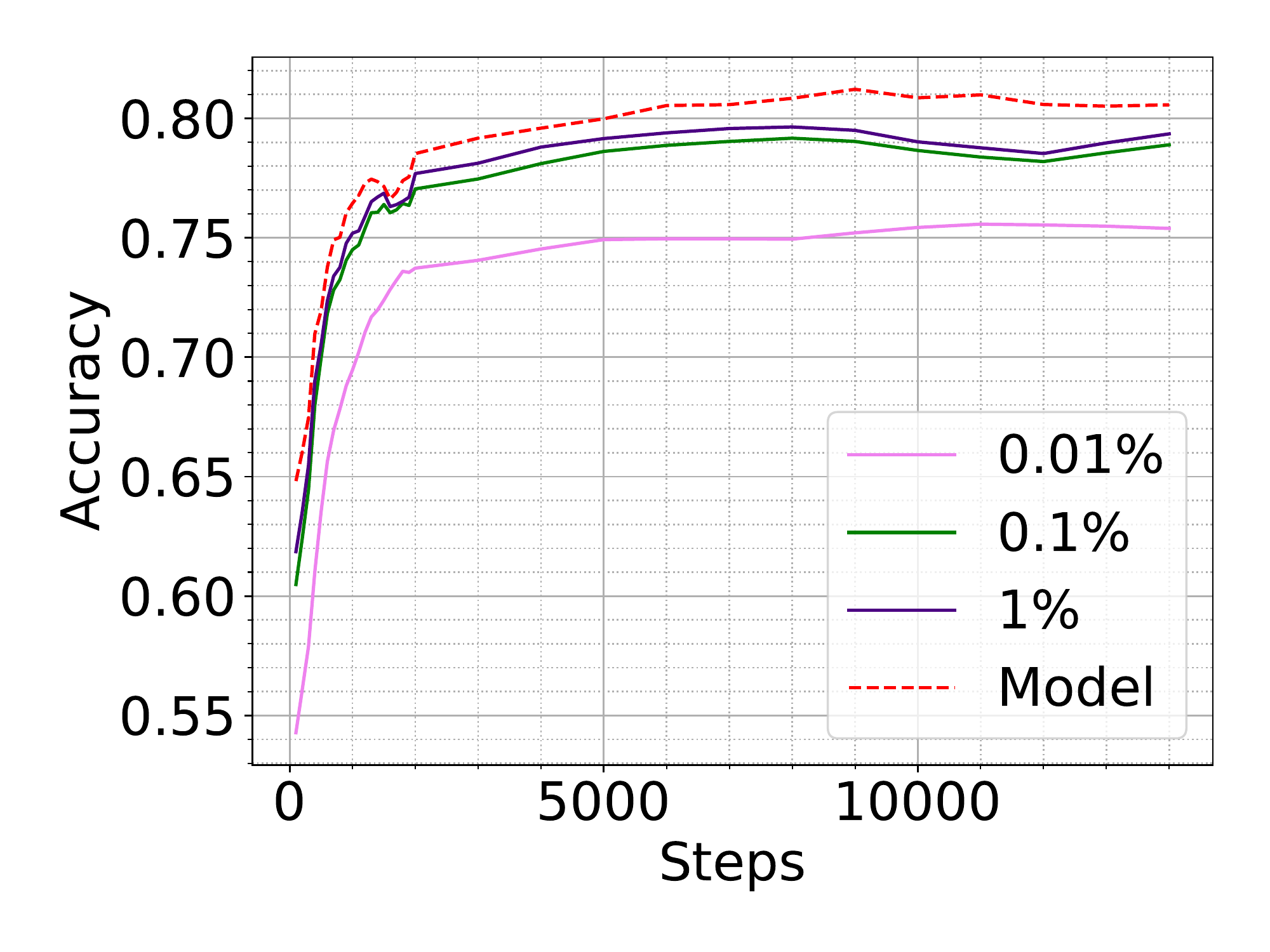}\caption { MNLI, 4096-shot. Checkpoint selected at step 2000. }
    \end{subfigure}\hfill
    \caption{We track the evolution of  the grafting regions of various sizes of an intermediate checkpoint over time.  We created grafted models with the model checkpoints at every step, using the grafting regions. We observe that the grafting region tracks the model's performance over training before saturating.
    }
    \label{fig:stitch_track}
\end{figure*}

%% file: core_skills.tex
\subsection{Core skills are learned first.}
\label{sec:core_skills}


We track the accuracy of grafted models of varying sparsity as fine-tuning progresses in \Cref{fig:stitch_track}.
To do so, we select an intermediate checkpoint to compute grafting regions at different sparsity levels that can match the labels of the model rather than the ground-truth labels.
For each run, we selected the checkpoint, where the train and test accuracy plots seemed to diverge, implying the phase where the model starts to overfit on the training dataset.
We construct grafted model at each model checkpoint using the pre-computed grafting regions and track their performance.
As we can observe, the grafting region tracks the model's performance over training before saturating earlier.
This aligns with the observations of \citet{kalimeris2019sgd} that SGD learns simple functions first.